%% file: main.tex
\documentclass[11pt]{article}

\usepackage[final]{acl}

\input{macros.tex}

\usepackage{times}
\usepackage{latexsym}
\usepackage[T1]{fontenc}
\usepackage[utf8]{inputenc}
\usepackage{microtype}
\usepackage{inconsolata}
\usepackage{graphicx}

\title{\benchmark{}: Benchmarking Multimodal Models for \\Visual Programming in Turtle Graphics}

\author{Chao Wen \\
MPI-SWS \\
\texttt{chaowen@mpi-sws.org} \\
\And
Jacqueline Staub \\
University of Trier \\
\texttt{staub@uni-trier.de} \\
\And
Adish Singla \\
MPI-SWS \\
\texttt{adishs@mpi-sws.org}
}

\begin{document}

\maketitle

%%%%%%%%%%%%%%%%%%%%%%%%%%%%%%%%%%%
%%%%%%%%%%%%%%%%%%%%%%%%%%%%%%%%%%%

\input{0_abstract}
\input{1_intro}
\input{2_relatedwork}

\input{3_background}

\input{4_method}

\input{5_experiment}
\input{6_conclusions}

\input{7_limitations}

\input{8_acknowledgments}

\bibliography{main}
\clearpage
\input{9_appendix}

\end{document}

%% file: macros.tex
\usepackage{booktabs}
\usepackage{subcaption}
\usepackage{multirow}
\usepackage{hyperref}
\usepackage{tikz}
\usepackage{url}           
\usepackage{amsfonts}      
\usepackage{nicefrac}       
\usepackage{xcolor}      
\usepackage{amsmath}
\usepackage{amssymb}
\usepackage{mathtools}
\usepackage{amsthm}
\usepackage[capitalize,noabbrev]{cleveref}

\usepackage{pgfplots}
\pgfplotsset{compat=1.18}

\usepackage{alltt}
\usepackage{adjustbox} 
\usepackage{listings}    
\usepackage{mdframed} 
\usepackage{framed}
\usepackage{wrapfig} 
\usepackage{etoolbox}
\usepackage{enumitem} 
\usepackage{colortbl}

\theoremstyle{plain}

\theoremstyle{definition}

\theoremstyle{remark}

\newcommand{\task}{\texttt{{T}}}
\newcommand{\image}{\texttt{img}}
\newcommand{\instruction}{\texttt{ins}}
\newcommand{\code}{\texttt{{C}}}

\newcommand{\ft}{\textsc{Turtle}}

\newcommand{\categoryBasic}{\textit{Basic geometry}}
\newcommand{\categoryRotation}{\textit{Rotation}}
\newcommand{\categorySpiral}{\textit{Spiral}}

\newcommand{\categoryTranslation}{\textit{Translation}}
\newcommand{\categoryComposite}{\textit{Composite}}
\newcommand{\categoryScaling}{\textit{Scaling}}

\newcommand{\difficultyEasy}{\textit{Easy}}
\newcommand{\difficultyMedium}{\textit{Medium}}
\newcommand{\difficultyHard}{\textit{Hard}}

\newcommand{\platformAll}{XLogoOnline}
\newcommand{\benchmark}{\textsc{TurtleAI}}
\newcommand{\eval}{\textsc{TurtleAI}-\text{Eval}}

\newcommand{\datagenOurs}{\textsc{TurtleAI}-Datagen}

\newcommand{\datasetAll}{\textsc{TurtleAI}-DS}
\newcommand{\datasetBasic}{\textsc{TurtleAI}-DSReal}
\newcommand{\datasetHand}{\textsc{TurtleAI}-DSCraft}
\newcommand{\datasetSyn}{\textsc{TurtleAI}-DSSyn}

\newcommand{\datasetTrain}{\textsc{TurtleAI}-Train}

\newcommand{\datasetBasicNoPrefix}{DSReal}
\newcommand{\datasetHandNoPrefix}{DSCraft}
\newcommand{\datasetSynNoPrefix}{DSSyn}
\newcommand{\datasetAllNoPrefix}{DS}

\newcommand{\gptfouro}{{GPT-4o}} 
\newcommand{\gptfourv}{{GPT-4V}} 
\newcommand{\llavaov}{{Llava-OneVision}}
\newcommand{\qwentwovl}{{Qwen2-VL}}  
\newcommand{\molmo}{{Molmo}} 
\newcommand{\pixtral}{{Pixtral}} 
\newcommand{\nvlmd}{{NVLM-1.0-D}} 
\newcommand{\internvltwo}{{InternVL2}} 
 
\newcommand{\glmfourv}{{GLM-4V}}

\definecolor{promptinputcolor}{rgb}{0,0,2.55}
\newcommand{\promptheader}[1]{{\large{\textcolor{blue}{\textbf{#1}}}}}

\DeclareRobustCommand{\successmark}{\tikz[baseline=-0.5ex]\node[circle,draw=none,fill=green!60!black,inner sep=0.08pt]{\textcolor{white}{\tiny$\checkmark$}};}
\DeclareRobustCommand{\failmark}{\tikz[baseline=-0.5ex]\node[circle,draw=none,fill=red!70!white,inner sep=0.08pt]{\textcolor{white}{\tiny$\times$}};}

\lstdefinestyle{python}{
    language=Python,
    basicstyle=\ttfamily\tiny,
    keywordstyle=\color{blue}\bfseries,
    commentstyle=\color{gray}\itshape,
    stringstyle=\color{red},
    identifierstyle=\color{black},
    frame=single,
    rulecolor=\color{black},
    tabsize=2,
    breaklines=true,
    showstringspaces=false,
    backgroundcolor=\color{lightgray!20},
    morekeywords={self, cls},
    emphstyle=\color{magenta},
    framesep=0pt,
    xleftmargin=0pt,
    xrightmargin=0pt,
    framexleftmargin=1pt,
    framexrightmargin=0pt,
    aboveskip=2pt,
    belowskip=0pt,
}

%% file: 0_abstract.tex
% !TEX root =  main.tex
%%%%%%%%%%%%%%%%%%%%%%%%%%%%%%%%%%%%%%%%%%%%%%%
%%%%%%%%%%%%%%%%%%%%%%%%%%%%%%%%%%%%%%%%%%%%%%%

\begin{abstract}

Vision-language models (VLMs) have been explored for visual programming, where they generate code to solve visual tasks. However, most prior work focuses on visual programming for productivity; it remains unclear how well current VLMs perform on education-oriented visual programming and what factors limit their performance. To bridge this gap, we introduce \benchmark{}, a benchmark containing 823 tasks curated based on real-world visual programming tasks in the Turtle Graphics domain. Solving these tasks requires models to perceive geometric patterns, reason about spatial relationships, and synthesize Python code that faithfully reproduces geometric patterns. We evaluate 20+ VLMs, including GPT-5, GPT-4o, and Qwen2-VL-72B, and find that they struggle significantly, with most achieving success rates below 30\%. To address these limitations, we propose a data generation technique that requires only a small set of seed samples. Fine-tuning Qwen2-VL-72B on the resulting synthetic data yields an improvement of about 20\% on real-world tasks. Our failure analysis reveals that GPT-4o struggles with spatial reasoning and precise visual replication, whereas fine-tuning primarily improves the alignment between visual reasoning and code implementation.

\end{abstract}

%% file: 1_intro.tex
% !TEX root =  main.tex
%%%%%%%%%%%%%%%%%%%%%%%%%%%%%%%%%%%%%%%%%%%%%%%
%%%%%%%%%%%%%%%%%%%%%%%%%%%%%%%%%%%%%%%%%%%%%%%

%%%%%%%%%%%%%%%%%%%%%%%%%%%%%%%%%%%%%%%%% 

\section{Introduction} \label{sec.intro}

Vision-language models (VLMs) have shown strong performance on visual understanding tasks~\citep{DBLP:conf/icml/RameshPGGVRCS21,DBLP:conf/icml/RadfordKHRGASAM21,DBLP:conf/cvpr/YueNZ0LZSJRSWYY24,DBLP:conf/iclr/LuBX0LH0CG024}. Beyond visual understanding, recent work has explored \emph{visual programming}, in which VLMs are given visual inputs and prompted to generate executable code that solves visually grounded tasks, such as solving visual reasoning problems~\cite{DBLP:conf/cvpr/GuptaK23,DBLP:conf/emnlp/LiTHLH024,DBLP:conf/acl/WenSS25} and generating visual artifacts (e.g., UIs, SVGs)~\cite{DBLP:conf/naacl/SiZLYLY25,DBLP:conf/iclr/0002SLS0JXZLZLN25,DBLP:conf/naacl/WuLGGLWSL25,DBLP:conf/aaai/RodriguezPALR0P25,DBLP:conf/emnlp/ZouCZL24,DBLP:journals/corr/abs-2512-18388}.

\looseness-1
However, most existing work on visual programming is motivated by improving \emph{productivity} through code generation in applied domains, such as software engineering, data visualization, and computer graphics~\cite{DBLP:conf/www/Gui0LZC0SCZ0025,DBLP:conf/naacl/SiZLYLY25,DBLP:conf/naacl/WuLGGLWSL25,DBLP:conf/iclr/0002SLS0JXZLZLN25,DBLP:conf/aaai/RodriguezPALR0P25,DBLP:conf/emnlp/ZouCZL24}. Beyond productivity, visual programming is also a central paradigm in education-oriented settings, particularly in K-12 programming education for developing students' computational thinking~\cite{grover2013computational}, as exemplified by platforms such as Scratch~\cite{DBLP:journals/jeric/MaloneyRRSE10}, Code.org~\cite{codeorg}, and Blockly~\cite{blocklygames}.
Despite its widespread use in education and the potential of generative models to support it, education-oriented visual programming remains largely underexplored. It remains unclear how well current VLMs perform on educational visual programming tasks and which factors most limit their performance.

\input{figs/1_intro/examples.tex}

To bridge this gap, we introduce \benchmark{}, a benchmark for assessing VLMs' visual programming capabilities on education-oriented Turtle Graphics tasks~\cite{python_turtle}. The benchmark comprises 823 visual programming tasks curated based on the visual programming platform XLogoOnline~\cite{xlogoonline}, which is used by tens of thousands of students each year~\cite{DBLP:journals/eatcs/Staub21}. In \benchmark{}, each task requires a VLM to generate Python code that reproduces a target image. \autoref{fig:example_code} shows representative target images along with corresponding model outputs.
Solving these tasks requires recognizing visual patterns, reasoning about spatial relationships (e.g., layout, scale, position, and angles), and translating this analysis into executable Python code that accurately reproduces the target image.

We evaluate 20+ state-of-the-art VLMs on \benchmark{} and find that they struggle: even leading models such as GPT-5~\cite{gpt5}, \gptfouro{}~\citep{gpt4o}, and \qwentwovl{}-72B~\citep{DBLP:journals/corr/abs-2409-12191} achieve success rates below 30\% on real-world tasks. To improve performance, we propose a data generation technique that synthesizes training data from a small set of seed samples; fine-tuning on this synthetic data yields roughly a 20\% improvement on real-world tasks.
Our failure analysis reveals that \gptfouro{} most often fails due to limitations in spatial reasoning and faithful visual replication, whereas \qwentwovl{}-72B more frequently fails to align its code implementation with its visual reasoning; fine-tuning primarily reduces these alignment errors rather than improving visual understanding or spatial reasoning.

Our contributions are threefold.
First, we introduce \benchmark{}, a multimodal benchmark for evaluating VLMs in the Turtle Graphics domain, with datasets curated based on real-world visual programming tasks.
Second, we propose \datagenOurs{}, a data generation technique that creates large-scale synthetic training data, and show that fine-tuning VLMs on this data improves performance on \benchmark{}.
Third, we conduct extensive experiments and analyses on \benchmark{}, providing insights into VLMs' capabilities and limitations in educational visual programming settings. We
publicly release our benchmark for further research.\footnote{\url{https://github.com/machine-teaching-group/acl2026-turtleai}}

%% file: figs/1_intro/examples.tex
% !TEX root =  main.tex
%%%%%%%%%%%%%%%%%%%%%%%%%%%%%%%%%%%%%%%%%%%%%%%
%%%%%%%%%%%%%%%%%%%%%%%%%%%%%%%%%%%%%%%%%%%%%%%

\begin{figure*}[ht]
  \begin{subfigure}[b]{0.542\textwidth}
    \centering
    \includegraphics[width=\textwidth]{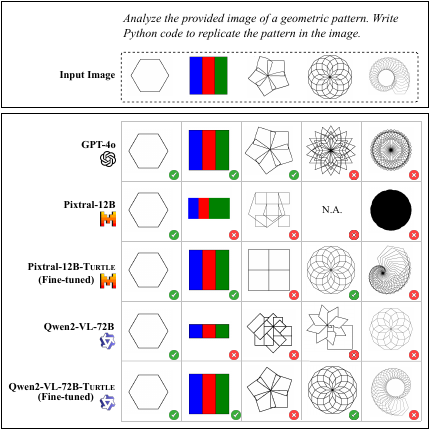}
    \caption{VLMs' outputs for replicating different input images.}
    \vspace{-0.25cm}
    \label{fig:example_response}
  \end{subfigure}
  \begin{subfigure}[b]{0.452\textwidth}
    \centering
    \framebox{%
      \begin{minipage}{\linewidth}
        \centering
        \includegraphics[width=0.99\textwidth]{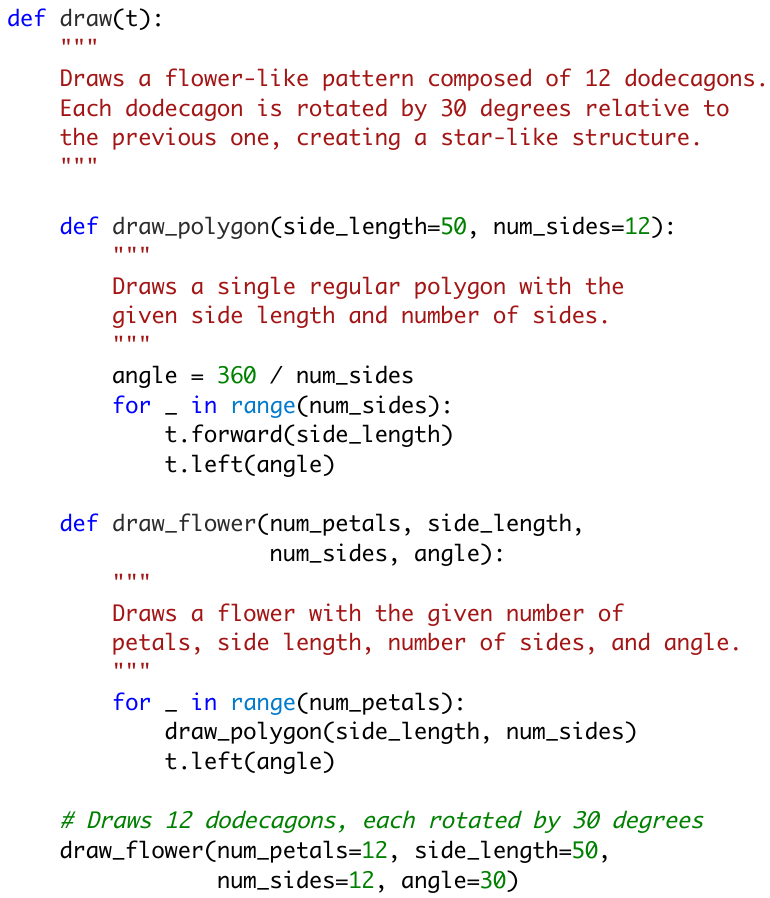}
      \end{minipage}%
    }
    \caption{Solution code for replicating the image \raisebox{-0.25\height}{\includegraphics[width=0.045\textwidth]{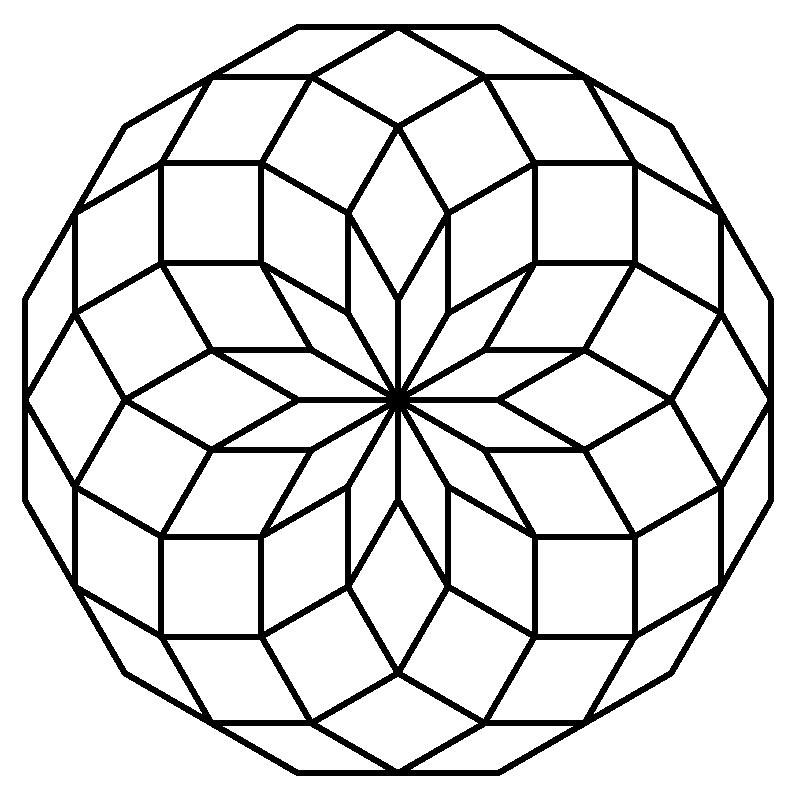}}.}
    \vspace{-0.2cm}
    \label{fig:code}
  \end{subfigure}
  \caption{Outputs of VLMs on visual-to-code generation tasks and an example solution code. (a) shows the input images and the visual outputs produced by executing each VLM's generated Python code, with success ($\successmark$) or failure ($\failmark$) shown for each output. (b) shows an example solution code for replicating the image {\includegraphics[width=0.02\textwidth]{figs/1_intro/task_examples/midi_9c--chatcmpl-562deed6-fcf0-47a5-b79e-e9d1a40edb6b.png}}.}
  \label{fig:example_code}
  \vspace{-0.4cm}
\end{figure*}

%% file: 2_relatedwork.tex
% !TEX root =  main.tex
%%%%%%%%%%%%%%%%%%%%%%%%%%%%%%%%%%%%%%%%%%%%%%%
%%%%%%%%%%%%%%%%%%%%%%%%%%%%%%%%%%%%%%%%%%%%%%%

%%%%%%%%%%%%%%%%%%%%%%%%%%%%%%%%%%%%%%%%% 

\section{Related Work}\label{sec.related-work}

\looseness-1\paragraph{Visual programming benchmarks.} Visual programming involves generating executable code to solve visually grounded tasks.
Prior work explores visual programming across various productivity-oriented domains such as UI generation~\cite{DBLP:conf/naacl/SiZLYLY25,DBLP:conf/www/Gui0LZC0SCZ0025}, data visualization~\cite{DBLP:conf/naacl/WuLGGLWSL25,DBLP:conf/iclr/0002SLS0JXZLZLN25}, and graphics synthesis~\cite{DBLP:conf/emnlp/ZouCZL24,DBLP:conf/aaai/RodriguezPALR0P25}. 
Recent work has also explored visual programming in educational contexts. However, these benchmarks typically focus on discrete, grid-based navigation or multiple-choice reasoning, often utilizing domain-specific languages with restricted syntax~\citep{DBLP:conf/acl/WenSS25,DBLP:journals/corr/abs-2406-09891,DBLP:conf/icer/Singla22}. In contrast, our benchmark targets the synthesis of complex geometric patterns in continuous space using Python, which necessitates a more expressive and complex code space.

\paragraph{Program synthesis for inverse graphics.}
\looseness-1
Our tasks can be viewed as a form of inverse graphics, where the goal is to generate code that reconstructs a given visual input.
Prior work has studied inverse graphics in domains such as SVGs~\citep{DBLP:conf/aaai/RodriguezPALR0P25,DBLP:conf/emnlp/ZouCZL24}, scientific figures in \LaTeX~\citep{DBLP:conf/nips/BelouadiPE24}, and charts~\citep{DBLP:conf/naacl/WuLGGLWSL25,DBLP:conf/iclr/0002SLS0JXZLZLN25}. 
Existing work has also explored Turtle Graphics~\citep{DBLP:conf/pldi/EllisWNSMHCST21,DBLP:conf/nips/LiE24,rismanchian2024turtlebench}. The most relevant benchmark is TurtleBench~\citep{rismanchian2024turtlebench}. Our work differs in three key ways: (i) Datasets: Our datasets are curated based on real-world visual programming tasks actively used by students, whereas TurtleBench relies on manually created tasks; (ii) Scope: Our benchmark spans six task categories, three difficulty levels, and three datasets, whereas TurtleBench is built around a single dataset with two task types and simpler black-and-white geometric patterns; and (iii) Scale: We evaluate 20+ VLMs (including fine-tuning) on 823 tasks, while they evaluated 3 VLMs on 260 tasks.

%% file: 3_background.tex
% !TEX root =  main.tex
%%%%%%%%%%%%%%%%%%%%%%%%%%%%%%%%%%%%%%%%%%%%%%%
%%%%%%%%%%%%%%%%%%%%%%%%%%%%%%%%%%%%%%%%%%%%%%%

%%%%%%%%%%%%%%%%%%%%%%%%%%%%%%%%%%%%%%%%% 

\section{Background and Synthesis Objective} \label{sec:background}

In this section, we provide background on Turtle Graphics and XLogoOnline, and introduce the program synthesis objective.

\paragraph{Background on Turtle Graphics.} \label{sec:background:turtle_graphics}
Turtle Graphics is a programmable method for creating vector graphics and has been used in K-12 programming education to teach programming concepts and computational thinking~\cite{codehs_tracy,xlogoonline,turtleacademy}.
In Turtle Graphics, one can create vector graphics using a relative cursor (the ``turtle'') on a Cartesian plane~\citep{python_turtle}. Basic commands like ``forward'', ``turn left'', and ``pen down'' control the turtle's movement to draw lines and shapes. These commands can be combined with programming constructs such as loops, conditionals, and functions to generate visually appealing geometric patterns.

\paragraph{Background on XLogoOnline platform.} \label{sec:background:xlogoonline}
XLogoOnline~\cite{xlogoonline} is a Logo-based visual programming platform~\cite{pea1987logo} used by tens of thousands of students each year~\cite{DBLP:journals/eatcs/Staub21}. 
It offers four levels (Mini, Midi, Maxi, Mega); the Midi (grades 3-4) and Maxi (grades 5-6) levels provide Turtle Graphics tasks, where students learn programming concepts by replicating visual patterns through writing code. We curate our benchmark primarily based on these tasks (e.g., the hexagon pattern in \autoref{fig:example_code}).

\paragraph{Task specification.} \label{sec:background:task_specification}
We define a visual programming task in Turtle Graphics as $\task{}:=(\image{}, \instruction{})$, a tuple consisting of a target image $\image{}$ and a text-based instruction $\instruction{}$. The target image specifies the desired visual output, while the instruction specifies the requirements for generating the code that will replicate the pattern shown in the target image.

\paragraph{Code specification.} \label{sec:background:code_specification}
The code space for Turtle Graphics tasks is defined using the Python programming language.
A \emph{solution code} for a task $\task{}$ is Python code $\code{}$ that, after being executed, can accurately replicate the target image $\image{}$ and satisfy the requirements specified by the instruction $\instruction{}$.
For consistent evaluation, the task's instruction requires the solution code to be synthesized as a function \texttt{draw(t)}, which takes a turtle object \texttt{t} as input.
For instance, \autoref{fig:code} shows a solution code that generates the image \raisebox{-0.25\height}{\includegraphics[width=0.023\textwidth]{figs/1_intro/task_examples/midi_9c--chatcmpl-562deed6-fcf0-47a5-b79e-e9d1a40edb6b.png}}.

\paragraph{Program synthesis objective.} \label{sec:background:objective}
\looseness-1
The synthesis objective is to develop a synthesizer function, $\mathcal{M}: \task \rightarrow \code$, which generates a solution code $\code$ for a given task $\task$ in Turtle Graphics.
To evaluate $\mathcal{M}$ on a task $\task$, we first use $\mathcal{M}$ to synthesize a code $\hat{\code}$ that contains a Python function \texttt{draw(t)}. To evaluate the correctness of this Python function,
one straightforward way is to compare the images generated by $\hat{\code}$ and $\code$ pixel-by-pixel. However, this pixel-wise comparison fails to account for differences in the size, position, or line width of patterns being drawn in images~\citep{DBLP:conf/issep/MarbachMS22}.
In the next section, as part of our benchmark \benchmark{}, we will address this by introducing an evaluation framework.

%% file: 4_method.tex
% !TEX root =  main.tex
%%%%%%%%%%%%%%%%%%%%%%%%%%%%%%%%%%%%%%%%%%%%%%%
%%%%%%%%%%%%%%%%%%%%%%%%%%%%%%%%%%%%%%%%%%%%%%%

%%%%%%%%%%%%%%%%%%%%%%%%%%%%%%%%%%%%%%%%% 

\input{figs/4_method/fig_benchmark_overview}

\input{figs/4_method/fig_dataset_stats}

\section{The \benchmark{} Benchmark} \label{sec:benchmark}

\benchmark{} is a visual programming benchmark for Turtle Graphics.
\autoref{fig:benchmark_overview} illustrates the components of \benchmark{}, which consists of (i) \datasetAll{}, a collection of evaluation datasets; (ii) \eval{}, an evaluation framework for assessing the correctness of synthesized code; and (iii) \datagenOurs{}, a data generation technique for synthesizing high-quality data. We describe each component in the following sections.

\subsection{Evaluation Datasets \datasetAll{}} \label{sec:benchmark:dataset}

\benchmark{} includes \datasetAll{}, a collection of $823$ evaluation tasks organized into three distinct datasets. \autoref{fig:dataset_stats} shows example tasks and the distribution of these datasets.

\paragraph{\datasetBasic{} (Size $\textbf{102}$).}
\looseness-1
This dataset contains real-world tasks curated from the Midi and Maxi levels of \platformAll{}~\citep{xlogoonline}. The tasks were originally designed by experts to teach programming concepts to students in grades 3-6. We manually wrote Python reference solutions for each task and verified their correctness.

\paragraph{\datasetHand{} (Size $\textbf{102}$).}
This dataset contains hand-crafted tasks created by manually hand-drawing each image from \datasetBasic{}, while reusing the same solution code. It is designed to evaluate model robustness to hand-drawn inputs that may include imperfections.

\paragraph{\datasetSyn{} (Size $\textbf{619}$).} This dataset contains synthetic tasks generated using \datasetBasic{} as a seed dataset for our data generation technique (see Section~\ref{sec:method:datagen}), followed by manual selection of high-quality tasks. See Appendix~\ref{sec:appendix:dataset} for more details about the dataset generation process.

\subsection{Evaluation Framework \eval{}} \label{sec.eval}

\looseness-1
In our benchmark, we evaluate whether the synthesized code $\hat{\code}$ produces a drawing that is \emph{semantically} equivalent to the ground-truth code $\code$. One natural way is to compare the images generated by $\hat{\code}$ and $\code$ pixel-by-pixel. However, pixel-wise comparison is sensitive to non-semantic factors (e.g., translation, scale, and line width); for example, the same square drawn at $(0,0)$ and $(1,1)$ would be judged different despite being semantically identical. To address this, we introduce \eval{}, an evaluation framework that canonicalizes drawings before comparison to achieve invariance to translation, scale, and line width.
Specifically, \eval{} executes $\code$ and $\hat{\code}$ with a custom Turtle emulator that records drawing states (e.g., vertices, strokes, fills, colors). It then normalizes coordinates to a fixed range, centers the drawing at the origin, and standardizes line width to $1$, producing canonical renderings $\image$ and $\hat{\image}$.\footnote{For simplicity, we use $\image$ to denote the canonical rendering when the context is clear.} Finally, we compare $\image$ and $\hat{\image}$ using two methods:

\begin{itemize}[leftmargin=*]
    \item \textit{Symbolic comparison}: we compare $\image$ and $\hat{\image}$ pixel-wise. A prediction is marked as \emph{success} if the fraction of matching pixels exceeds a pre-defined threshold; otherwise, it is \emph{fail}.

    \item \textit{Embedding-based comparison}: We embed $\image$ and $\hat{\image}$ using a pre-trained image encoder (e.g., ResNet18~\citep{DBLP:conf/cvpr/HeZRS16}) and compute a normalized similarity score in $[0,1]$. A prediction is \emph{success} if the similarity exceeds a threshold; otherwise, it is \emph{fail}.
\end{itemize}
Thresholds are selected based on manual evaluations, i.e., by choosing the values that maximize agreement with human judgments. This yields $99.1\%$ accuracy for symbolic comparison and $98.1\%$ for embedding-based comparison against human judgments (see Appendix~\ref{sec:appendix:evaluation_reliability}).

\subsection{Data Generation Technique \datagenOurs{}} \label{sec:method:datagen}

As mentioned in Section~\ref{sec.intro}, existing VLMs struggle on \benchmark{}. 
To make these models practically useful in educational settings, it is important to improve their performance. One natural way is to fine-tune VLMs for Turtle Graphics; however, fine-tuning is limited by the scarcity of training data in this domain. To address this, we propose \datagenOurs{}, a technique that leverages large models to synthesize high-quality training data for Turtle Graphics.
The key idea is to evolve a large dataset from a small seed through iterative stages of code mutation and elite selection.
We describe the two stages below (see \autoref{fig:data_framework}).

\paragraph{Stage 1: Code mutation.}
\looseness-1
This stage aims to generate a larger set of codes from a given small seed dataset $\mathcal{D}_t = \{(\image_i, \code_i)\}$~\citep{DBLP:conf/iclr/XuSZG0FTLJ24,chao2024xlogo}. This can be done by pre-defining instructions for large language models (LLMs) to mutate the codes~\citep{DBLP:conf/iclr/XuSZG0FTLJ24}. For example, one can use the instruction ``add a loop to the code'' to guide the LLM to mutate an input code $\code_{\text{in}}$:
\begin{equation*}
{\small
    \code_{\text{out}} = \text{LLM}(\code_{\text{in}}, \text{instruction}=``\text{add a loop to the code}").
}
\end{equation*}
\looseness-1However, a fixed set of pre-defined instructions may limit the diversity of mutated codes by capturing only specified mutation patterns.  
To address this, we use an LLM to infer mutation patterns. Specifically, the LLM is given a pair of reference codes $(\code_{\texttt{ref}_1}, \code_{\texttt{ref}_2})$ and prompted to infer the high-level mutation pattern $m(\code_{\texttt{ref}_1}, \code_{\texttt{ref}_2})$. It then applies this pattern to another input code $\code_{\text{in}}$, producing a new mutated output code:
\begin{equation*}
{\small
    \code_{\text{out}} = \text{LLM}\big(\code_{\text{in}}, \text{instruction}=m({\code_{\texttt{ref}_1}, \code_{\texttt{ref}_2}})\big).
}
\end{equation*}
We apply this process to extend the seed dataset: for each $\code_{\text{in}} \in \mathcal{D}_t$, we randomly sample $p$ reference code pairs from $\mathcal{D}_t$ to generate mutated codes. These codes are then executed to obtain images, producing a larger dataset of image-code pairs.

\paragraph{Stage 2: Elite selection.}
\looseness-1
After obtaining a larger dataset of image-code pairs, we first use a clustering algorithm to remove duplicate image-code pairs. Then we use a VLM to score image quality based on predefined rubrics. Finally, we select the top $k\%$ image-code pairs, which serve as the seed dataset for the next iteration $t+1$. After $n$ iterations, this produces a large-scale dataset $\mathcal{D}_{t+n}$.

\input{figs/4_method/data_framework}

\paragraph{Generating training dataset \datasetTrain{}.}
\looseness-1
We construct the training dataset \datasetTrain{} using \datagenOurs{} in three steps. First, we start from a manually curated seed set of only $10$ image-code pairs. 
Second, we run \datagenOurs{} for five iterations to expand the seed set.
Third, we enrich each resulting $(\image{}, \code{})$ pair in the expanded dataset with Chain-of-Thought (CoT) labels~\citep{DBLP:conf/nips/Wei0SBIXCLZ22,star_cot_datasynthesis} by prompting a VLM to describe the image, explain the solution code step by step, and add documentation. 
We find that adding CoT labels improves fine-tuning and generalization performance (see Appendix~\ref{sec:exp_ood_analysis}). 
Finally, we obtain the final training dataset containing $738,126$ samples. 
More details are provided in Appendix~\ref{sec:appendix:dataset:generation}.

%% file: figs/4_method/fig_benchmark_overview.tex
% !TEX root =  main.tex
%%%%%%%%%%%%%%%%%%%%%%%%%%%%%%%%%%%%%%%%%%%%%%%
%%%%%%%%%%%%%%%%%%%%%%%%%%%%%%%%%%%%%%%%%%%%%%%

%%%%%%%%%%%%%%%%%%%%%%%%%%%%%%%%%%%%%%%%% 

\begin{figure*}[ht]
    \centering
    \includegraphics[width=0.97\textwidth]{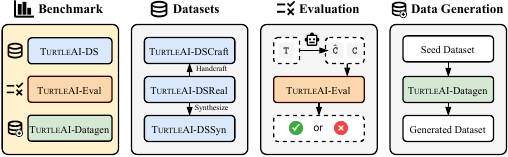}
    \caption{Overview of the \benchmark{} benchmark. It comprises three key components: (i) a collection of datasets \datasetAll{}, (ii) an evaluation framework \eval{} for assessing the correctness of generated code, and (iii) a data generation technique \datagenOurs{} for generating synthetic datasets.}
    \label{fig:benchmark_overview}
\end{figure*} 

%% file: figs/4_method/fig_dataset_stats.tex
% !TEX root =  main.tex
%%%%%%%%%%%%%%%%%%%%%%%%%%%%%%%%%%%%%%%%%%%%%%%
%%%%%%%%%%%%%%%%%%%%%%%%%%%%%%%%%%%%%%%%%%%%%%%

%%%%%%%%%%%%%%%%%%%%%%%%%%%%%%%%%%%%%%%%% 

\begin{figure*}[ht!]
    \centering
    \begin{minipage}[b]{0.26\textwidth}
        \begin{subfigure}{\linewidth}
            \includegraphics[width=\columnwidth]{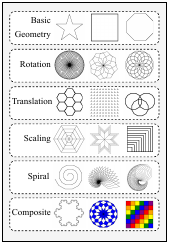}
            \caption{Examples for each category}
            \label{fig:examples.six_categories}
        \end{subfigure}
    \end{minipage}
    \hfill
    \begin{minipage}[b]{0.72\textwidth}
        \begin{subfigure}{0.48\textwidth}
            \centering
            \includegraphics[width=\columnwidth]{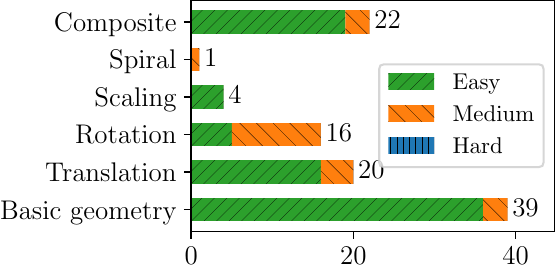}
            \caption{\datasetBasicNoPrefix{}/\datasetHandNoPrefix{}}
            \label{fig:dataset_stats_graphics}
        \end{subfigure}
        \hfill
        \begin{subfigure}{0.48\textwidth}
            \centering
            \includegraphics[width=\columnwidth]{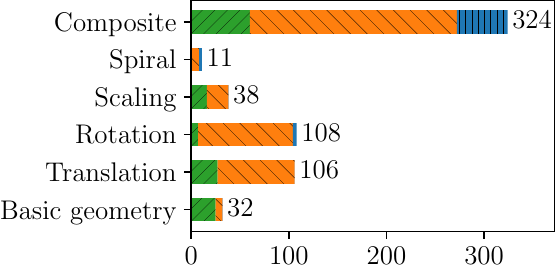}
            \caption{\datasetSynNoPrefix{}}
            \label{fig:dataset_stats_graphics_syn}
        \end{subfigure}
        \vspace{0.01\textwidth}
        
        \begin{subfigure}{0.48\textwidth}
            \centering
            \includegraphics[width=\columnwidth]{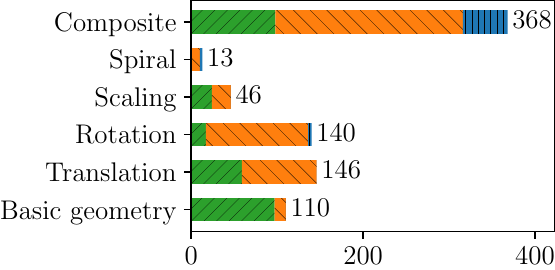}
            \caption{\datasetAll{}}
            \label{fig:dataset_stats_graphics_plus}
        \end{subfigure}
        \hfill
        \begin{subfigure}{0.49\textwidth}
            \centering
            \scalebox{0.72}{
            \begin{tabular}{lrrrr}
                \toprule
                Dataset & Easy & Medium & Hard & Total \\
                \midrule
                \datasetBasicNoPrefix{} & 80 & 22 & 0 & 102 \\
                \datasetHandNoPrefix{} & 80 & 22 & 0 & 102 \\
                \datasetSynNoPrefix{} & 135  & 425 & 59 & 619 \\
                \midrule
                \datasetAllNoPrefix{} & 295 & 469 & 59 & 823 \\
                \bottomrule
            \end{tabular}
            }
            \vspace{0.06\linewidth}
            \caption{Difficulty distributions}
            \label{fig:dataset_difficulty_distribution}
        \end{subfigure}
    \end{minipage}
    \caption{Dataset composition and statistics. Tasks are categorized into six task categories and three difficulty levels. (a) shows representative examples for each category. (b-d) show the distributions of these categories in \datasetBasicNoPrefix{}, \datasetSynNoPrefix{}, and \datasetAll{}, respectively. Note that \datasetHandNoPrefix{} has the same distribution as \datasetBasicNoPrefix{}, as it is a hand-drawn rendering of the same set of tasks. (e) shows the difficulty distribution across different datasets. The detailed labeling process for task categories and difficulty levels is provided in Appendix~\ref{sec:appendix:dataset}.}
    \label{fig:dataset_stats}
\end{figure*}

%% file: figs/4_method/data_framework.tex
% !TEX root =  main.tex
%%%%%%%%%%%%%%%%%%%%%%%%%%%%%%%%%%%%%%%%%%%%%%%
%%%%%%%%%%%%%%%%%%%%%%%%%%%%%%%%%%%%%%%%%%%%%%%

%%%%%%%%%%%%%%%%%%%%%%%%%%%%%%%%%%%%%%%%% 

\begin{figure}[t!]
  \centering
  \includegraphics[width=\linewidth]{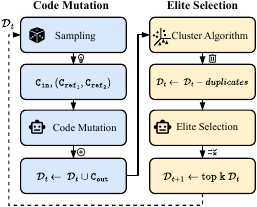}
  \caption{Overview of the data generation technique \datagenOurs{}. It consists of two stages: (i) code mutation, which creates diverse code variants; and (ii) elite selection, which deduplicates candidates and retains high-quality samples.}
  \label{fig:data_framework}
\end{figure}

%% file: 5_experiment.tex
% !TEX root =  main.tex
%%%%%%%%%%%%%%%%%%%%%%%%%%%%%%%%%%%%%%%%%%%%%%%
%%%%%%%%%%%%%%%%%%%%%%%%%%%%%%%%%%%%%%%%%%%%%%%

%%%%%%%%%%%%%%%%%%%%%%%%%%%%%%%%%%%%%%%%% 

\section{Experimental Evaluation}
\label{sec.exp}

\looseness-1
In this section, we evaluate VLMs on \benchmark{}. We first describe the experimental setup in Section~\ref{sec:exp_setup} and present the main results in Section~\ref{sec:exp_results}. Then we provide fine-grained performance analyses (Section~\ref{sec:experiments:comparative}), failure analysis (Section~\ref{sec:exp_failure_analysis}), and code length analysis (Section~\ref{sec:length_analysis}).

\subsection{Experimental Setup} \label{sec:exp_setup}

\input{figs/5_exp/tab_main_results}

\paragraph{Evaluation procedure and metrics.}  
\looseness-1
We use the datasets described in Section~\ref{sec:benchmark:dataset} for evaluation.
Each evaluation dataset consists of $(\task{}, \code{})$ pairs. For each pair, we provide the task image $\image{}$ and a fixed prompt instructing the model to generate the code snippet $\hat{\code{}}$ in the desired format. The model may also output explanations, but we only extract $\hat{\code{}}$.  
Then we evaluate $\hat{\code{}}$ against the ground-truth $\code{}$ using our evaluation framework (see Section~\ref{sec.eval}). The \emph{success rate} is the number of successful predictions divided by the total number of samples. We report both symbolic- and embedding-based success rates. For the main experiments, success rates are based on greedy decoding; additional Pass@K results from random sampling are provided in Appendix~\ref{sec:appendix:experiments:topk}.

\paragraph{Models evaluated.}  
We compare various VLMs:  
(i) \emph{Reasoning VLMs}, including GPT-family models: GPT-5~\citep{gpt5}, o3, and o4-mini~\citep{o3_and_o4_mini};  
(ii) \emph{Non-reasoning base VLMs}, covering model families from GPT~\citep{gpt4o}, Qwen~\citep{DBLP:journals/corr/abs-2409-12191}, Molmo~\citep{DBLP:journals/corr/abs-2409-17146}, Llava~\citep{DBLP:journals/corr/abs-2408-03326}, Pixtral~\citep{agrawal2024pixtral12b}, and InternVL~\citep{DBLP:journals/corr/abs-2504-10479};  
and (iii) \emph{Fine-tuned VLMs}, trained on our \datasetTrain{} dataset with 738k samples, denoted with the \ft{} suffix.
Full model details and fine-tuning details are in Appendix~\ref{sec:appendix:finetuning_details}.

\subsection{Main Results} \label{sec:exp_results}

\paragraph{Current VLMs struggle, even on real-world tasks designed for students in grades 3-6.}
As shown in \autoref{fig:main_performance}, all evaluated models struggle on our benchmark. On the real-world dataset \datasetBasicNoPrefix{} containing tasks designed for students in grades 3-6, even the top-performing model, o3, achieves only a $40.2\%$ symbolic success rate, suggesting current VLMs are still far from ready for classroom deployment. Performance is even lower on the full dataset \datasetAll{}: among base models, o4-mini achieves the highest symbolic success rate at $15.9\%$, while the strongest open-source base model, Qwen2.5-VL, reaches only $6.80\%$. Overall, these results show that \benchmark{} is challenging for all existing VLMs.

\paragraph{Fine-tuning improves performance, but yields limited gains on hand-drawn tasks.}
Fine-tuning models (e.g., \pixtral{}-12B-\ft{}, \qwentwovl{}-7B-\ft{}, and \qwentwovl{}-72B-\ft{}) consistently improves performance on \datasetAll{}, increasing symbolic success rates by over $10\%$ over their base counterparts. This suggests that data generated by \datagenOurs{} is effective for improving model performance. However, performance improvements mainly come from \datasetBasicNoPrefix{} and \datasetSynNoPrefix{}, while performance on \datasetHandNoPrefix{} remains low, indicating that fine-tuning primarily improves performance on clean and synthetic inputs but does not fully address robustness to out-of-distribution hand-drawn tasks. Additional out-of-distribution results are provided in Appendix~\ref{sec:exp_ood_analysis}.

\input{figs/5_exp/fig_spider_comparative_analysis}

\subsection{Analysis of Model Performance Across Different Dimensions} \label{sec:experiments:comparative}

To enable a more fine-grained analysis of current VLMs, we examine their performance across three dimensions: task categories, difficulty levels, and dataset types. The results are shown in \autoref{fig:spider_compare_all}.

Across task categories (see \autoref{fig:spider_category}), models perform best on Basic Geometry, while Spiral is consistently the most challenging category for all models, as it requires long-horizon, sequential control with tightly coupled steps. Besides, Composite tasks are not uniformly harder than single-transformation tasks (i.e., Rotation, Translation, and Scaling). Overall, these results suggest that models struggle more with tasks requiring high precision and long procedural sequences than with tasks that combine transformations.

Across difficulty levels (see \autoref{fig:spider_difficulty}), all models exhibit a clear monotonic trend, with success rates highest on easy tasks and decreasing on medium and hard tasks, suggesting that the difficulty levels in our datasets capture meaningful differences in task complexity.

Across datasets (see \autoref{fig:spider_dataset}), most models perform slightly worse on \datasetHandNoPrefix{} than on \datasetBasicNoPrefix{}. Since \datasetHandNoPrefix{} is constructed by hand-drawing the images from the same tasks as \datasetBasicNoPrefix{}, this gap indicates that hand-drawn input variations introduce additional challenges.
Performance also drops on \datasetSynNoPrefix{} across most models, consistent with its harder difficulty distribution relative to \datasetBasicNoPrefix{} (see \autoref{fig:dataset_stats}).
Furthermore, while fine-tuning improves performance on both \datasetBasicNoPrefix{} and \datasetSynNoPrefix{}, it yields limited gains on \datasetHandNoPrefix{}. This suggests that fine-tuning on clean synthetic data still faces challenges on out-of-distribution tasks, offering limited improvement in robustness to hand-drawn inputs. We provide additional analysis of out-of-distribution generalization for fine-tuned models in Appendix~\ref{sec:exp_ood_analysis}.

\input{figs/5_exp/fig_failure_distribution}

\subsection{Failure Analysis} \label{sec:exp_failure_analysis}

To examine the limitations of VLMs, we conduct a systematic failure analysis on three representative models: \gptfouro{}, \qwentwovl{}-72B, and \qwentwovl{}-72B-\ft{}. We examine these models using \datasetBasicNoPrefix{}, since it consists of real-world tasks collected from visual programming platforms and therefore best reflects real-world challenges.
We manually review generated images, code, and available explanations to identify root causes of errors, attributing each case to the failure type that contributes most.\footnote{For failure analysis, we apply CoT prompting to \qwentwovl{}-72B to elicit image descriptions and reasoning, yielding a $10.78\%$ success rate on \datasetBasicNoPrefix{}, close to non-CoT's $11.76\%$.} The distribution of failure types is shown in \autoref{fig:failure_and_finetune_analysis}, with definitions and examples in Appendix~\ref{sec:appendix:dataset:labeling:failure_types}.

\paragraph{Models struggle with spatial reasoning.}
We find that all models struggle with spatial reasoning, which is the ability to reason about the spatial relationships among different patterns in the image, such as relative positions, distances, angles, and sizes of patterns. This might be due to the scarcity of training data that captures spatial relationships when training VLMs.

\paragraph{Models struggle with precise visual details.}  
We find that despite correct visual reasoning, models still face difficulties in achieving visual precision during replication. 
For instance, both \gptfouro{} and \qwentwovl{}-72B can often replicate the intended image from a high-level perspective but fail to achieve low-level visual accuracy, such as ignoring tiny details like angles and relative positions.

\paragraph{Models often miss crucial details in images.}
Visual understanding errors remain common between \qwentwovl{}-72B and \qwentwovl{}-72B-\ft{}. During review, we found that they often overlook small but crucial details and describe images using approximate common patterns. For instance, if an image shows a square with a unique cut-off, the models might just describe and draw a regular square, ignoring the specific cut-off.

\paragraph{Fine-tuning improves code-reasoning alignment.}
By comparing \qwentwovl{}-72B and \qwentwovl{}-72B-\ft{}, we observe that fine-tuning increases the success rate from $10.8\%$ to $35.3\%$, mainly by reducing programming errors (from $30.4\%$ to $6.9\%$). Visual understanding and spatial reasoning errors remain largely unchanged, suggesting that fine-tuning primarily improves the alignment between code generation and visual reasoning, rather than visual understanding or spatial reasoning itself.

\input{figs/5_exp/tab_code_quality}

\subsection{Code Length Analysis} \label{sec:length_analysis}

Beyond code correctness, code quality also matters for education-oriented visual programming. In Turtle Graphics tasks, one aspect of code quality is compactness, as unnecessarily long solutions often indicate redundant operations or missed structural patterns such as loops. To capture this aspect, we use the \emph{length ratio} as a proxy for code compactness. For each task, length ratio is computed as the number of lines in the model-generated code divided by the number of lines in the corresponding reference solution, after removing comments and blank lines. Although this metric does not directly capture all aspects of code quality, it provides a lightweight and automated measure of code compactness, a relevant aspect of code quality in Turtle Graphics tasks.

\looseness-1
We analyze length ratio on \datasetBasicNoPrefix{} for three representative models: \gptfouro{}, \qwentwovl{}-72B, and \qwentwovl{}-72B-\ft{}. 
As shown in Table~\ref{tab:code_quality}, \gptfouro{} maintains a relatively stable ratio across successful and failed cases (ratio = 1.25). \qwentwovl{}-72B exhibits a moderate length ratio on successful cases (ratio = 1.52), but its outputs become substantially longer in failure cases (ratio = 4.01), indicating a substantial increase in code verbosity upon failure. In contrast, the fine-tuned \qwentwovl{}-72B-\ft{} maintains a length ratio close to $1.0$ across all cases, suggesting that fine-tuning is associated not only with improved correctness, but also with more consistent and compact generation behavior, particularly in failure cases.

%% file: figs/5_exp/tab_main_results.tex
% !TEX root =  main.tex
%%%%%%%%%%%%%%%%%%%%%%%%%%%%%%%%%%%%%%%%%%%%%%%
%%%%%%%%%%%%%%%%%%%%%%%%%%%%%%%%%%%%%%%%%%%%%%%

%%%%%%%%%%%%%%%%%%%%%%%%%%%%%%%%%%%%%%%%% 

\begin{table*}[t!]
  \centering
  \scalebox{0.76}{
  \begin{tabular}{lrrrrrrrrrr}
    \toprule
    \multicolumn{2}{c}{} & \multicolumn{2}{c}{\datasetBasic{}} & \multicolumn{2}{c}{\datasetHand{}} & \multicolumn{2}{c}{\datasetSyn{}} & \multicolumn{2}{c}{\datasetAll{}} \\
    \cmidrule(lr){3-4} \cmidrule(lr){5-6} \cmidrule(lr){7-8} \cmidrule(lr){9-10}  
    & Size & Sym. (\%) & Emb. (\%) & Sym. (\%) & Emb. (\%) & Sym. (\%) & Emb. (\%) & Sym. (\%) & Emb. (\%) \\

    \midrule

    \emph{Reasoning:} & & & & & & & & & \\
    \quad o3  & - & \textbf{40.20} &	\textbf{44.12} & \underline{8.82}	& \underline{9.80} & \underline{10.18} & \underline{9.21} & \underline{13.73} & \underline{13.61} \\
    \quad o4-mini & - &	\underline{36.27} & \underline{38.24} &	\textbf{28.43}	& \textbf{29.41} & \textbf{10.50} & \textbf{10.50} & \underline{15.92} & \underline{16.28} \\
    \quad GPT-5 (medium) & - & 27.45 & 29.41 & 0.98 & 0.98 & 7.43 & 6.79 & 9.11 & 8.87 \\

    \midrule
    \emph{Non-reasoning ($>$ 30B):} \\
    \quad \gptfouro{} & - & \underline{26.47} & \underline{28.43} & 12.75 & 13.73 & \underline{5.82} & \underline{5.17} & \underline{9.23} & \underline{9.11} \\
    \quad \gptfourv{} & - & 15.69 & 17.65 & 8.82 & 10.78 & 3.23 & 2.75 & 5.47 & 5.59 \\

    \quad \pixtral{}-Large & 124B & 10.78 & 11.76 & \underline{13.73} & \underline{15.69} & 4.68 & 3.88 & 6.56 & 6.32 \\
    \quad \llavaov{} & 72B & 4.90 & 3.92 & 6.86 & 5.88 & 0.97 & 0.97 & 2.19 & 1.94 \\
    \quad InternVL3 & 78B & 12.75 & 13.73 & 7.84 & 9.80 & 3.23 & 2.58 & 4.98 & 4.86 \\
    \quad \molmo{} & 72B & 3.92 & 4.90 & 4.90 & 4.90 & 1.62 & 1.45 & 2.31 & 2.31 \\
    \quad \nvlmd{} & 72B & 0.00 & 0.00 & 0.00 & 0.00 & 0.16 & 0.16 & 0.12 & 0.12 \\
    \quad Qwen2.5-VL & 72B & 15.69 & 18.63 & \textbf{17.65} & \textbf{17.65} & 3.55 & 3.55 & 6.80 & 7.17 \\
    \quad \qwentwovl{} & 72B & 11.76 & 14.71 & 7.84 & 8.82 & 1.45 & 1.62 & 3.52 & 4.13 \\
    \quad \qwentwovl-\ft{} & 72B & \textbf{35.29} & \textbf{39.22} & 6.86 & 6.86 & \textbf{19.06} & \textbf{17.12} & \textbf{19.56} & \textbf{18.59} \\

    \midrule

    \emph{Non-reasoning ($\leq$ 30B):} \\
    \quad Qwen3-VL & 30B & 18.63 & 16.67 & \textbf{13.73} & \textbf{14.71} & 2.91 & 2.26 & 6.20 & 5.59 \\
    \quad \pixtral{} & 12B & 9.80 & 9.80 & 2.94 & 2.94 & 0.97 & 0.97 & 2.31 & 2.31 \\
    \quad \pixtral{}-\ft{} & 12B & \underline{27.45} & \underline{29.41} & \underline{9.80} & \underline{9.80} & \textbf{13.41} & \textbf{12.12} & \textbf{14.70} & \textbf{13.97} \\
    \quad \llavaov{} & 7B & 3.92 & 3.92 & 2.94 & 2.94 & 0.32 & 0.48 & 1.09 & 1.22 \\
    \quad \glmfourv{} & 9B & 0.00 & 0.00 & 0.00 & 0.98 & 0.00 & 0.00 & 0.00 & 0.12 \\
    \quad \molmo{} & 7B & 0.00 & 0.00 & 0.00 & 0.00 & 0.00 & 0.00 & 0.00 & 0.00 \\
    \quad \qwentwovl{} & 7B & 0.98 & 0.98 & 0.00 & 0.00 & 0.00 & 0.00 & 0.12 & 0.12 \\
    \quad \qwentwovl{}-\ft{} & 7B & \textbf{28.43} & \textbf{30.39} & 6.86 & 8.82 & \underline{11.95} & \underline{11.15} & \underline{13.37} & \underline{13.24} \\

    \bottomrule
  \end{tabular}
  }
  \caption{Performance comparison of VLMs on different datasets. We evaluate VLMs using both symbolic comparison (Sym.) and embedding-based comparison (Emb.), with results shown as success rates (\%). Fine-tuned models are denoted by the suffix \ft{}. The best performance within each group is shown in \textbf{bold}, and the second-best is \underline{underlined}. See Appendix~\ref{sec:appendix:experiments:topk} for Pass@K results.}
  \label{fig:main_performance}
\end{table*}

%% file: figs/5_exp/fig_spider_comparative_analysis.tex
% % !TEX root =  main.tex
%%%%%%%%%%%%%%%%%%%%%%%%%%%%%%%%%%%%%%%%%%%%%%%
%%%%%%%%%%%%%%%%%%%%%%%%%%%%%%%%%%%%%%%%%%%%%%%

%%%%%%%%%%%%%%%%%%%%%%%%%%%%%%%%%%%%%%%%% 

\begin{figure*}[ht!]
    \centering
    \includegraphics[width=\textwidth]{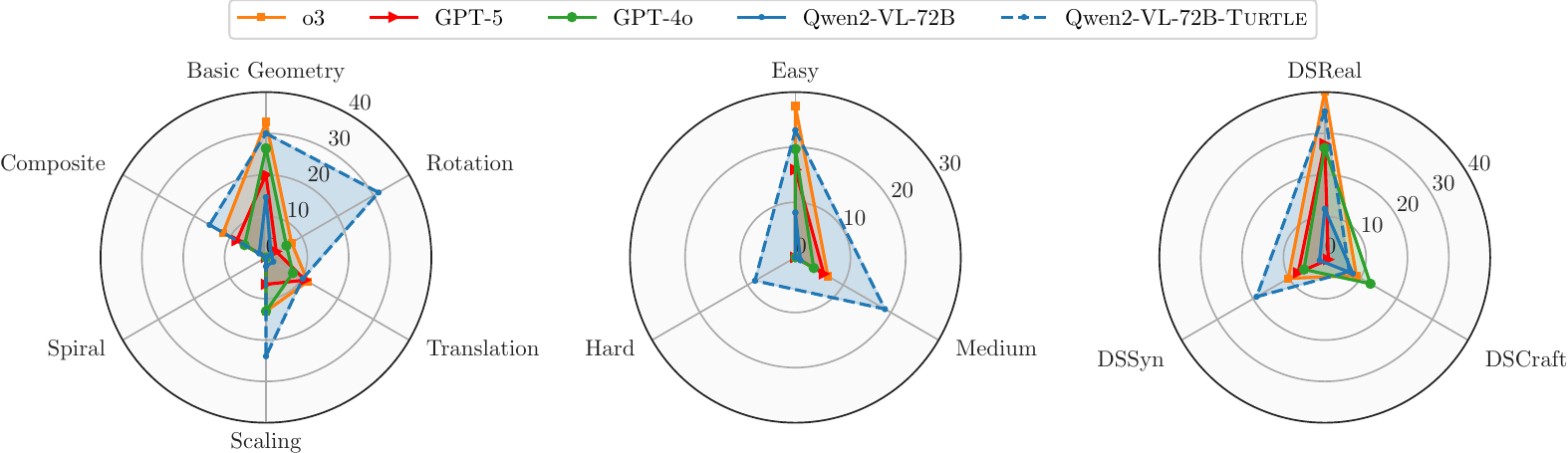}
    \begin{minipage}{0.32\textwidth}
        \begin{subfigure}{0.99\textwidth}
            \centering
            \caption{Task Categories}
            \label{fig:spider_category}
        \end{subfigure}
    \end{minipage}
    \begin{minipage}{0.32\textwidth}
        \begin{subfigure}{0.99\textwidth}
            \centering
            \caption{Difficulty levels}
            \label{fig:spider_difficulty}
        \end{subfigure}
    \end{minipage}
    \begin{minipage}{0.32\textwidth}
        \begin{subfigure}{0.99\textwidth}
            \centering
            \caption{Datasets}
            \label{fig:spider_dataset}
        \end{subfigure}
    \end{minipage}
    \caption{Symbolic success rates (\%) of representative VLMs across task categories, difficulty levels, and datasets. (a) shows performance by task category, where Basic Geometry is the easiest and Spiral is consistently the most challenging. (b) shows performance by difficulty level, decreasing consistently from easy to medium, and to hard. (c) shows performance by dataset, where most models drop relative to \datasetBasicNoPrefix{} on both \datasetHandNoPrefix{} and \datasetSynNoPrefix{}.}
    \label{fig:spider_compare_all}
\end{figure*}

%% file: figs/5_exp/fig_failure_distribution.tex
% !TEX root =  main.tex
%%%%%%%%%%%%%%%%%%%%%%%%%%%%%%%%%%%%%%%%%%%%%%%
%%%%%%%%%%%%%%%%%%%%%%%%%%%%%%%%%%%%%%%%%%%%%%%

%%%%%%%%%%%%%%%%%%%%%%%%%%%%%%%%%%%%%%%%% 

\begin{figure}[ht!]
    \centering
    \begin{subfigure}{0.493\textwidth}
        \centering
        \includegraphics[width=\textwidth]{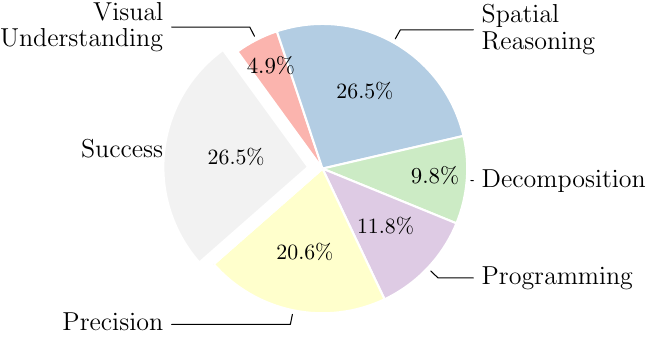}
        \caption{\gptfouro{}}
        \label{fig:failure_dist_gpt4o}
    \end{subfigure}

    \vspace{0.01\textwidth}

    \begin{subfigure}{0.493\textwidth}
        \centering
        \includegraphics[width=\textwidth]{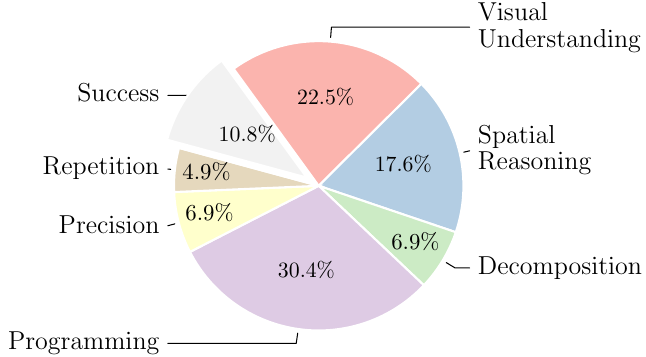}
        \caption{\qwentwovl{}-72B}
        \label{fig:failure_dist_qwen2vl}
    \end{subfigure}

    \vspace{0.01\textwidth}

    \begin{subfigure}{0.493\textwidth}
        \centering
        \includegraphics[width=\textwidth]{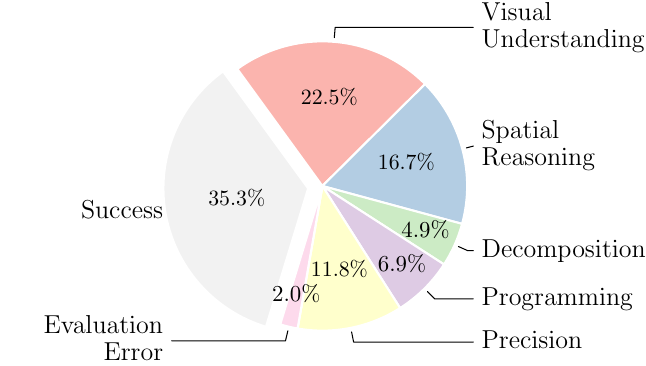}
        \caption{\qwentwovl{}-72B-\ft{}}
        \label{fig:failure_dist_qwen2vl_ft}
    \end{subfigure}
    \caption{Distribution of failure types on \datasetBasicNoPrefix{} dataset for three representative VLMs.}
    \label{fig:failure_and_finetune_analysis}
    \vspace{-1em}
\end{figure}

%% file: figs/5_exp/tab_code_quality.tex
% !TEX root =  main.tex
%%%%%%%%%%%%%%%%%%%%%%%%%%%%%%%%%%%%%%%%%%%%%%%
%%%%%%%%%%%%%%%%%%%%%%%%%%%%%%%%%%%%%%%%%%%%%%%

%%%%%%%%%%%%%%%%%%%%%%%%%%%%%%%%%%%%%%%%%

\begin{table*}[t!]
  \centering
  \scalebox{0.9}{
  \begin{tabular}{lrrr}
    \toprule
    Model & Length Ratio (Success-Only) & Length Ratio (Failure-Only) & Length Ratio (All) \\
    \midrule
    \gptfouro{} & $1.25 \pm 0.08$ & $1.25 \pm 0.05$ & $1.25 \pm 0.04$ \\
    \qwentwovl{}-72B & $1.52 \pm 0.16$ & $4.01 \pm 1.27$ & $3.74 \pm 1.14$ \\
    \qwentwovl{}-72B-\ft{} & $1.02 \pm 0.05$ & $1.03 \pm 0.06$ & $1.03 \pm 0.04$ \\
    \bottomrule
  \end{tabular}
  }
  \caption{Length ratios on \datasetBasicNoPrefix{} dataset for three models. We report length ratio for three cases: \textit{Success-Only} (computed over successful codes), \textit{Failure-Only} (computed over failed codes), and \textit{All} (computed over all codes). Length ratios close to $1.0$ indicate generated code lengths similar to the reference solution code, while larger ratios suggest more verbose or potentially redundant codes. Length ratio values are reported as mean $\pm$ standard error.}
  \label{tab:code_quality}
  \vspace{-0.5em}
\end{table*}

%% file: 6_conclusions.tex
% !TEX root =  main.tex
%%%%%%%%%%%%%%%%%%%%%%%%%%%%%%%%%%%%%%%%%%%%%%%
%%%%%%%%%%%%%%%%%%%%%%%%%%%%%%%%%%%%%%%%%%%%%%%

%%%%%%%%%%%%%%%%%%%%%%%%%%%%%%%%%%%%%%%%% 

\section{Concluding Discussions}
\label{sec.conclusion}

\looseness-1
In this paper, we introduced \benchmark{}, a benchmark of 823 tasks for evaluating vision-language models on education-oriented visual programming in Turtle Graphics. Our evaluation showed that current VLMs struggled substantially, even on real-world tasks designed for students in grades 3-6, highlighting a clear gap between current model capabilities and the reliability required for classroom deployment. To improve performance, we proposed \datagenOurs{}, a data generation technique that synthesizes high-quality image-code pairs from a small seed set; fine-tuning \qwentwovl{}-72B on the synthesized data improved performance on real-world tasks by over $20\%$ compared to the base model. Our analysis showed that \gptfouro{} most often failed due to limitations in spatial reasoning and faithful visual replication, while fine-tuning primarily reduced programming errors by aligning code implementation with visual reasoning rather than improving visual understanding or spatial reasoning.

%% file: 7_limitations.tex
% !TEX root =  main.tex
%%%%%%%%%%%%%%%%%%%%%%%%%%%%%%%%%%%%%%%%%%%%%%%
%%%%%%%%%%%%%%%%%%%%%%%%%%%%%%%%%%%%%%%%%%%%%%%

%%%%%%%%%%%%%%%%%%%%%%%%%%%%%%%%%%%%%%%%% 

\section*{Limitations} \label{sec:limitations}
\looseness-1
We discuss some limitations of our work and propose ideas for addressing them in the future. 
First, our analysis of code quality focuses on code compactness as measured by length ratio, which does not capture all aspects of code quality. Future work could explore additional metrics for code quality, such as code readability and pedagogical alignment.
Second, our evaluation framework includes a normalization step that makes drawing comparison invariant to size, translation, and line width, which may discard meaningful geometric variations that are essential in inverse graphics tasks. Future work could explore evaluation methods that preserve these geometric properties.
Third, we did not provide a systematic ablation or comparison with other data synthesis techniques. Future work could conduct controlled ablations and method comparisons to better understand the contribution of each stage in \datagenOurs{} and how different synthesis choices affect fine-tuning performance.
Finally, although fine-tuning with our data generation technique improves overall performance, fine-tuned models still struggle on hand-drawn inputs (e.g., \datasetHandNoPrefix{}). This may be because our synthesized training data focuses on clean renderings and lacks sufficient variation that matches the noise and distortions in human drawings. Future work could improve robustness by augmenting training with hand-drawn style perturbations, such as random noise injection and transformations that simulate common hand-drawn artifacts.

%% file: 8_acknowledgments.tex
% !TEX root =  main.tex
%%%%%%%%%%%%%%%%%%%%%%%%%%%%%%%%%%%%%%%%%%%%%%%
%%%%%%%%%%%%%%%%%%%%%%%%%%%%%%%%%%%%%%%%%%%%%%%

%%%%%%%%%%%%%%%%%%%%%%%%%%%%%%%%%%%%%%%%% 

\section*{Acknowledgments}
Funded/Co-funded by the European Union (ERC, TOPS, 101039090). Views and opinions expressed are however those of the author(s) only and do not necessarily reflect those of the European Union or the European Research Council. Neither the European Union nor the granting authority can be held responsible for them.

%% file: 9_appendix.tex
% !TEX root =  main.tex
%%%%%%%%%%%%%%%%%%%%%%%%%%%%%%%%%%%%%%%%%%%%%%%
%%%%%%%%%%%%%%%%%%%%%%%%%%%%%%%%%%%%%%%%%%%%%%%
\appendix

\allowdisplaybreaks
\input{appendix_table-of-contents}

\input{appendix_impact} 
\input{appendix_dataset}

\input{appendix_experiments}

\input{appendix_evaluation_reliability}

\input{appendix_implementation}
\input{appendix_case_study}
\input{appendix_prompt}

%%%%%%%%%%%%%%%%%%%%%%%%%%%%%%%%%%%%%%%%%%%%%%%%%%%%%%%%%%%%

%% file: appendix_table-of-contents.tex
% !TEX root =  main.tex
%%%%%%%%%%%%%%%%%%%%%%%%%%%%%%%%%%%%%%%%%%%%%%%
%%%%%%%%%%%%%%%%%%%%%%%%%%%%%%%%%%%%%%%%%%%%%%%

\section{Table of Contents}
\label{sec:appendix:overview}

In this section, we briefly summarize the contents of the paper's appendices.

\begin{itemize}
    \item \autoref{sec:appendix:impact} provides details about broader impacts and declaration of LLM usage.
    \item \autoref{sec:appendix:dataset} provides more details about the dataset generation and labeling process.
    \item \autoref{sec:appendix:additional_experiments} provides additional experiments and analysis.
    \item \autoref{sec:appendix:evaluation_reliability} provides more details about the reliability of the evaluation framework.
    \item \autoref{sec:appendix:implementation_details} provides more details about the implementation of the benchmark and experiments.
    \item \autoref{sec:appendix:case_study} provides a case study of model failures.
    \item \autoref{sec:appendix:prompt} provides more details about the prompts used for fine-tuning and evaluation.
\end{itemize}

%% file: appendix_impact.tex
% !TEX root =  main.tex
%%%%%%%%%%%%%%%%%%%%%%%%%%%%%%%%%%%%%%%%%%%%%%%
%%%%%%%%%%%%%%%%%%%%%%%%%%%%%%%%%%%%%%%%%%%%%%%

%%%%%%%%%%%%%%%%%%%%%%%%%%%%%%%%%%%%%%%%% 

\section{Broader Impacts and Declaration of LLM Usage} \label{sec:appendix:impact}

\subsection{Broader Impacts} \label{sec:appendix:impact:broader}
This work introduces a benchmark for evaluating the performance of existing vision-language models (VLMs) in solving visual programming tasks in the Turtle Graphics domain. In addition, we propose a novel data generation technique, \datagenOurs{}, for generating large-scale synthetic data to train VLMs.

Our benchmark and data generation technique have several positive broader impacts: (i) our work can facilitate programming education, especially in K-12 settings where Turtle Graphics is commonly used; (ii) our work can help track and improve VLMs' ability to understand and generate Turtle Graphics code, making it easier for both beginners and experts to create vector graphics and complex geometric art; and (iii) our data generation technique can advance synthetic data generation in fields where real data is scarce or difficult to collect, potentially helping researchers and developers build better models in a variety of domains beyond Turtle Graphics.

However, it is essential to acknowledge the potential risks associated with synthetic data generation. For instance, our framework could be misused to generate images with political or sensitive content. We emphasize the need for careful oversight and ethical considerations in the application of our framework to ensure that it is used responsibly and for the benefit of society.

\subsection{Declaration of LLM Usage} \label{sec:appendix:impact:llm}
We declare that large language models (LLMs) were used only to assist with writing and formatting the manuscript. LLMs were not involved in the design of experiments, analysis of results, or any other important aspects in this paper.

%% file: appendix_dataset.tex
% !TEX root =  main.tex
%%%%%%%%%%%%%%%%%%%%%%%%%%%%%%%%%%%%%%%%%%%%%%%
%%%%%%%%%%%%%%%%%%%%%%%%%%%%%%%%%%%%%%%%%%%%%%%

%%%%%%%%%%%%%%%%%%%%%%%%%%%%%%%%%%%%%%%%% 

\section{Additional Details About the Dataset Generation and Labeling Process} \label{sec:appendix:dataset}

In this section, we provide detailed information and generation processes for the datasets used in this paper. We then describe our labeling process for task categories, difficulty levels, and failure types.

\input{appendix-figs/fig_dataset_summary}

\subsection{Dataset License} \label{sec:appendix:dataset:license}

The \datasetBasic{} dataset is collected from the visual programming platform \platformAll{} and is licensed under CC BY-NC 4.0.\footnote{\url{https://xlogo.inf.ethz.ch/terms.html}}. 

The datasets \datasetTrain{} and \datasetSyn{} were generated with vision-language models. The models used and their licenses are listed below:
\begin{itemize}
  \item \emph{Qwen2-VL-72B-Instruct} uses Qwen License Agreement.\footnote{\url{https://huggingface.co/Qwen/Qwen2-VL-72B-Instruct/blob/main/LICENSE}}  
  \item \emph{Pixtral-Large} uses Mistral Research License (MRL) for research/educational use.\footnote{\url{https://mistral.ai/news/pixtral-large}}  
  \item \emph{Llama 3.1-70B-Instruct} uses Llama 3.1 Community License Agreement.\footnote{\url{https://github.com/meta-llama/llama-models/blob/main/models/llama3_1/LICENSE}}  
\end{itemize}

\subsection{Dataset Generation Process} \label{sec:appendix:dataset:generation}

In this subsection, we provide more details about the generation process for different datasets used in this paper, including the evaluation datasets \datasetBasic{}, \datasetHand{}, and \datasetSyn{}, and the training dataset \datasetTrain{}. \autoref{fig:appendix:dataset_summary} gives a summary of the datasets used in this paper.

\paragraph{\datasetBasic{}.} This dataset is curated from the visual programming platform \platformAll{}. The tasks in this platform are carefully designed by domain experts and have been used by tens of thousands of students for learning programming every year~\citep{DBLP:journals/eatcs/Staub21}. The involvement of domain experts ensures that this dataset includes a diverse range of high-quality tasks, reflecting real-world learning scenarios.

\paragraph{\datasetHand{}.}
%
\input{appendix-figs/fig_example_reference_handdrawn}

This dataset is generated by manually drawing the task images from \datasetBasic{} using a drawing tool. Specifically, we use each task image in \datasetBasic{} as the reference image and ask a human without any prior knowledge of Turtle Graphics or professional drawing skills to manually draw the task image using a digital drawing tool (i.e., an iPad). 
Finally, we replace each task image in \datasetBasic{} with the corresponding hand-drawn image, resulting in the dataset \datasetHand{}. \autoref{fig:appendix:examples:handdrawn} shows some examples of the reference images and the corresponding hand-drawn images in the dataset \datasetHand{}. This dataset can be used to evaluate the generalization capabilities of the model to real-world drawing tasks.

\paragraph{\datasetSyn{}.} The dataset \datasetSyn{} is generated by our proposed data synthesis framework \datagenOurs{}. 
Specifically, we use \datasetBasic{} (Size $102$) as the seed dataset for \datagenOurs{}. 
We run our \datagenOurs{} for $3$ iterations; in each iteration, we keep the top $30\%$ of the generated samples for the next iteration of \datagenOurs{}, resulting in a synthetic dataset with $8,214$ image-code pairs. To further ensure the quality, we manually select from this dataset using the rubrics defined in the elite selection stage and make a binary decision on whether to keep an image-code pair. When making the decision, we adopt the rubrics used in the elite selection stage of \datagenOurs{}, including (i) geometric structure and symmetry, (ii) visual appeal, clarity, and simplicity, (iii) structural coherence, (iv) alignment and positioning, (v) educational value and solvability, and (vi) color usage and necessity. When evaluating the quality of each sample, we make a binary decision (i.e., ``good'' or ``bad'') for each dimension, and only keep the sample if all dimensions are evaluated as ``good''. After this process, we obtain the final dataset \datasetSyn{} with $619$ high-quality samples. Note that we do not apply the CoT labeling for generating \datasetSyn{} since this dataset is not used for training. The implementation details of \datagenOurs{} are provided in Appendix~\ref{sec:appendix:datagen_details}.

\paragraph{\datasetAll{}.} The dataset \datasetAll{} is a union of \datasetBasic{}, \datasetHand{}, and \datasetSyn{} datasets, including a total of $102 + 102 + 619 = 823$ samples. 

\paragraph{\datasetTrain{}.} The dataset \datasetTrain{} is a large-scale training dataset containing $738,126$ samples. This dataset is generated using a seed dataset with only $10$ seed examples. These seed examples are provided in \autoref{fig:examples.dataset_train.seed} and are selected based on the following three principles:
\begin{itemize}
\item Minimal manual effort: The set should be as small as possible to reduce manual effort.
\item Simplicity: The pairs should be easy to design and understand.
\item Conceptual diversity: The set should cover a broad range of geometric transformation concepts.
\end{itemize}
By following these principles, we arrived at a set of 10 pairs that is both minimal and simple, while remaining diverse. These pairs capture a range of geometric transformation types observed in our domain, including:
\begin{itemize}
\item Adding or removing edges (e.g., transforming a triangle into a square or vice versa)
\item Rotating shapes (e.g., turning a square into a diamond)
\item Translating shapes (e.g., placing two squares side by side)
\item Scaling (e.g., comparing a large square with a smaller one)
\item Changing edge color (e.g., a square with red edges)
\item Changing fill color (e.g., a red-filled square)
\item Combining shapes (e.g., combining a square and a triangle)
\end{itemize}
After preparing the seed examples, we generate the training dataset using the same settings as generating \datasetSyn{}, except that (i) we use a different seed dataset with only $10$ manually designed examples; (ii) we iterate $5$ times with \datagenOurs{}; (iii) we use $k=70\%$ for the elite selection stage; and (iv) we apply the CoT labeling after iterating $5$ times. Examples of generated samples and the generated CoT label are shown in \autoref{fig:examples.dataset_train}.

\input{appendix-figs/fig_seed_images}

\subsection{Labeling Process} \label{sec:appendix:dataset:labeling}

We describe the dataset labeling process for the task categories and difficulty levels in the evaluation datasets.

\paragraph{Labeling process of task categories.} \label{sec:appendix:dataset:labeling:categories}
We identify $6$ different geometric categories of task images based on the visual patterns and transformations in the evaluation datasets. The definitions of these categories are as follows:
\begin{itemize}
    \item \categoryBasic{}: simple and basic shapes like squares, circles, triangles, and lines without complex patterns or arrangements. Tasks in this category require understanding of basic geometry.
    \item \categoryRotation{}: patterns formed by rotating basic geometric shapes around a central point to create symmetrical designs, such as a spirograph. These tasks require reasoning about rotation angles and the numbers of repetitions. 
    \item \categoryTranslation{}: patterns formed by translating a basic pattern to different positions, forming tiling or grid structures. Tasks in this category require reasoning about translation distance and the numbers of basic patterns.
    \item \categoryScaling{}: patterns formed by scaling basic geometric shapes, creating nested or expanding structures. Tasks in this category require reasoning about scaling factors and the numbers of repetitions.
    \item \categorySpiral{}: sequential shapes arranged in spiraling paths, creating dynamic patterns with radial symmetry, like an Archimedean spiral. Tasks in this category require reasoning about the spiral pattern, the numbers of repetitions, the degrees of rotation, and scaling factors.
    \item \categoryComposite{}: complex arrangements combining different transformations (scaling, rotation, spiral, and translation) with varied shapes or colors.
\end{itemize}

Given above definitions, we manually label each task-code pair in the dataset \datasetAll{} into one of the $6$ categories. Note that an image may involve multiple transformations, we categorize each image based on its predominant geometric characteristic.

\input{appendix-figs/fig_example_diff}

\paragraph{Labeling process of difficulty levels.} \label{sec:appendix:dataset:labeling:difficulty}
We assigned a difficulty level for each task-code pair in the dataset \datasetAll{}. The assigned difficulty level is based on the complexity of the visual patterns, and difficulty of drawing the image using Turtle Graphics. We define $3$ difficulty levels as follows:
\begin{itemize}
  \item \difficultyEasy{}: basic patterns consisting of single geometric shapes like squares, circles, or lines without complex combinations. Tasks in this category require entry-level visual understanding of geometry, basic math reasoning about transformation parameters (e.g., length, angles), and fundamental programming skills to implement simple geometric shapes.

  \item \difficultyMedium{}: patterns involving combinations of basic shapes with one or more transformations. Tasks in this category require the ability to decompose patterns into simpler components, understand relationships between different transformations, and programming skills to implement multiple transformation steps.

  \item \difficultyHard{}: sophisticated patterns featuring multiple complex transformations. Tasks in this category require high-level visual understanding, reasoning about spatial and temporal relationships between components, mathematical reasoning about transformation parameters (e.g., length, angles, scaling factors), and programming skills to convert complex reasoning into executable programs.
\end{itemize}
Given above definitions, we manually label each image in the dataset \datasetAll{} into one of the $3$ difficulty levels.

\paragraph{Labeling process of failure types.} \label{sec:appendix:dataset:labeling:failure_types}
We identify 7 different failure types based on the failure cases of different models on the \datasetBasic{} dataset. The definitions of these failure types are as follows:
\begin{itemize}
    \item \emph{Visual understanding}: This involves misinterpreting the image's overall design, layout, or composition, leading to code generation that does not match the target image. For instance, the image shows a square, but the model describes it as a circle and generates the code for a circle.

    \item \emph{Decomposition}: This involves errors in decomposing the image into feasible drawing steps or errors in decomposing a complex pattern into its constituent parts (shapes, patterns, elements).
    For example, when a complex pattern results from repeated rotations of a simple shape. The model may fail to decompose it into the base shape and its transformations, instead treating the entire pattern as an indivisible unit. This leads to either an attempt to generate code that represents the entire complex pattern directly or a failure to plan a feasible sequence of steps to generate the pattern.

    \item \emph{Spatial reasoning}: This involves errors in understanding relative positions, distances, angles, and sizes of patterns within the image. For instance, the model misinterprets the relative positioning of two shapes, placing one above the other when they are actually side-by-side.

    \item \emph{Programming}: This involves the model having correct visual understanding and reasoning but the code implementation is not consistent with the visual reasoning results or the code implementation contains syntax or logical errors. For instance, the model understands correctly that the image shows a square, but implements the code for a circle instead.

    \item \emph{Visual precision}: This involves the model having correct visual understanding and reasoning, but failing to achieve very precise details during the code implementation.
    For instance, the model generates code that captures the overall structure but deviates in specifics, such as lines being slightly too long, angles that are a few degrees off.

    \item \emph{Repetition}: This involves unnecessary or incorrect repetition of code blocks. For instance, the model keeps generating the same redundant code repeatedly without stopping.

    \item \emph{Evaluation error}: This is due to the evaluation framework's incorrect evaluation results that are not consistent with the manual evaluation results. For instance, the symbolic comparison incorrectly identifies the generated image as a \emph{success}, but it's actually a \emph{fail} from the manual evaluation.
    
\end{itemize}

%% file: appendix-figs/fig_dataset_summary.tex
% !TEX root =  main.tex
%%%%%%%%%%%%%%%%%%%%%%%%%%%%%%%%%%%%%%%
%%%%%%%%%%%%%%%%%%%%%%%%%%%%%%%%%%%%%%%

\begin{table*}[ht!]
\centering
  \scalebox{1}{
  \begin{tabular}{lrrrr}
  \toprule
  Dataset & \# Samples & Purpose & Seed Dataset & Seed Size \\
  \midrule
  \datasetAll{}  & 823 & Evaluation & - & - \\
  \datasetBasic{} & 102 & Evaluation & - & - \\
  \datasetHand{} & 102 & Evaluation & \datasetBasic{} & 102 \\
  \datasetSyn{} & 619 & Evaluation& \datasetBasic{} & 102 \\
  \datasetTrain{} & 738,126 & Train \& Validation & Manually curated & 10 \\
  \bottomrule
  \end{tabular}
  }
\caption{A summary of the datasets used in this paper.}
\label{fig:appendix:dataset_summary}
\end{table*}

%% file: appendix-figs/fig_example_reference_handdrawn.tex
% !TEX root =  main.tex
%%%%%%%%%%%%%%%%%%%%%%%%%%%%%%%%%%%%%%%
%%%%%%%%%%%%%%%%%%%%%%%%%%%%%%%%%%%%%%%

\begin{figure}[h]
    \centering
      \begin{subfigure}[b]{0.3\linewidth}
        \centering
        \includegraphics[width=0.45\textwidth]{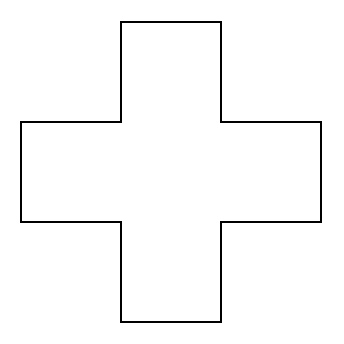}
        \includegraphics[width=0.46\textwidth]{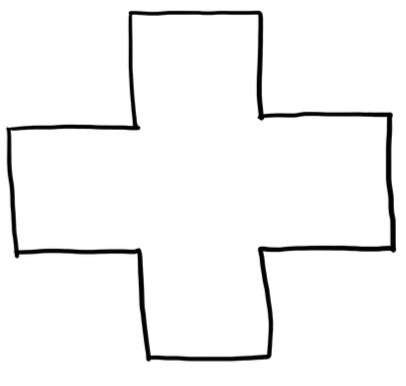}
        \caption{\categoryBasic}
        \label{fig:appendix:examples:handdrawn:basic_geometry}
      \end{subfigure}
      \hspace{0.01\linewidth}
      \begin{subfigure}[b]{0.3\linewidth}
        \centering
        \includegraphics[width=0.45\textwidth]{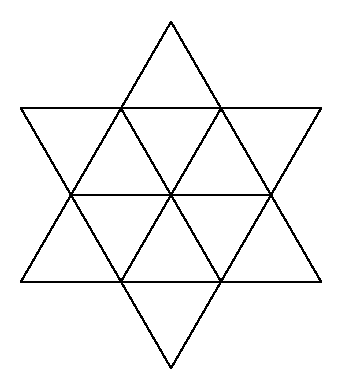}
        \includegraphics[width=0.46\textwidth]{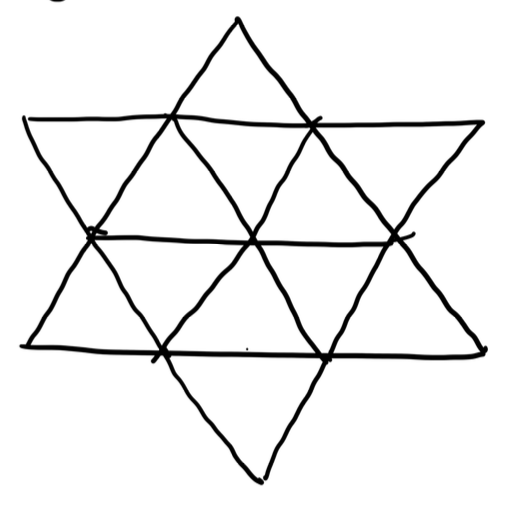}
        \caption{\categoryRotation}
        \label{fig:appendix:examples:handdrawn:rotation}
      \end{subfigure}
      \hspace{0.01\linewidth}
      \begin{subfigure}[b]{0.3\linewidth}
        \centering
        \includegraphics[width=0.45\textwidth]{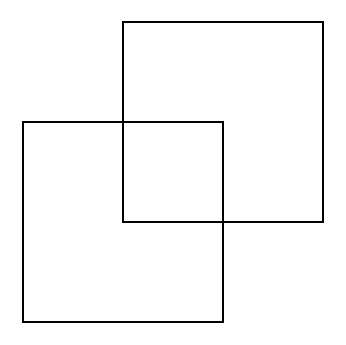}
        \includegraphics[width=0.46\textwidth]{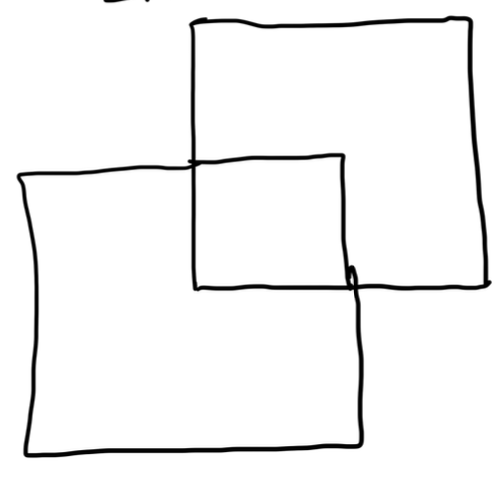}
        \caption{\categoryTranslation}
        \label{fig:appendix:examples:handdrawn:translation}
      \end{subfigure}
   
      \hspace{0.01\linewidth}

      \hspace{0.01\linewidth}
      \begin{subfigure}[b]{0.3\linewidth}
        \centering
        \includegraphics[width=0.45\textwidth]{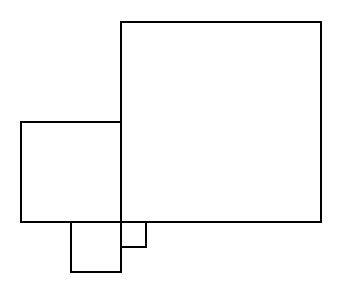}
        \includegraphics[width=0.46\textwidth]{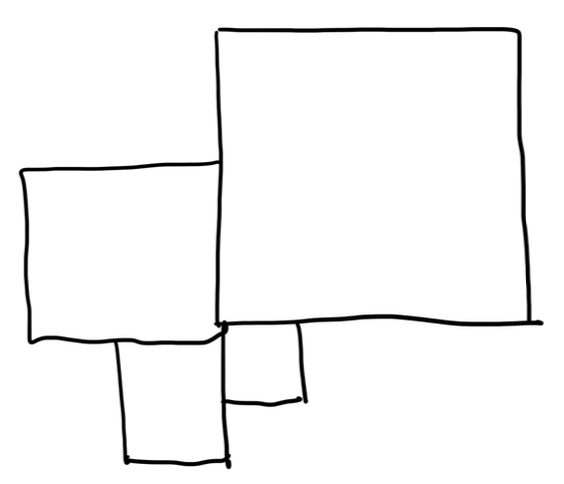}
        \caption{\categoryScaling}
        \label{fig:appendix:examples:handdrawn:scaling}
      \end{subfigure}
      \hspace{0.01\linewidth}
      \begin{subfigure}[b]{0.3\linewidth}
        \centering
        \includegraphics[width=0.45\textwidth]{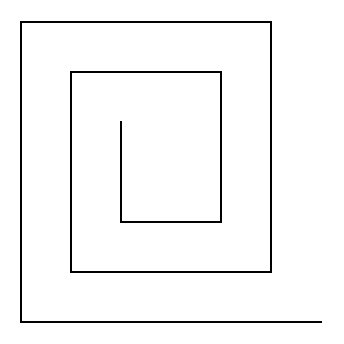}
        \includegraphics[width=0.46\textwidth]{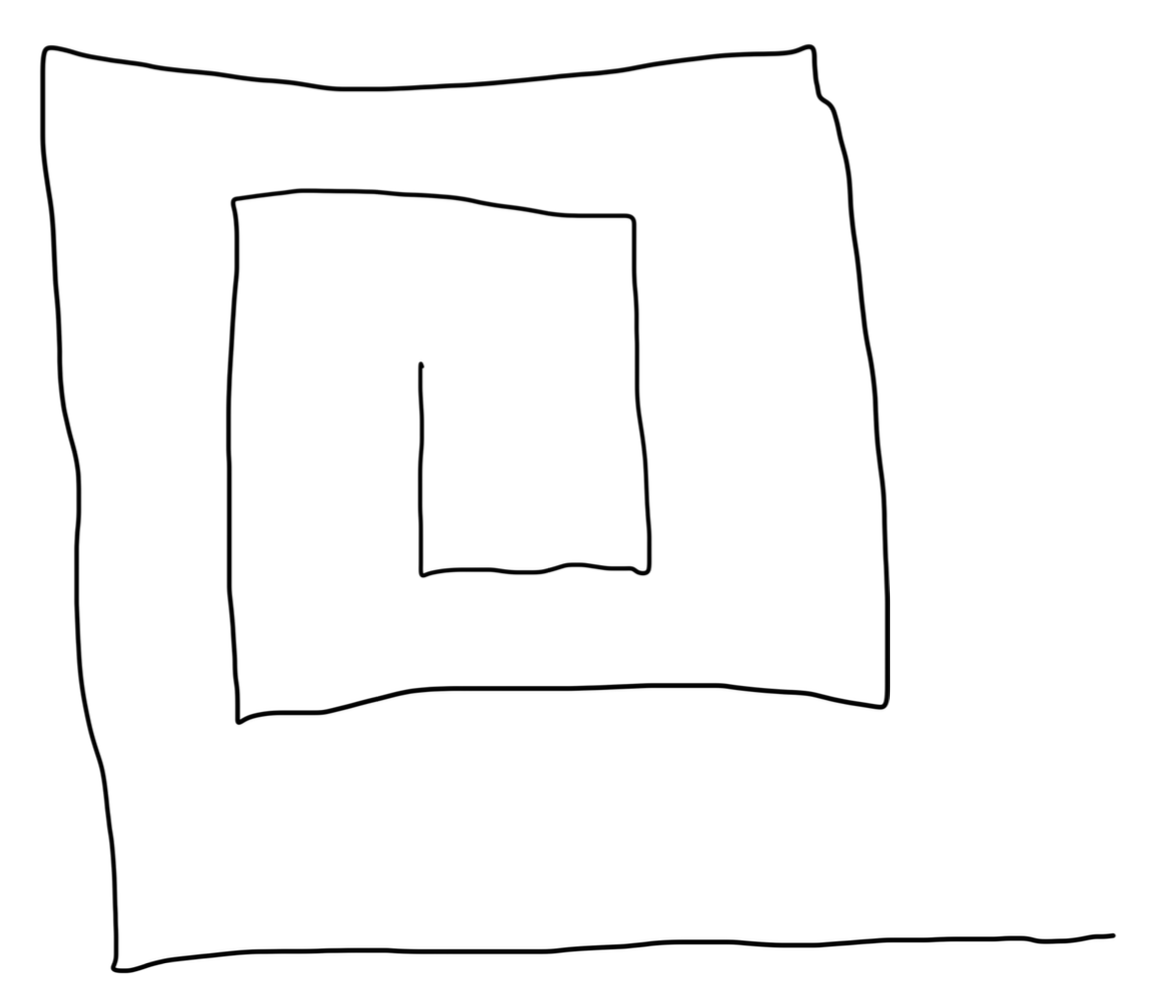}
        \caption{\categorySpiral}
        \label{fig:appendix:examples:handdrawn:spiral}
      \end{subfigure}
      \hspace{0.01\linewidth}
      \begin{subfigure}[b]{0.3\linewidth}
        \centering
        \includegraphics[width=0.45\textwidth]{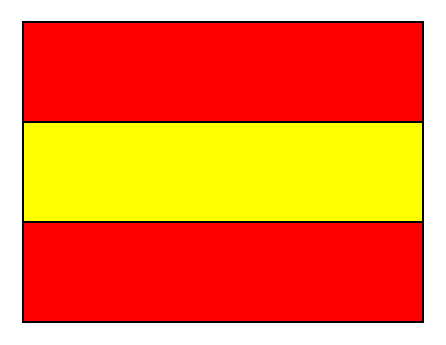}
        \includegraphics[width=0.46\textwidth]{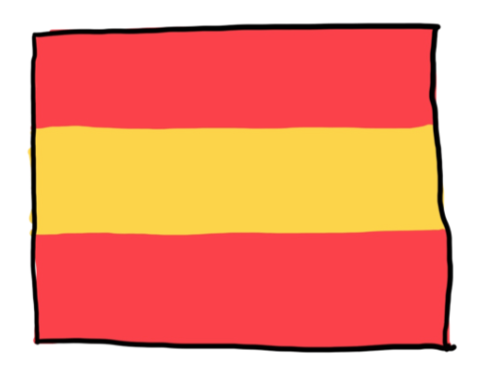}
        \caption{\categoryComposite}
        \label{fig:appendix:examples:handdrawn:composite}
      \end{subfigure}
 
      \caption{Examples of the reference images and the corresponding hand-drawn images in the dataset \datasetHand{}. The reference images are shown on the left, and the corresponding hand-drawn images are shown on the right. One example is shown for each task category.}
      \label{fig:appendix:examples:handdrawn}
\end{figure}

%% file: appendix-figs/fig_seed_images.tex
% !TEX root =  main.tex
%%%%%%%%%%%%%%%%%%%%%%%%%%%%%%%%%%%%%%%
%%%%%%%%%%%%%%%%%%%%%%%%%%%%%%%%%%%%%%%

\begin{figure*}[p!]
    \centering
    \begin{minipage}[b]{0.4\textwidth}
        \begin{subfigure}{\textwidth}
            \centering
            \begin{tabular}{ccccc}
              \includegraphics[width=0.12\textwidth]{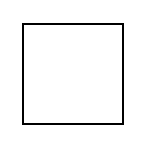} &
              \includegraphics[width=0.12\textwidth]{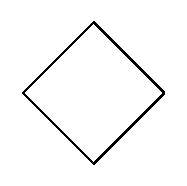} &
              \includegraphics[width=0.12\textwidth]{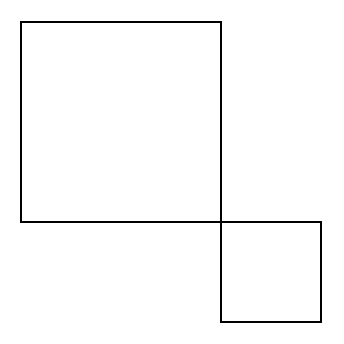} &
              \includegraphics[width=0.12\textwidth]{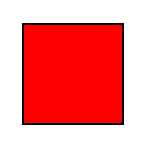} &
              \includegraphics[width=0.10\textwidth]{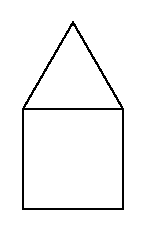} \\
              \\
              \includegraphics[width=0.12\textwidth]{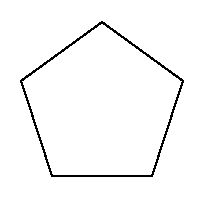} &
              \includegraphics[width=0.12\textwidth]{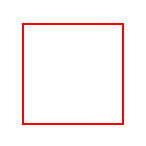} &
              \includegraphics[width=0.12\textwidth]{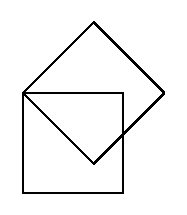} &
              \includegraphics[width=0.13\textwidth]{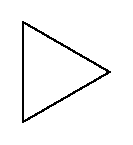} &
              \includegraphics[width=0.12\textwidth]{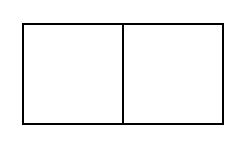} \\
            \end{tabular}
            \caption{All the $10$ task images in the seed dataset.}
            \label{fig:examples.dataset_train.seed}
        \end{subfigure}

        \vspace{0.1\linewidth}

        \begin{subfigure}{\textwidth}
            \centering
            \begin{tabular}{ccccc} % 5 columns
                \includegraphics[width=0.12\textwidth]{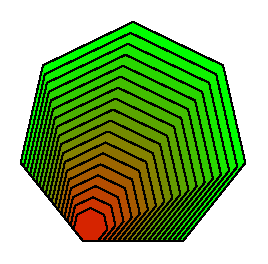} &
                \includegraphics[width=0.12\textwidth]{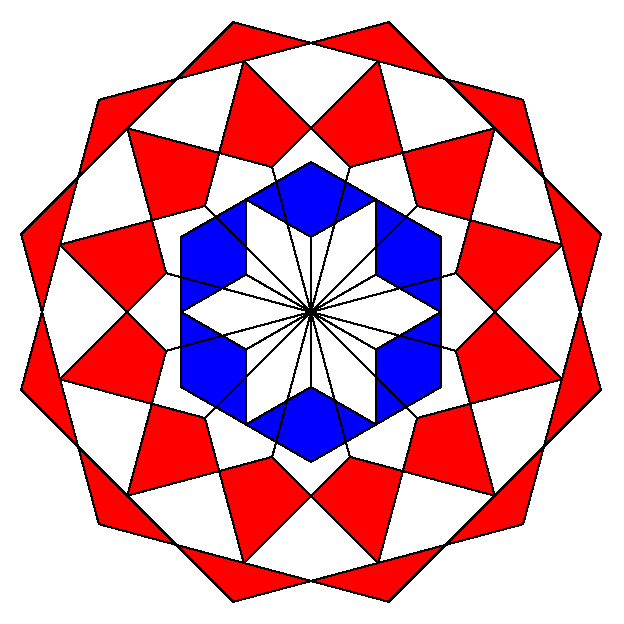} &
                \includegraphics[width=0.12\textwidth]{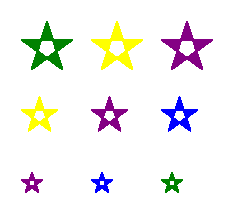} &
                \includegraphics[width=0.12\textwidth]{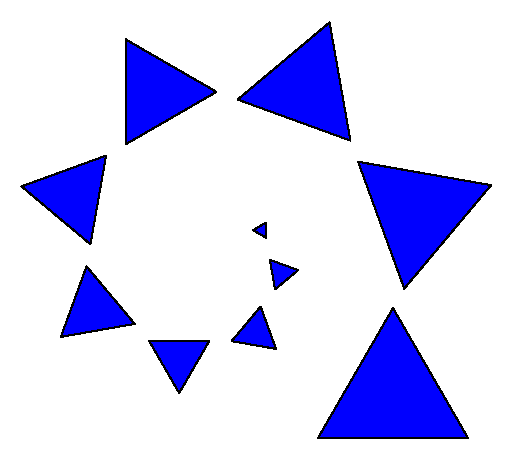} &
                \includegraphics[width=0.12\textwidth]{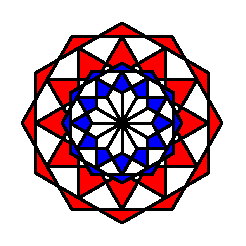} \\
                % \hline
                \\
                \includegraphics[width=0.12\textwidth]{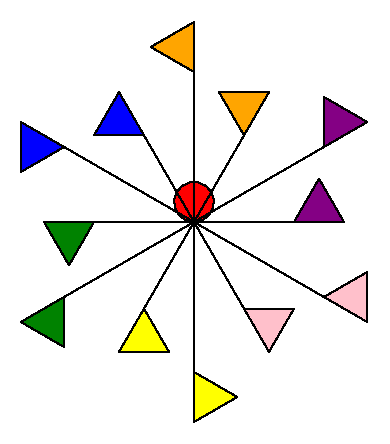} &
                \includegraphics[width=0.12\textwidth]{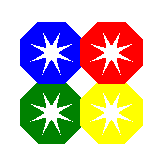} &
                \includegraphics[width=0.12\textwidth]{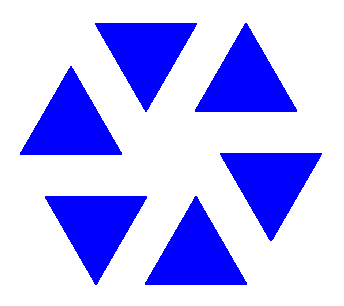} &
                \includegraphics[width=0.12\textwidth]{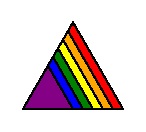} &
                \includegraphics[width=0.12\textwidth]{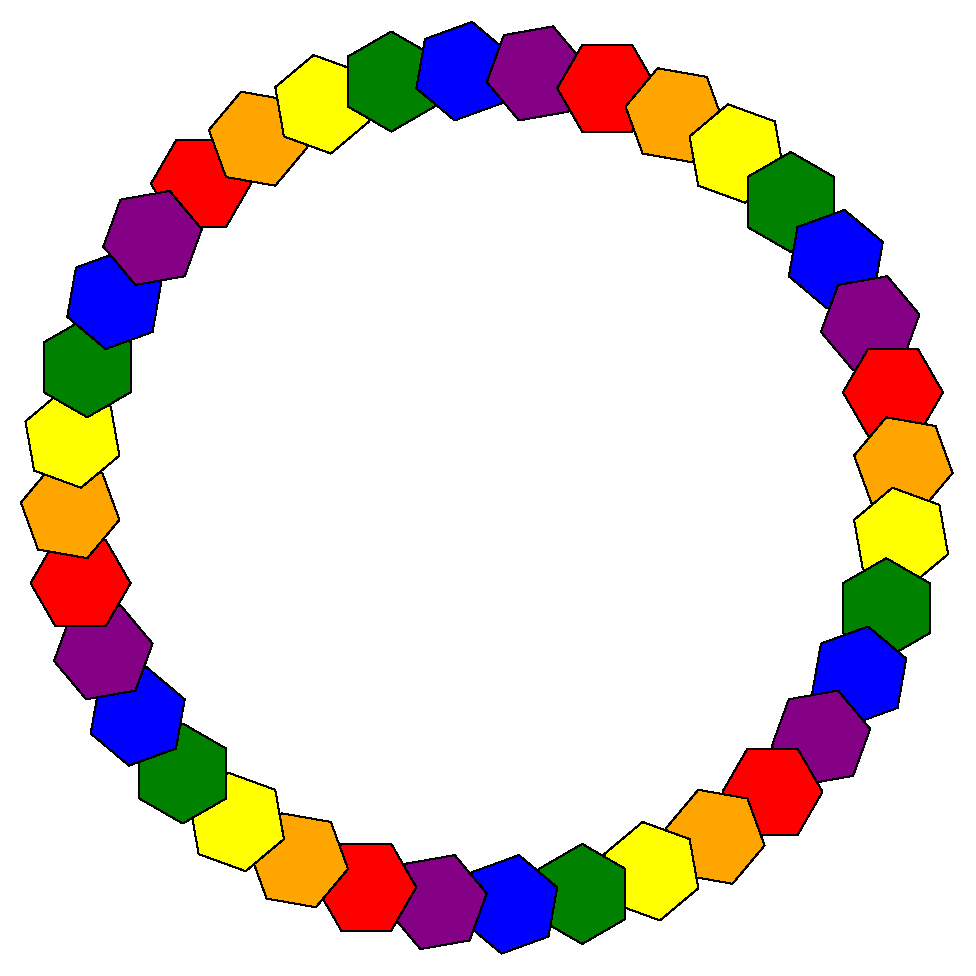} \\
                % \hline
                \\
                \includegraphics[width=0.12\textwidth]{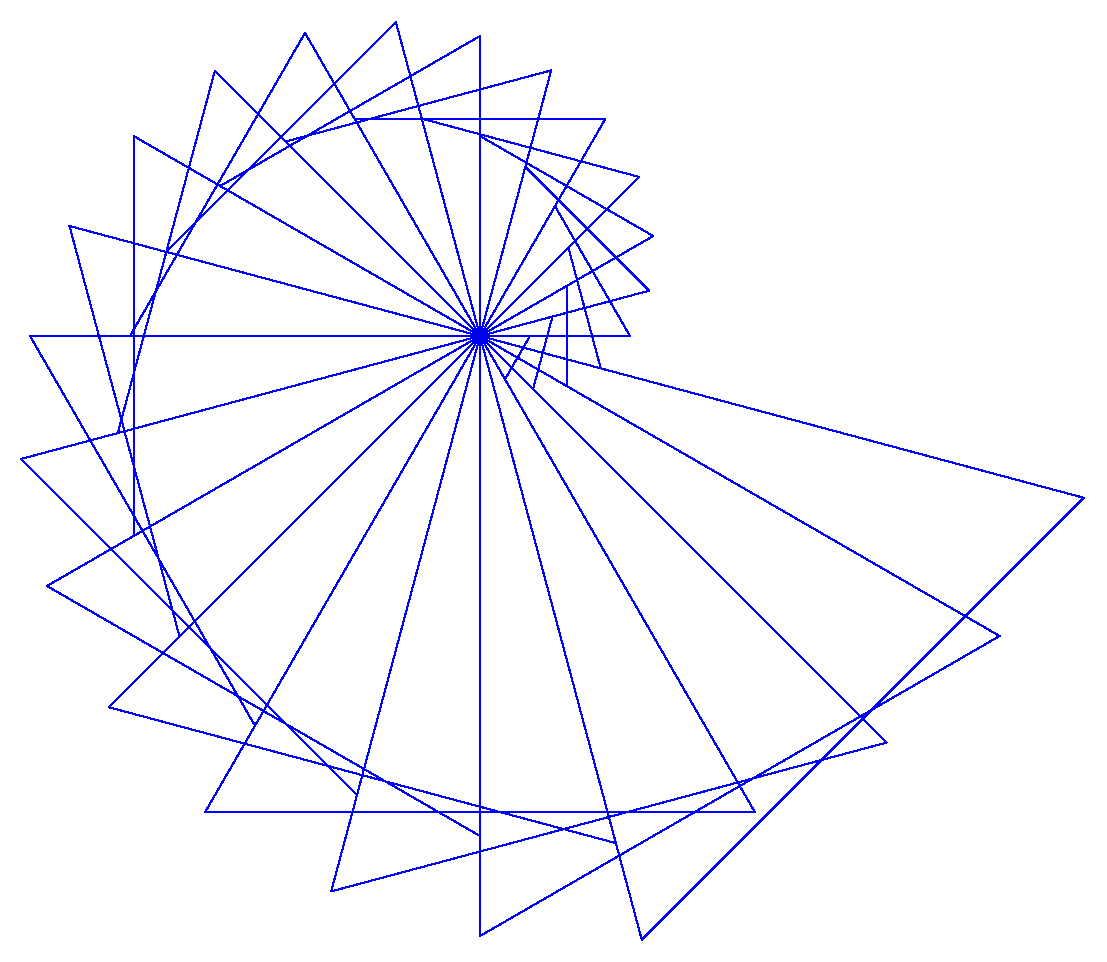} &
                \includegraphics[width=0.12\textwidth]{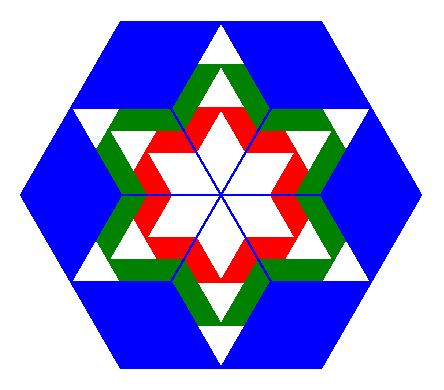} &
                \includegraphics[width=0.12\textwidth]{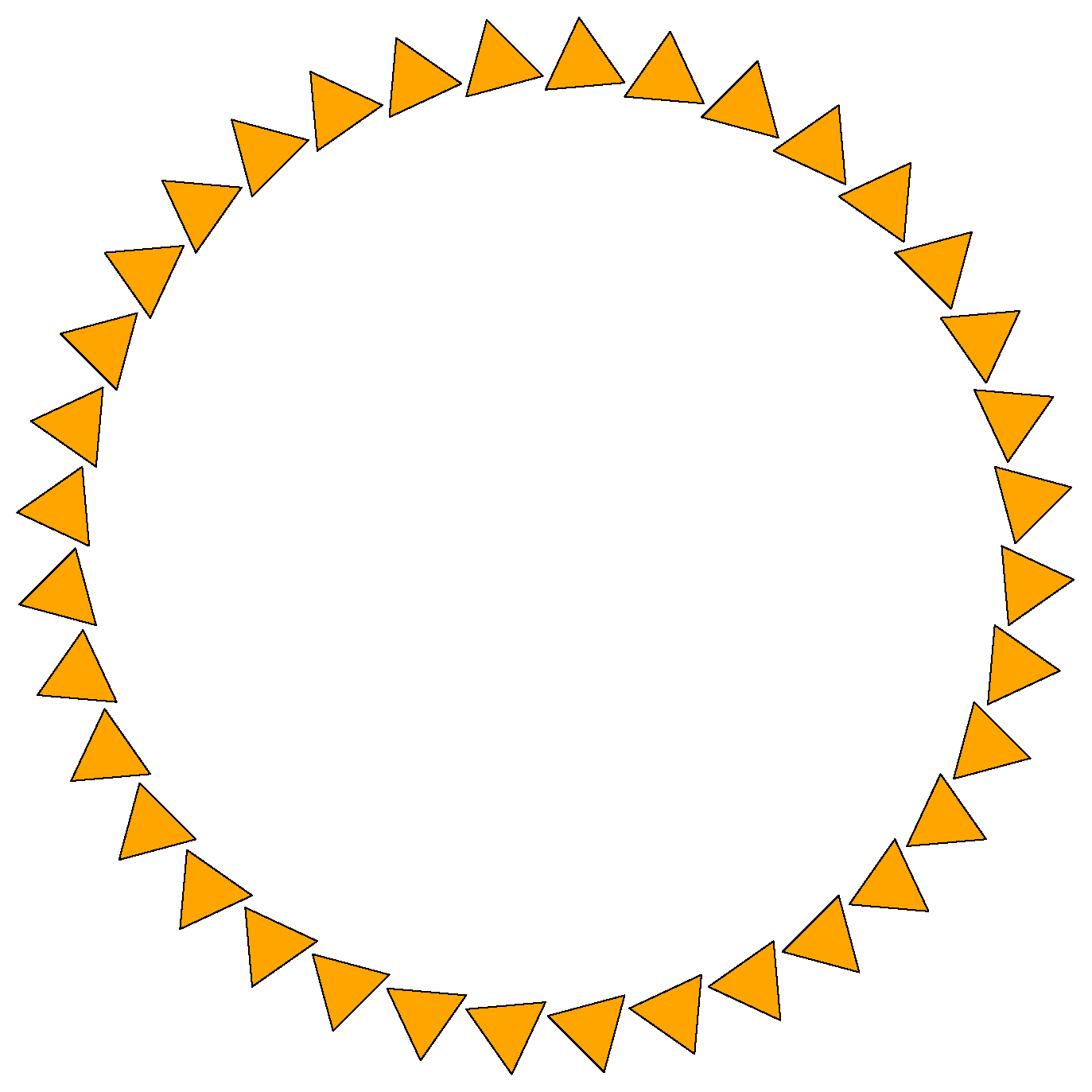} &
                \includegraphics[width=0.12\textwidth]{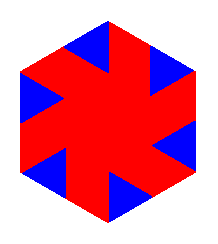} &
                \includegraphics[width=0.12\textwidth]{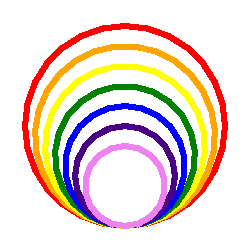} \\
                % \hline
                \\
                \includegraphics[width=0.12\textwidth]{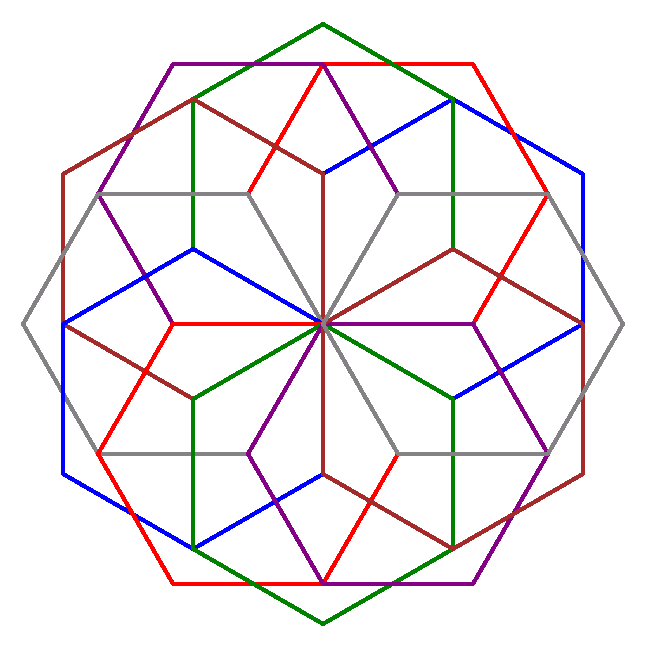} &
                \includegraphics[width=0.12\textwidth]{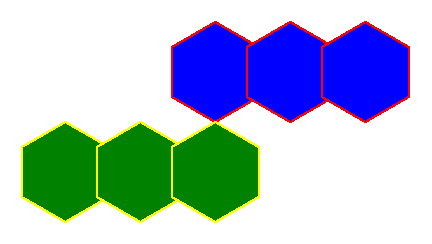} &
                \includegraphics[width=0.12\textwidth]{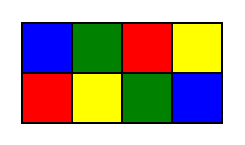} &
                \includegraphics[width=0.12\textwidth]{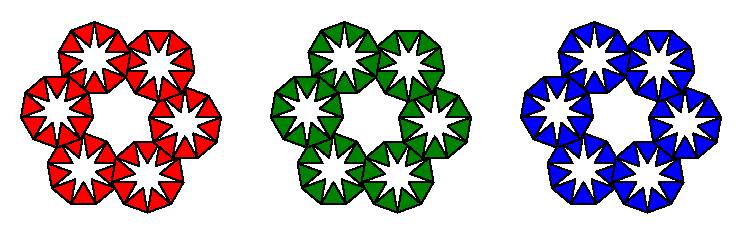} &
                \includegraphics[width=0.12\textwidth]{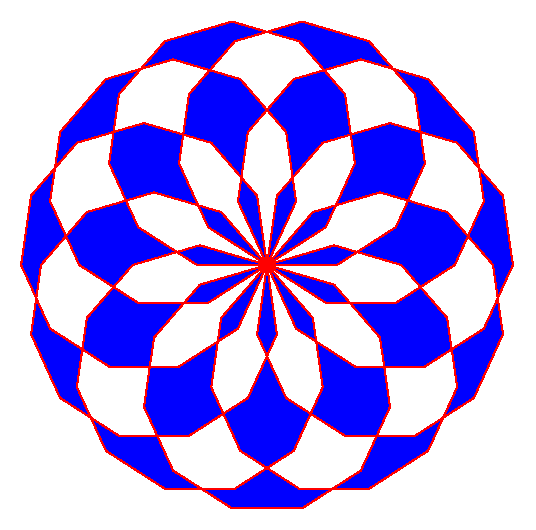} \\
                % \hline
                \\
                \includegraphics[width=0.12\textwidth]{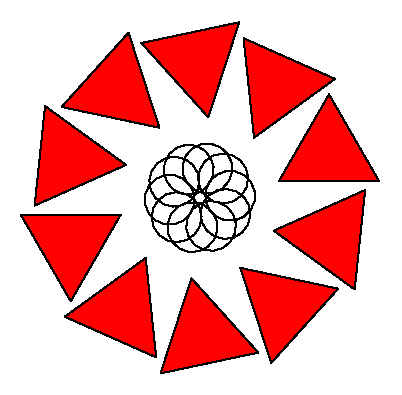} &
                \includegraphics[width=0.12\textwidth]{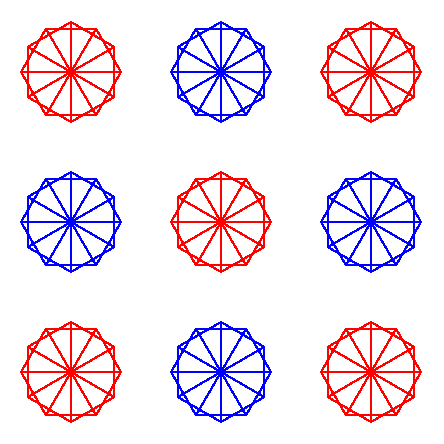} &
                \includegraphics[width=0.12\textwidth]{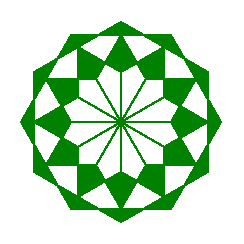} &
                \includegraphics[width=0.12\textwidth]{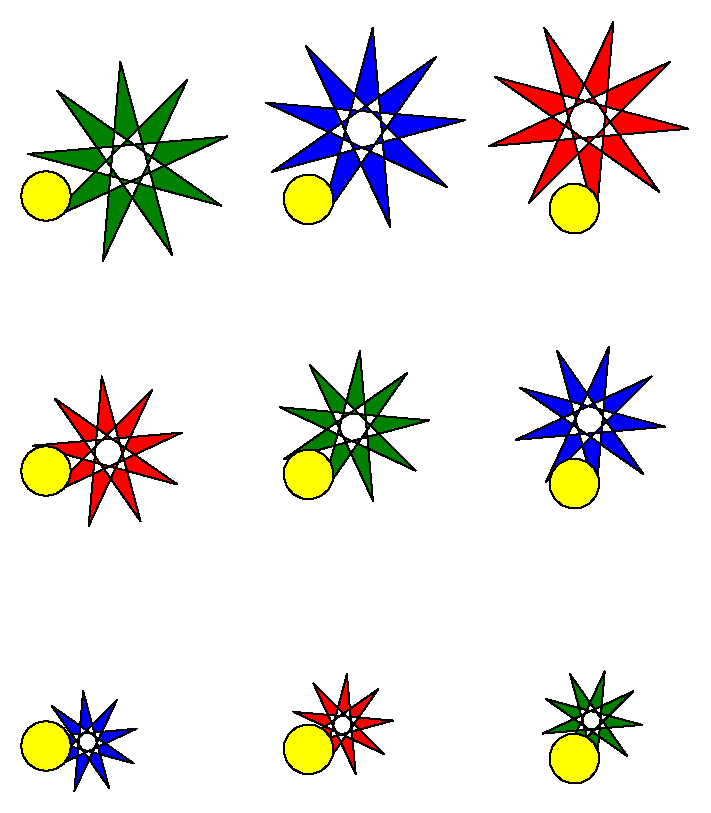} &
                \includegraphics[width=0.12\textwidth]{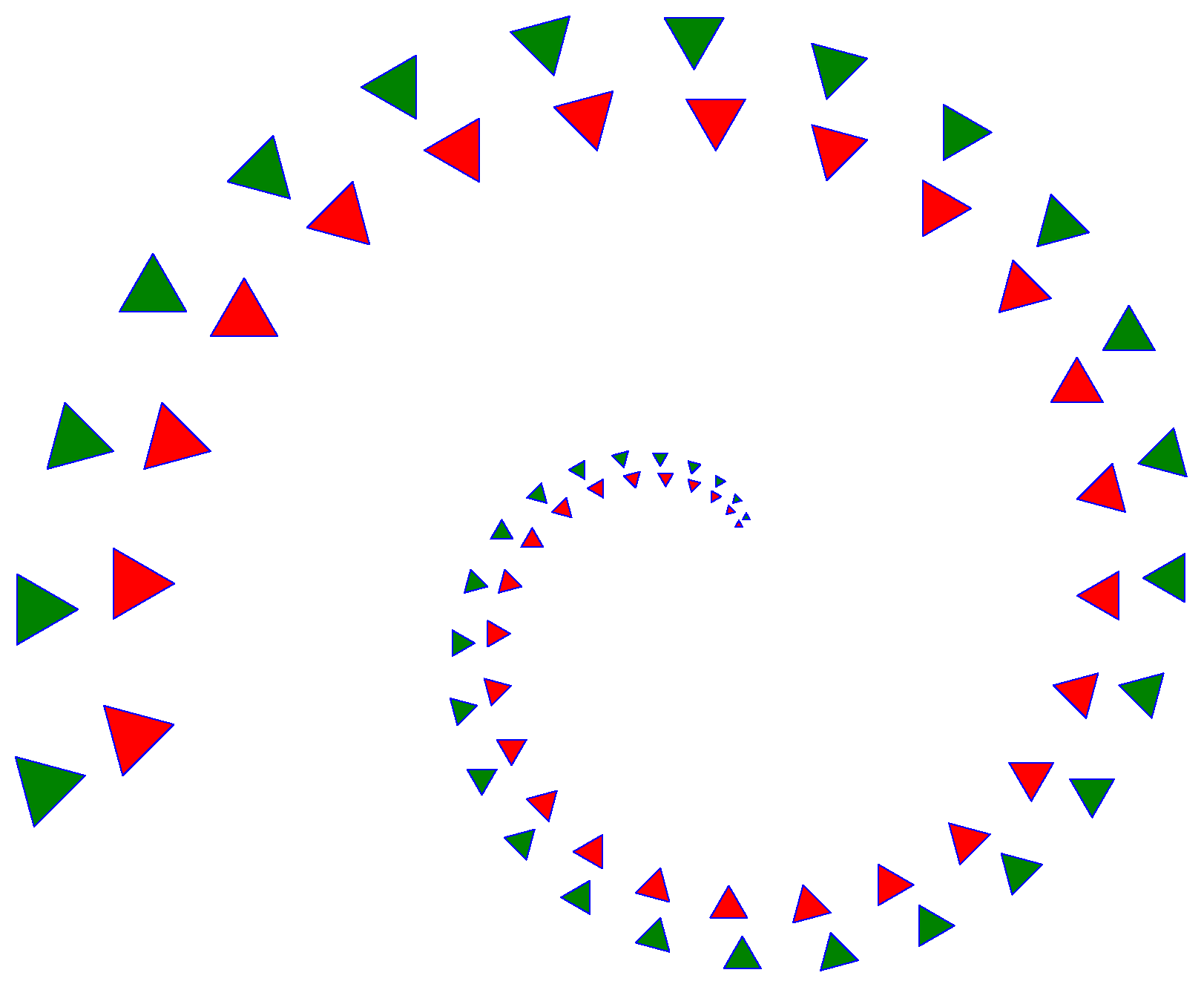} \\
                % \hline
                \\
                \includegraphics[width=0.12\textwidth]{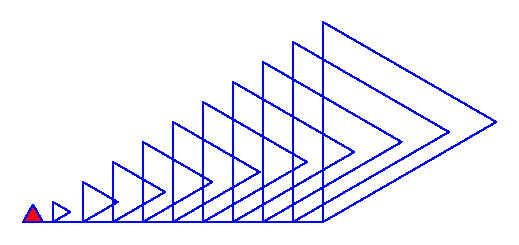} &
                \includegraphics[width=0.12\textwidth]{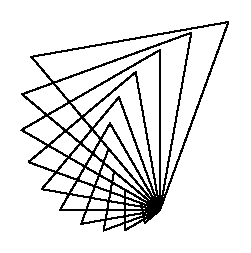} &
                \includegraphics[width=0.12\textwidth]{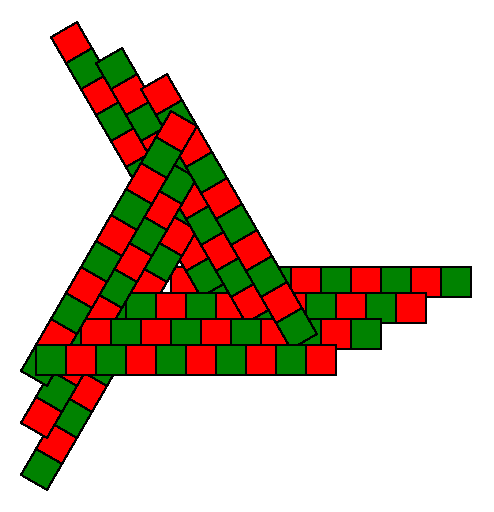} &
                \includegraphics[width=0.12\textwidth]{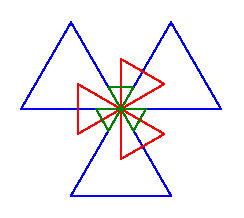} &
                \includegraphics[width=0.12\textwidth]{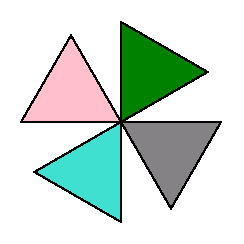} \\
                % \hline
                \\
                \includegraphics[width=0.12\textwidth]{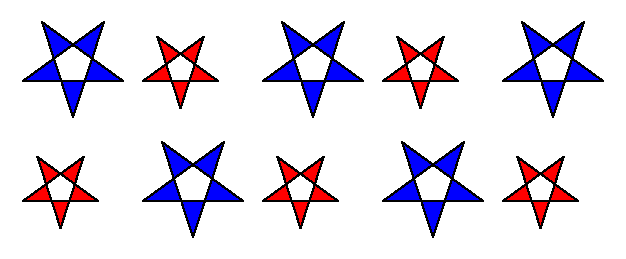} &
                \includegraphics[width=0.12\textwidth]{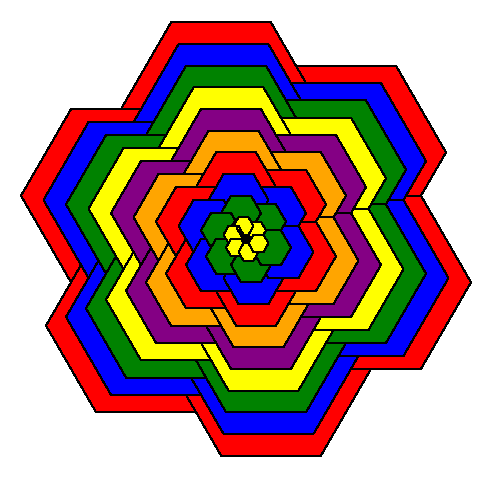} &
                \includegraphics[width=0.12\textwidth]{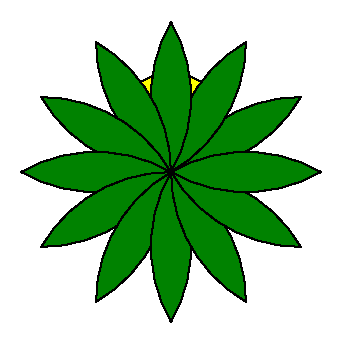} &
                \includegraphics[width=0.12\textwidth]{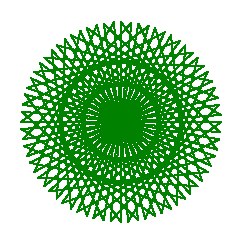} &
                \includegraphics[width=0.12\textwidth]{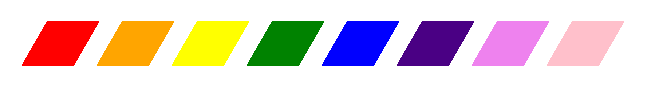} \\
                \\
                \includegraphics[width=0.12\textwidth]{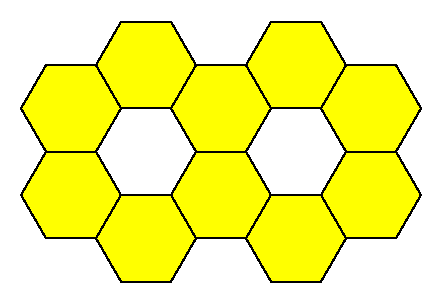} &
                \includegraphics[width=0.12\textwidth]{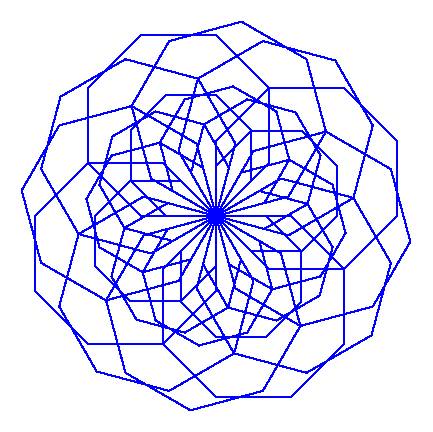} &
                \includegraphics[width=0.12\textwidth]{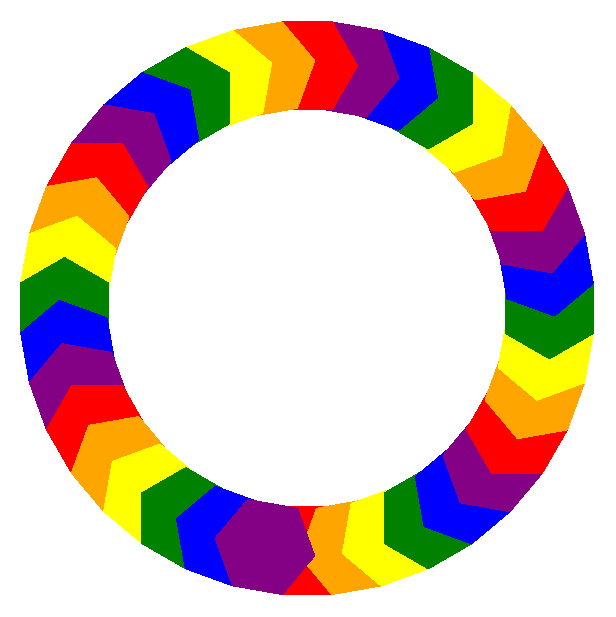} &
                \includegraphics[width=0.12\textwidth]{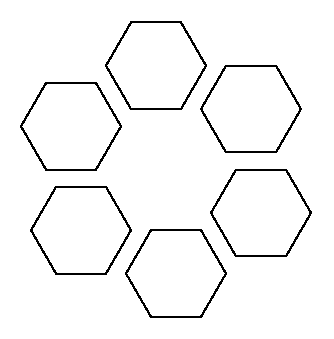} &
                \includegraphics[width=0.12\textwidth]{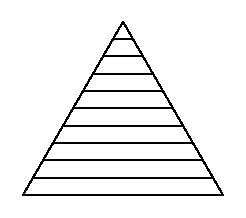} \\
                % \bottomrule
            \end{tabular}
            \vspace{0.05\linewidth}
            \caption{Generated images in \datasetTrain{}.} 
            \label{fig:examples.dataset_train.generated}
        \end{subfigure}
    \end{minipage}
    \hfill
    \begin{minipage}[b]{0.55\textwidth}
        \begin{subfigure}{\textwidth}
            \tiny
            \begin{lstlisting}[breaklines=true,breakindent=0pt,xleftmargin=0cm,xrightmargin=0cm,framexleftmargin=0.05cm,framexrightmargin=0.05cm]
## Image Description

The image displays a vibrant, symmetrical spiral pattern composed of nested heptagons (seven-sided polygons). The heptagons decrease in size as they spiral inward, creating a dynamic and visually appealing effect. The colors transition smoothly from green on the outer edges to red at the center, enhancing the depth and dimensionality of the spiral. The overall structure is highly symmetrical, with each heptagon rotated slightly relative to the previous one, contributing to the spiral effect.

## Analysis & Solution Code

To create the pattern shown in the image using Turtle Graphics, 
we need to follow these steps:

1. $\textbf{Initialize the Turtle}$: Set up the turtle graphics environment and configure the drawing speed.
2. $\textbf{Define the Heptagon}$: Create a function to draw a heptagon with a specified size.
3. $\textbf{Draw the Spiral}$: Use a loop to draw a series of heptagons, decreasing their size and adjusting their position and rotation with each iteration to form a spiral.
4. $\textbf{Color Gradient}$: Apply a color gradient to the heptagons, transitioning from green to red as the size decreases.

Here is the code with comprehensive docstrings and comments to 
create the pattern:

```python
def draw(t):
    """
    Draws a spiral pattern of nested heptagons with a color 
    gradient from green to red.
    """
    t.speed(0)  # Set the drawing speed to the maximum
    angle = 10  # Initial angle for the spiral rotation
    size = 100  # Initial size for the heptagons

    def draw_heptagon(size):
        """
        Draws a heptagon with the given size.
        """
        for _ in range(7):
            t.forward(size)  # Draw each side of the heptagon
            t.left(360 / 7)  # Turn to the next side

    # Draw the spiral pattern of heptagons
    while size > 10:
        # Set the fill color with a gradient from green to red
        t.fillcolor((1 - size / 100, size / 100, 0))
        t.begin_fill()  # Start filling the shape
        draw_heptagon(size)  # Draw the heptagon
        t.end_fill()  # End filling the shape
        t.right(angle)  # Rotate right to create the spiral effect
        size -= 5  # Decrease the size of the heptagon for the next iteration
        t.left(10)  # Adjust the rotation slightly to maintain the spiral pattern
```
            \end{lstlisting}
            \caption{Generated CoT label for the first image \raisebox{-0.25\height}{\includegraphics[width=0.04\textwidth]{appendix-figs/images_seed/pentagon_rect_addedge--chatcmpl-b414e34e-a884-48fa-ac2d-25b4a7f18b06.png}} in \autoref{fig:examples.dataset_train.generated}.}
            \label{fig:examples.dataset_train.label}
        \end{subfigure}
    \end{minipage}
    \caption{Examples of images in the seed dataset and the \datasetTrain{} dataset. The seed dataset includes $10$ task images and their corresponding solution codes. The \datasetTrain{} dataset includes $738$k images generated from the seed dataset using our data generation technique \datagenOurs{}. \datagenOurs{} can generate diverse and high-quality images by evolving from a small seed dataset.}
    \label{fig:examples.dataset_train}
\end{figure*}

%% file: appendix-figs/fig_example_diff.tex
% !TEX root =  main.tex
%%%%%%%%%%%%%%%%%%%%%%%%%%%%%%%%%%%%%%%
%%%%%%%%%%%%%%%%%%%%%%%%%%%%%%%%%%%%%%%

\begin{figure}[t]
    \centering
    \begin{subfigure}[b]{0.32\linewidth}
      \centering
      \includegraphics[width=0.7\textwidth]{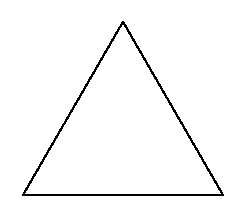}
      \caption{\difficultyEasy}
      \label{fig:appendix:difficulty_examples.easy}
    \end{subfigure}
    \begin{subfigure}[b]{0.32\linewidth}
      \centering
      \includegraphics[width=0.7\textwidth]{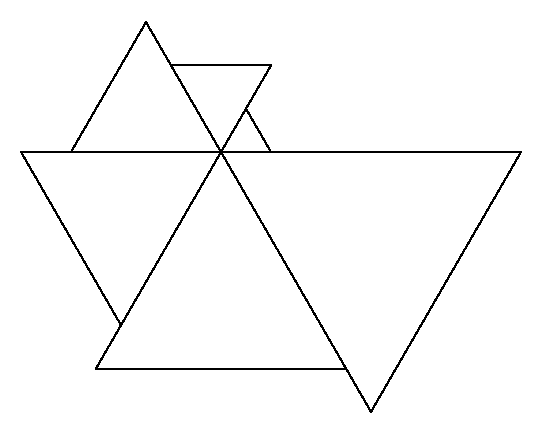}
      \caption{\difficultyMedium}
      \label{fig:appendix:difficulty_examples.medium}
    \end{subfigure}
    \begin{subfigure}[b]{0.32\linewidth}
      \centering
      \includegraphics[width=0.5\textwidth]{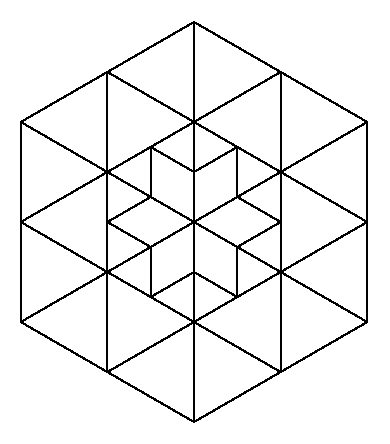}
      \caption{\difficultyHard}
      \label{fig:appendix:difficulty_examples.hard}
    \end{subfigure}
    \caption{Examples showing images of different difficulty levels from the \datasetAll{} dataset.}
    \label{fig:appendix:difficulty_examples}
\end{figure}

%% file: appendix_experiments.tex
% !TEX root =  main.tex
%%%%%%%%%%%%%%%%%%%%%%%%%%%%%%%%%%%%%%%%%%%%%%%
%%%%%%%%%%%%%%%%%%%%%%%%%%%%%%%%%%%%%%%%%%%%%%%

%%%%%%%%%%%%%%%%%%%%%%%%%%%%%%%%%%%%%%%%% 

\section{Additional Experiments and Analysis} \label{sec:appendix:additional_experiments}

In this section, we provide additional experiments and analysis of our synthetic data generation technique \datagenOurs{} and the model performance on \benchmark{}.

\subsection{Analysis of the Size of the Dataset Generated by \datagenOurs{}} \label{sec:appendix:experiments:dataset_size}
We analyze the exponential growth rate of datasets generated by the \datagenOurs{} framework. Specifically, we derive a mathematical formulation that describes how the dataset size expands iteratively, starting from an initial seed dataset and growing with each iteration. 
% %
Formally, assume that we have an initial seed dataset $\mathcal{D}_0$ containing $|\mathcal{D}_0|$ samples. Starting from $\mathcal{D}_0$, the framework generates a dataset $\mathcal{D}_t$ after $t$ iterations, where the size $|\mathcal{D}_t|$ can be expressed as:
\begin{equation}
\begin{split}
    |\mathcal{D}_{t}| &= |\mathcal{D}_{t-1}| \cdot p \cdot k \cdot (1-d_{t-1}) \\
    &= |\mathcal{D}_{0}| \cdot (pk)^t \cdot \prod_{i=1}^{t} (1-d_{i-1}),
\end{split}
\label{eq:dataset_growth}
\end{equation}
where $p \geq 1$ is the number of pairs of code sampled from the dataset in each iteration as the reference-guided code-to-code mutation example (e.g., $p=4$, $8$, $16$), $d_i \in [0, 1]$ is the duplicate rate in the $i$-th iteration, and $k \in [0, 1]$ is the top percentage of samples selected from the dataset in the elite selection stage. If we assume that the duplicate rate is the same for all iterations (i.e., $d_i = d$), then the size of the dataset $|\mathcal{D}_t|$ can be simplified as:
\begin{equation}
\begin{split}
    |\mathcal{D}_{t}| &= |\mathcal{D}_{0}| \cdot \big(pk(1-d)\big)^t,
    \end{split}
\label{eq:dataset_growth_simplified}
\end{equation}
where the size of the dataset $|\mathcal{D}_t|$ grows exponentially with the number of iterations $t$ with a growth rate of $pk(1-d)$. In our implementation, we set $p=16$ and $k=0.7$ for generating \datasetTrain{}, resulting in a growth rate of around $9.7$ at each iteration.

\subsection{Scaling of Fine-tuning Performance with Dataset Size} \label{sec:experiments:scaling}

%
\input{appendix-figs/fig_dataset_iteration_performance}
We study how fine-tuning performance scales with the size of datasets generated by \datagenOurs{} across different iterations. To this end, we fine-tune Pixtral-12B on datasets from each iteration and evaluate the resulting models, as shown in \autoref{fig:dataset_iteration_performance}.  
The dataset grows exponentially with the number of iterations, at a rate of roughly $9.7$, meaning each iteration produces a dataset approximately $9.7$ times larger than the previous one.  
Performance improvements depend on the dataset: for \datasetBasic{}, fine-tuning scales linearly with dataset size, while for \datasetSyn{}, the improvement is nearly exponential. In contrast, for the out-of-distribution dataset \datasetHand{}, performance saturates after the first iteration and remains stable in subsequent iterations. These results suggest that exponentially larger datasets generated by \datagenOurs{} can improve fine-tuning performance, although out-of-distribution datasets like \datasetHand{} do not benefit from the increased data.

\begin{table*}[t]
    \centering
    \begin{subfigure}[b]{0.43\textwidth}
        \scalebox{0.72}{
        \centering
        \begin{tabular}{l>{\raggedleft\arraybackslash}p{1.3cm}>{\raggedleft\arraybackslash}p{1.8cm}}
            \toprule
            &  \datasetBasicNoPrefix{} & \datasetHandNoPrefix{} \\
            \midrule
            \qwentwovl{}-72B & 11.76 & \textbf{7.84} \\
            \qwentwovl{}-72B-\ft{} & \textbf{35.29} & 6.86 \\
            \midrule
            \qwentwovl{}-7B & 0.98 & 0.00 \\
            \qwentwovl{}-7B-\ft{} & \textbf{28.43} & \textbf{6.86} \\
            \midrule
            \pixtral{}-12B & 9.80 & 2.94 \\
            \pixtral{}-12B-\ft{} (w/ CoT) & \textbf{27.45} & \textbf{9.80} \\
            \pixtral{}-12B-\ft{} (w/o CoT) & 22.55 & 0.00 \\
            \bottomrule
        \end{tabular}
        }
        \caption{Success rates (\%) on in-domain datasets.}
        \label{fig:analysis.ood}
    \end{subfigure}%
    \hfill
    \begin{subfigure}[b]{0.5\textwidth}
        \centering
        \scalebox{0.9}{
        \begin{tabular}{lrr}
            \toprule
            & HumanEval+ & MBPP+ \\
            \midrule
            \qwentwovl{}-72B       & \textbf{85.4} & \textbf{77.5} \\
            \qwentwovl{}-72B-\ft{} & 67.7 & 73.5 \\
            \midrule
            \qwentwovl{}-7B       & \textbf{70.7} & \textbf{55.3} \\
            \qwentwovl{}-7B-\ft{}  & 28.0 & 33.1 \\
            \bottomrule
        \end{tabular}
        }
        \caption{Success rates (\%) on out-of-domain datasets.}
        \label{fig:ood_benchmarks}
    \end{subfigure}
    \caption{(a) Success rates (\%) on in-domain datasets; \datasetHandNoPrefix{} contains hand-drawn OOD tasks from the same domain. (b) Pass@1 success rates (\%) on out-of-domain program synthesis benchmarks; fine-tuned models are not tuned for these tasks.}
    \label{fig:ood_combined}
\end{table*}

\subsection{Analysis of Out-of-distribution Performance of Fine-tuned Models} \label{sec:exp_ood_analysis}

We analyze fine-tuned models on both in-domain and out-of-domain out-of-distribution (OOD) tasks, providing insights into their generalization capabilities. Results are shown in \autoref{fig:ood_combined}.

We examine whether fine-tuned models can solve in-domain out-of-distribution tasks by comparing them with their base models on \datasetBasicNoPrefix{} and the hand-drawn OOD dataset \datasetHandNoPrefix{}. \datasetHandNoPrefix{} is created by hand-drawing images from \datasetBasicNoPrefix{} tasks and thus serves as an in-domain OOD dataset.
As shown in \autoref{fig:analysis.ood}, fine-tuning consistently improves performance on \datasetBasicNoPrefix{}, but offers little benefit on \datasetHandNoPrefix{} for moderately performing models (\qwentwovl-72B) and noticeable gains for weaker ones (\qwentwovl-7B, \pixtral-12B).
We hypothesize that the CoT labeling in the data generation process may enhance OOD performance, as it generates image descriptions that help ignore irrelevant variations in hand-drawn images. To test this, we ablate the CoT labeling step in the data generation and fine-tune \pixtral{}-12B to obtain \pixtral{}-12B-\ft{} (w/o CoT). As shown in \autoref{fig:analysis.ood}, this model fails on \datasetHandNoPrefix{} despite outperforming \pixtral{}-12B on \datasetBasicNoPrefix{}, showing that CoT labeling is useful for OOD generalization.

We investigate whether fine-tuning affects performance on out-of-domain tasks.
To this end, we test fine-tuned \qwentwovl{} models against their base versions on out-of-domain benchmarks, including HumanEval+ and MBPP+~\citep{DBLP:conf/nips/LiuXW023}. 
Pass@1 success rates are shown in \autoref{fig:ood_benchmarks}. The results show a clear drop in performance after fine-tuning, consistent with prior work showing that fine-tuning may lead to some degree of forgetting~\citep{DBLP:conf/acl/ZengLLWLD024,DBLP:conf/emnlp/Li0FT24}. This forgetting issue is more pronounced in the 7B model, suggesting that smaller models are more prone to overfitting and losing general capabilities. The 72B model retains better generality, which might be due to the higher capacity in retaining general knowledge.

\input{appendix-figs/fig_cot_prompting}

\subsection{Influence of the CoT Prompting on Model Performance} \label{sec:appendix:experiments:cot}
We investigate the influence of the CoT prompting on base models' performance. We experiment with various open-source VLMs, with and without CoT prompting. For CoT prompting, we require the model to generate the solution code in the following step-by-step manner: (i) describe the image in detail, (ii) analyze the image and propose steps to create the pattern, and (iii) generate the solution code with comprehensive docstrings and comments. For non-CoT prompting, we only require the model to generate the code, without the above steps explicitly mentioned. The comparison results are shown in \autoref{fig:appendix:cot_prompting}.
The results indicate that the effectiveness of CoT prompting varies across different models and datasets, and there is no clear indication that CoT prompting can improve performance in our domain. For instance, \qwentwovl{}-7B shows improved performance with CoT prompting on both datasets, whereas \internvltwo{}-76B performs better without CoT prompting on both datasets. This inconsistency may stem from the reasoning-intensive nature of our tasks, where each type of task demands different reasoning steps, making it challenging to devise a consistent CoT prompting strategy applicable to all tasks. Furthermore, models trained on different datasets may develop distinct reasoning preferences, causing the same CoT strategy to enhance performance in some models while potentially confusing others, resulting in inconsistent performance of CoT prompting in our domain.

\input{appendix-figs/fig_lora_rank}

\subsection{Influence of LoRA Rank and Vision Tower for Fine-tuning Performance} \label{sec:appendix:experiments:lora}
We investigate the influence of LoRA rank and vision tower fine-tuning on model performance. To do this, we conduct fine-tuning experiments on \pixtral{}-12B with LoRA ranks of $64$, $128$, and $256$ using the $738$k \datasetTrain{} dataset (without CoT labeling), training each configuration for $1$ epoch. For each LoRA rank, we set the LoRA alpha parameter to twice the rank value. Additionally, we examine the impact of freezing versus unfreezing the vision tower during fine-tuning. By unfreezing the vision tower, we enable parameter tuning of the visual encoder component of the VLM, allowing the model to adapt its visual representations during fine-tuning.
The results are shown in \autoref{fig:appendix:lora_rank}. We find that unfreezing the vision tower can enhance performance. Specifically, in our experiments with LoRA rank $64$, unfreezing the vision tower increases the success rate from $10.78\%$ to $17.65\%$ on \datasetBasic{} and from $8.38\%$ to $9.96\%$ on \datasetAll{}. Additionally, the choice of LoRA rank also affects the performance, with rank $128$ achieving the best results in our case.

\subsection{Performance of VLMs Using Pass@K Metrics} \label{sec:appendix:experiments:topk}

\input{appendix-figs/fig_passk}

In the main paper, we report the evaluation results of different VLMs on our benchmark using a greedy decoding strategy (i.e., \texttt{temperature=0}). To provide a more comprehensive evaluation, we also experiment with a random sampling strategy by randomly sampling $N$ samples from the model and then calculating the Pass@K results. Following previous works~\citep{DBLP:journals/corr/abs-2308-12950,DBLP:conf/iclr/ZhuoVCH0WYZHPB025}, we compute Pass@K results with random sampling by generating $N=5$ samples with $\texttt{top\_p=0.95}$ and $\texttt{temperature = 0.8}$. Then we calculate Pass@1, Pass@3, and Pass@5, respectively. Although generating many more samples ($N \geq K$) is recommended to reduce bias, we adopt the lower bound due to limited computational resources. The results are provided in \autoref{fig:appendix:passk}.

%% file: appendix-figs/fig_dataset_iteration_performance.tex
% !TEX root =  main.tex
%%%%%%%%%%%%%%%%%%%%%%%%%%%%%%%%%%%%%%%
%%%%%%%%%%%%%%%%%%%%%%%%%%%%%%%%%%%%%%%

\begin{figure}[ht!]
    \centering
    \includegraphics[width=0.5\textwidth]{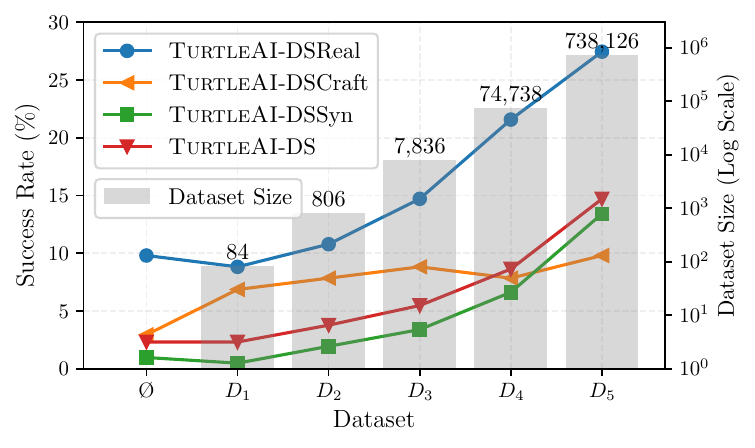}
    \caption{Performance of fine-tuned \pixtral{}-12B-\ft{} using datasets generated by \datagenOurs{} across different iterations.}

    \label{fig:dataset_iteration_performance}
\end{figure} 

%% file: appendix-figs/fig_cot_prompting.tex
% !TEX root =  main.tex
%%%%%%%%%%%%%%%%%%%%%%%%%%%%%%%%%%%%%%%
%%%%%%%%%%%%%%%%%%%%%%%%%%%%%%%%%%%%%%%

\begin{table*}[t]
    \centering
    \scalebox{0.9}{
        \begin{tabular}{lrrrr}
            \toprule
            & \multicolumn{2}{c}{\datasetBasic{}} & \multicolumn{2}{c}{\datasetAll{}} \\
            \cmidrule(lr){2-3} \cmidrule(lr){4-5}
            & CoT & Non-CoT & CoT & Non-CoT \\
            \midrule
            \pixtral{}-Large & \textbf{11.76} & 10.78 & 4.74 & \textbf{6.56}\\
            \qwentwovl{}-72B & 10.78 & \textbf{11.76} & \textbf{4.13} & 3.52 \\
            \llavaov{}-72B & \textbf{8.82} & 4.90 & \textbf{2.19} & \textbf{2.19} \\
            \internvltwo{}-76B & 8.82 & \textbf{11.76} & 2.79 & \textbf{3.28} \\
            \molmo{}-72B & \textbf{9.80} & 3.92 &  \textbf{2.92} & 2.31 \\
            \midrule
            \pixtral{}-12B & 1.96 & \textbf{9.80} & 1.09 & \textbf{2.31}\\
            \llavaov{}-7B & \textbf{3.92} & \textbf{3.92} & 0.97 & \textbf{1.09} \\
            \qwentwovl{}-7B & \textbf{4.90} & 0.98 & \textbf{1.09} & 0.12 \\
            \molmo{}-7B & 0.00 & 0.00 & \textbf{0.12} & 0.00 \\
            \internvltwo{}-8B & \textbf{2.94} & 0.00 & \textbf{0.73} & 0.12 \\
            \bottomrule
        \end{tabular}
    }
    \caption{Symbolic success rates of different base VLMs on \datasetBasic{} and \datasetAll{} with and without CoT prompting. The best performance is highlighted in \textbf{bold} for each model.}
    \label{fig:appendix:cot_prompting}
\end{table*}

%% file: appendix-figs/fig_lora_rank.tex
% !TEX root =  main.tex
%%%%%%%%%%%%%%%%%%%%%%%%%%%%%%%%%%%%%%%
%%%%%%%%%%%%%%%%%%%%%%%%%%%%%%%%%%%%%%%

\begin{table*}[t!]
    \centering
    \scalebox{1}{
        \begin{tabular}{lrrrr}
            \toprule
            & \multicolumn{2}{c}{Fine-tuning Parameters} & \multicolumn{2}{c}{Success Rate (\%)} \\
            \cmidrule(lr){2-3} \cmidrule(lr){4-5}
            & Vision Tower & LoRA rank & \datasetBasic{} & \datasetAll{} \\
            \midrule
            \pixtral{}-12B-\ft{} & Freeze & $64$ & $10.78$ & $8.38$ \\
            \pixtral{}-12B-\ft{} & Unfreeze & $64$ & $17.65$ & $9.96$ \\
            \pixtral{}-12B-\ft{} & Unfreeze & $128$ & $\textbf{22.55}$ & $\textbf{12.67}$ \\
            \pixtral{}-12B-\ft{} & Unfreeze & $256$ & $15.69$ & $10.81$ \\
            \bottomrule
        \end{tabular}
    }
    \caption{Influence of LoRA rank and vision tower on the performance of fine-tuning. We experiment with LoRA ranks of $64$, $128$, and $256$, freezing and unfreezing the vision tower to fine-tune \pixtral{}-12B model using the $738$k \datasetTrain{} dataset (without CoT labeling), with each setting trained for $1$ epoch. Unfreezing the vision tower and using LoRA rank $128$ yields the best performance.}
    \label{fig:appendix:lora_rank}
\end{table*}

%% file: appendix-figs/fig_passk.tex
% !TEX root =  main.tex
%%%%%%%%%%%%%%%%%%%%%%%%%%%%%%%%%%%%%%%%%%%%%%%
%%%%%%%%%%%%%%%%%%%%%%%%%%%%%%%%%%%%%%%%%%%%%%%

%%%%%%%%%%%%%%%%%%%%%%%%%%%%%%%%%%%%%%%%% 

\begin{table*}[ht!]
    \centering
    \scalebox{1}{
    \begin{tabular}{lrrrrr}
      \toprule
      & & \multicolumn{1}{c}{Greedy Decoding} & \multicolumn{3}{c}{Random Sampling} \\
      \cmidrule(lr){3-3} \cmidrule(lr){4-6}  
      & Size & Pass@1 & Pass@1 & Pass@3 & Pass@5 \\
      \midrule
    \internvltwo{} & 76B & $3.28$ & $2.72$ & $4.79$ & $5.83$ \\
    \llavaov{} & 72B & $2.19$ & $2.09$ & $4.02$ & $5.35$ \\

    \qwentwovl{} & 72B  & $3.52$ & $3.18$ & $5.53$ & $6.93$ \\
    \qwentwovl{}-\ft{} & 72B  & $19.56$ & $16.79$ & $25.65$ & $29.77$ \\
    \midrule
    \internvltwo{} & 8B & $0.12$ & $0.19$ & $0.51$ & $0.73$ \\
    \llavaov{} & 7B & $1.09$ & $0.83$ & $1.65$ & $2.19$ \\
    \qwentwovl{} & 7B  & $0.12$ & $0.10$ & $0.29$ & $0.49$ \\
    \qwentwovl{}-\ft{} & 7B  & $13.37$ & $11.96$ & $19.77$ & $23.69$ \\
    \pixtral{} & 12B & $2.31$ & $1.48$ & $3.23$ & $4.25$ \\
    \pixtral{}-\ft{} & 12B & $14.70$  & $10.28$ & $17.36$ &  $20.66$ \\
      \bottomrule
    \end{tabular}
    }
    \caption{Symbolic success rates (\%) of VLMs on the dataset \datasetAll{} with greedy decoding and random sampling. For greedy decoding, we report Pass@1 using $\texttt{temperature}=0$. For random sampling, we use $\texttt{temperature}=0.8$ and $\texttt{top\_p}=0.95$, where for each Pass@K metric, we generate $N=5$ samples.}
    \label{fig:appendix:passk}
  \end{table*}

%% file: appendix_evaluation_reliability.tex
% !TEX root =  main.tex
%%%%%%%%%%%%%%%%%%%%%%%%%%%%%%%%%%%%%%%%%%%%%%%
%%%%%%%%%%%%%%%%%%%%%%%%%%%%%%%%%%%%%%%%%%%%%%%

%%%%%%%%%%%%%%%%%%%%%%%%%%%%%%%%%%%%%%%%% 

\section{Reliability of the Evaluation Framework} \label{sec:appendix:evaluation_reliability}

\input{appendix-figs/fig_evaluation_confusion_matrix}

To assess the reliability of our evaluation framework, we perform a manual evaluation and compare it with the accuracies of both symbolic and embedding-based comparisons. To do this, we first perform a manual evaluation of all generated images in the \datasetAll{} dataset from \gptfouro{}. This involves comparing the ground-truth image with the corresponding image produced by executing the generated code from \gptfouro{}, and manually verifying whether each generated image is visually identical to the ground-truth image. This manual evaluation involves a total of $823$ image-code pairs. After this manual evaluation, we compare our results with both symbolic and embedding-based comparisons to evaluate the accuracy of our evaluation framework.

\paragraph{Accuracy of the symbolic comparison.}
After manual evaluation, we use our manual evaluation results as ground truth and calculate the precision, recall, F1 score, and accuracy for the results of the symbolic comparison. The results are shown in \autoref{fig:symbolic_confusion_matrix}. Our symbolic comparison achieves a precision of $0.974$, recall of $0.937$, F1 score of $0.955$, and accuracy of $0.991$, showing that the symbolic comparison can correctly identify almost all of the generated images compared against the manual evaluation.

\input{appendix-figs/fig_precision_recall_embedding}
\paragraph{Accuracy of the embedding-based comparison.}
The embedding-based comparison first calculates a similarity score and then determines \emph{success} or \emph{fail} by comparing the similarity score against a threshold value. To determine the optimal threshold value, we plot how different threshold values affect the precision, recall, and F1 score, and select the threshold value that maximizes the F1 score. \autoref{fig:precision_recall_embedding} shows the precision, recall, and F1 score at different threshold values. We find that using a threshold of $0.95$ achieves the highest F1 score of $0.896$. Therefore, we use a threshold of $0.95$ for the embedding-based comparison, i.e., if the embedding score is greater than $0.95$, we consider the generated image as a \emph{success}. \autoref{fig:embedding_confusion_matrix} shows detailed statistics of the precision, recall, F1 score, and accuracy for the embedding-based comparison with a threshold of $0.95$.

%% file: appendix-figs/fig_evaluation_confusion_matrix.tex
% !TEX root =  main.tex
%%%%%%%%%%%%%%%%%%%%%%%%%%%%%%%%%%%%%%%
%%%%%%%%%%%%%%%%%%%%%%%%%%%%%%%%%%%%%%%

\begin{table*}[ht!]
    \centering
    \begin{subfigure}[b]{0.8\linewidth}
        \scalebox{0.9}{
        \begin{tabular}{rrrr}
            \toprule
            & Positive (Symbolic) & Negative (Symbolic) \\
            \midrule
            Positive (Manual) & 74 & 5 \\
            Negative (Manual) & 2 & 742 \\
            \midrule
            \midrule
            \multicolumn{3}{c}{$Precision=0.974$ \quad $Recall=0.937$ \quad $F1=0.955$ \quad $Accuracy=0.991$} \\
            \bottomrule
        \end{tabular}
        }
        \caption{Confusion matrix for the symbolic comparison.}
        \label{fig:symbolic_confusion_matrix}
    \end{subfigure}

    \vspace{0.03\linewidth}

    \begin{subfigure}[b]{0.8\linewidth}
        \scalebox{0.9}{
        \begin{tabular}{rrrr}
            \toprule
            & Positive (Embedding) & Negative (Embedding) \\
            \midrule
            Positive (Manual) & 69 & 10 \\
            Negative (Manual) & 6 & 738 \\
            \midrule
            \midrule
            \multicolumn{3}{c}{$Precision=0.873$ \quad $Recall=0.920$ \quad $F1=0.896$ \quad $Accuracy=0.981$} \\
            \bottomrule
        \end{tabular}
        }
        \caption{Confusion matrix for the embedding-based comparison.}
        \label{fig:embedding_confusion_matrix}
    \end{subfigure}
    \caption{Confusion matrices illustrating the accuracy of our evaluation framework by comparing the results of the symbolic and embedding-based comparisons against the manual comparison. The evaluation is conducted by manually annotating \gptfouro{}'s results on the \datasetAll{} dataset. (a) shows the confusion matrix for the symbolic comparison, which demonstrates high accuracy with an F1 score of 0.955 when compared against manual evaluation. (b) shows the confusion matrix for the embedding-based comparison, achieving an F1 score of 0.896 at a threshold value of 0.95.}
    \label{fig:evaluation_confusion_matrix}
\end{table*} 

%% file: appendix-figs/fig_precision_recall_embedding.tex
% !TEX root =  main.tex
%%%%%%%%%%%%%%%%%%%%%%%%%%%%%%%%%%%%%%%
%%%%%%%%%%%%%%%%%%%%%%%%%%%%%%%%%%%%%%%

\begin{figure}[ht]
    \centering
    \includegraphics[width=0.95\columnwidth]{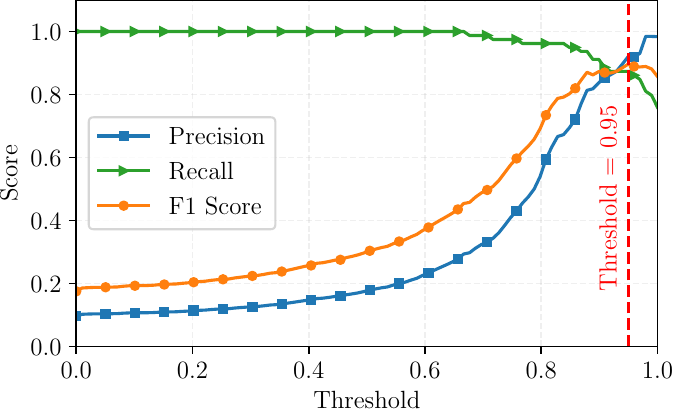}
    \caption{The relationship between precision, recall, and F1 score at different thresholds used in the embedding-based comparison. The best F1 score is achieved at a threshold of $0.95$, with F1 score of $0.896$.}
    \label{fig:precision_recall_embedding}
\end{figure} 

%% file: appendix_implementation.tex
% !TEX root =  main.tex
%%%%%%%%%%%%%%%%%%%%%%%%%%%%%%%%%%%%%%%%%%%%%%%
%%%%%%%%%%%%%%%%%%%%%%%%%%%%%%%%%%%%%%%%%%%%%%%

%%%%%%%%%%%%%%%%%%%%%%%%%%%%%%%%%%%%%%%%% 

\section{Implementation Details} \label{sec:appendix:implementation_details}

In this section, we detail the implementation of our dataset generation framework \datagenOurs{}, the model fine-tuning process, the evaluation process, and the evaluation framework \eval{}.

\subsection{Implementation Details of \datagenOurs{}} \label{sec:appendix:datagen_details}

We describe the implementation details of the dataset generation framework \datagenOurs{}.

\input{appendix-figs/fig_example_mutation}
\paragraph{Stage 1: Code mutation.} \label{sec:appendix:implementation:datagen:mutation} 
We use the Llama3.1-70B-Instruct model for code mutation. The model is queried with $\texttt{temperature}=0.5$ and $\texttt{top\_p}=1$. We use a higher temperature and $\texttt{top\_p}$ values to encourage the model to generate more diverse and creative code. During code mutation, we randomly sample $16$ pairs of $(\code{}_{\text{ref}_1}, \code{}_{\text{ref}_2})$ from the seed dataset for each input code $\code{}_{\text{in}}$. This results in $16$ possible mutated codes for $\code{}_{\text{in}}$ after applying the mutation for each pair of $(\code{}_{\text{ref}_1}, \code{}_{\text{ref}_2})$. 
An illustrative example of the code mutation process is provided in \autoref{fig:examples.code_mutation}.

\paragraph{Stage 2: Elite selection.} \label{sec:appendix:implementation:datagen:elite} 
The elite selection stage consists of two steps: deduplication and selection of elite samples. Given a dataset consisting of image-code pairs, we first perform deduplication to remove duplicate images. Specifically, we use a pre-trained ResNet18~\citep{DBLP:conf/cvpr/HeZRS16} image encoder to obtain the embedding for each image. We then use the DBSCAN clustering algorithm~\citep{DBLP:conf/kdd/EsterKSX96} to cluster these image embeddings, resulting in a set of clusters, where each cluster consists of similar images. For each cluster, we only preserve one sample and remove the rest. We use the DBSCAN clustering algorithm with parameters $\epsilon=0.2$, $\texttt{min\_samples}=2$, and the \texttt{euclidean} distance. After deduplication, we select the elite samples from the deduplicated dataset. To do this, we use \qwentwovl{}-72B as the model for selecting elite samples. We use the top $k=30\%$ for generating the \datasetSyn{} dataset and $k=70\%$ for generating the \datasetTrain{} dataset.

After the above two stages, we use Pixtral-Large as the model for CoT labeling. The model is queried with $\texttt{temperature}=0.1$ and $\texttt{top\_p}=0.001$. Note that this stage is only used to generate the \datasetTrain{} dataset for fine-tuning.

For querying the above models in different stages, we consistently use the vLLM inference engine to speed up inference. During inference, we use 8 $\times$ H100 GPUs, with \texttt{tensor\_parallel\_size} set to $8$, and \texttt{max\_num\_seqs} set to $64$. For every 100k samples generated during the elite selection (using Qwen2VL-72B-Instruct) or code mutation stage (using Llama3.1-70B-Instruct), the process takes approximately 8 hours. For CoT labeling (using Pixtral-Large), it takes around 13 hours to process every 100k samples.

\subsection{Implementation Details of Fine-tuning} \label{sec:appendix:finetuning_details}

We conduct fine-tuning experiments on three models: \qwentwovl{}-7B~\citep{DBLP:journals/corr/abs-2409-12191}, \qwentwovl{}-72B~\citep{DBLP:journals/corr/abs-2409-12191}, and Pixtral-12B~\citep{agrawal2024pixtral12b}. In our fine-tuning experiments, we use LoRA~\citep{DBLP:conf/iclr/HuSWALWWC22} for parameter-efficient fine-tuning. To determine the best LoRA rank and scaling factor, we experimented with LoRA ranks of $64$, $128$, and $256$, using a scaling factor of two times the LoRA rank in each case. We found that a rank of $128$ provides the best performance. Consequently, we use a LoRA rank of $128$ and a scaling factor of $256$ for all fine-tuning experiments. We also experimented with freezing and unfreezing the vision tower during fine-tuning and found that unfreezing the vision tower provides better performance. Therefore, we unfreeze the vision tower during fine-tuning for all fine-tuning experiments.

During fine-tuning, we use a learning rate schedule that combines a $10\%$ warmup phase where the learning rate linearly increases to $1e-4$, followed by cosine annealing, which gradually reduces the learning rate to $0$, ensuring stable training and smooth convergence~\citep{zheng-etal-2024-llamafactory}. 
We also reserve $1\%$ of the 738k training dataset for validation. For our fine-tuning experiments on \qwentwovl{}-7B and Pixtral-12B models, we observed that the validation loss increases after the first epoch, so we stopped fine-tuning after one epoch and report their performance at epoch 1 accordingly. Conversely, the \qwentwovl{}-72B model's performance continued to improve after the first epoch, so we fine-tuned it for two epochs in total. 

All fine-tuning experiments are conducted on an internal cluster using 8 $\times$ H100 GPUs, with each epoch taking approximately $15$ hours for \qwentwovl{}-7B, $112$ hours for \qwentwovl{}-72B, and $22$ hours for Pixtral-12B.

\subsection{Implementation Details of Evaluation} \label{sec:appendix:evaluation_details} 

\input{appendix-figs/fig_model_version}

\paragraph{Inference details of VLMs.} \label{sec:appendix:implementation:evaluation:inference}
For open-source VLMs, we download their pre-trained weights and perform inference locally.
We use the vLLM~\citep{DBLP:conf/sosp/KwonLZ0ZY0ZS23} engine for VLM inference to obtain the outputs of the evaluated open-source VLMs.
During inference, we set the \texttt{temperature} to $0$ and use different numbers of GPUs for different models depending on their parameter sizes: (i) 1 $\times$ A100 GPU for models with parameter sizes less than 7B, (ii) 2 $\times$ A100 GPUs for models with parameter sizes between 7B and 70B, and (iii) 8 $\times$ A100 GPUs for models with parameter sizes larger than 70B. We set the vLLM parameter \texttt{tensor\_parallel\_size} to 1, 2, and 8 for the three cases, respectively, and set \texttt{max\_num\_seqs} to \texttt{tensor\_parallel\_size} $\times$ 8.
We use the OpenAI API to evaluate proprietary models from OpenAI. For reasoning models, we set \texttt{reasoning\_effort} to \texttt{medium} and \texttt{max\_completion\_tokens} to 8192.

\paragraph{Details of the evaluation procedure.} \label{sec:appendix:implementation:evaluation:procedure}
For each task in our evaluation datasets, we provide the task image along with a fixed prompt template (see Figure~\ref{fig:appendix:prompt_template:code_synthesis}) to guide the VLMs in generating Turtle Graphics Python code. The model's output often includes an explanation along with the predicted code. We extract only the code snippet and disregard the rest. If multiple code snippets are present, we handle them differently based on the comparison method: 
\begin{itemize}
    \item \emph{Symbolic comparison}: we evaluate all the code snippets and consider the result a \emph{success} if any code snippet is successful.  
    \item \emph{Embedding-based comparison}: we evaluate only the longest code snippet. This is because embedding-based comparison involves batch processing when extracting image embeddings, making it inefficient to consider multiple code snippets.
\end{itemize}

\subsection{Implementation Details of the Evaluation Framework} \label{sec:appendix:evaluation_framework_details}
We describe the implementation details of our evaluation framework. Given a task image $\image{}$, the predicted code snippet $\hat{\code{}}$, and the solution code $\code{}$, our evaluation framework works as follows to evaluate the correctness of the predicted code $\hat{\code{}}$.
First, we execute both the solution code $\code{}$ and the predicted code $\hat{\code{}}$ using a customized Turtle Graphics emulator. This emulator inherits the built-in Turtle Graphics module and enables us to record all the drawing states, including the coordinates and the colors when drawing a line, filling a polygon, etc.
Second, we transform the recorded drawing states for $\code{}$ and $\hat{\code{}}$ into the same space to ensure the invariance to size, position, and line width of the drawing. Specifically, we perform the following three steps: 
\begin{itemize}
    \item \emph{Normalizing length of lines}: We rescale all recorded coordinates such that the maximum dimension of the entire pattern's bounding box is set to 300. This ensures that the drawings are uniformly scaled regardless of their original size, making our comparison invariant to the size of the drawing.
    \item \emph{Centering around the origin}: We translate all recorded lines so that they are centered around the origin. This involves calculating the center of the bounding box and shifting all coordinates accordingly. This ensures that the comparison is invariant to the position of the drawing.
    \item \emph{Standardizing pen size}: We standardize the pen size of all lines to $1$. This ensures that drawing line width does not affect the comparison of drawings, making our comparison invariant to the line width.
\end{itemize}
Third, we render these normalized drawing states into images in the sequence as they are recorded, resulting in standardized images $\image$ and $\hat{\image}$ for $\code{}$ and $\hat{\code{}}$, respectively.
Finally, our evaluation framework provides two comparison methods to compare these two images, which are described in detail as follows.

\paragraph{Symbolic comparison.} \label{sec:appendix:implementation:evaluation:symbolic} 
This compares the standardized images $\image$ and $\hat{\image}$ pixel-by-pixel. The high-level idea is to first count non-white pixels in both images and calculate the percentage of differing pixels among them. If this percentage is below a predefined value, the comparison result is \emph{success}; otherwise, the comparison result is \emph{fail}. 
More specifically, assume the images $\image$ and $\hat{\image}$ are of dimensions $H \times W$ and $\image{}_{i,j}$ and $\hat{\image}{}_{i,j}$ are the pixels at position $(i, j)$, respectively. We define a candidate set of pixels $\mathcal{P}$ that are considered for symbolic comparison:
\begin{equation}
    \begin{split}
\mathcal{P} = \big\{ (i, j) \mid \image_{i,j} \neq \text{white} \lor \hat{\image}_{i,j} \neq \text{white}, \\
\forall \, i \in \{1, \ldots, H\}, \, j \in \{1, \ldots, W\} \big\}.
\end{split}
\end{equation}
The pixel-wise difference between $\image$ and $\hat{\image}$ is computed as:
\begin{equation}
    \text{pixel\_diff}(\image{}, \hat{\image}) = \frac{\sum_{(i,j) \in \mathcal{P}} \mathbb{I}(\image{}_{i,j} \neq \hat{\image}{}_{i,j})}{|\mathcal{P}|},
\end{equation}
where $\mathbb{I}(\image{}_{i,j} \neq \hat{\image}{}_{i,j})$ is the indicator function that returns $1$ if $\image{}_{i,j} \neq \hat{\image}{}_{i,j}$ and $0$ otherwise. 
We establish a $\text{threshold}$ for pixel-wise differences to determine whether the image $\hat{\image}$ is a \emph{success} to match $\image$ or not. If $\text{pixel\_diff}(\image, \hat{\image}) < 1-\text{threshold}$, the image $\hat{\image}$ is considered a \emph{success} to match $\image$; otherwise, it is considered a \emph{fail}. For our symbolic evaluation, we use a $\text{threshold} = 0.95$ for drawings with fill colors and $0.92$ for those without fill colors. The higher threshold for filled drawings (i.e., using \texttt{begin\_fill()} and \texttt{end\_fill()} in the code) is due to the typically larger candidate set $\mathcal{P}$ for these drawings. The pixel-wise similarity in filled areas can overshadow differences in non-filled areas, making them harder to detect. Thus, we set a stricter threshold for filled drawings.

\paragraph{Embedding-based comparison.} \label{sec:appendix:implementation:evaluation:embedding} 
This method compares the standardized images $\image{}$ and $\hat{\image{}}$ within the embedding space. This is achieved by extracting image embeddings from both $\image$ and $\hat{\image}$ using a pre-trained image encoder model and then calculating a similarity score between these embeddings using a distance metric. 
During implementation, when comparing two standardized images $\image{}$ and $\hat{\image{}}$, we first resize them to 256x256 pixels, apply a center crop to 224x224 pixels, convert them to tensors, and normalize them using standard ImageNet statistics to ensure consistency and accuracy. Then we extract 512-dimensional feature vectors from these images using the ResNet18 model pre-trained on ImageNet~\citep{DBLP:conf/cvpr/HeZRS16}.\footnote{We use ResNet-18 primarily because it is a widely adopted, well-performing, and lightweight model (with only 11.7 million parameters) that offers a reasonable trade-off between performance and speed.}
The similarity score between these embeddings is computed using the Euclidean distance, normalized to the range $[0,1]$, where a higher score indicates higher similarity in the embedding space. Images are processed in batches of 128 for efficient computation. Any pairs where image processing fails (e.g., empty images or images that are too large) are assigned a similarity score of 0. Finally, we select the optimal $\text{threshold} = 0.95$ because it achieves the best F1 score (see \autoref{fig:precision_recall_embedding}). If the similarity score exceeds $0.95$, the image $\hat{\image}$ is considered a \emph{success} in matching $\image$; otherwise, it is considered a \emph{fail}.

%% file: appendix-figs/fig_example_mutation.tex
% !TEX root =  main.tex
%%%%%%%%%%%%%%%%%%%%%%%%%%%%%%%%%%%%%%%
%%%%%%%%%%%%%%%%%%%%%%%%%%%%%%%%%%%%%%%

\begin{figure}[htbp!]
    \centering
      \begin{subfigure}{0.95\columnwidth}
          \centering
          \includegraphics[width=0.2\textwidth]{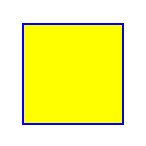}
          \hspace{5em}
          \includegraphics[width=0.15\textwidth]{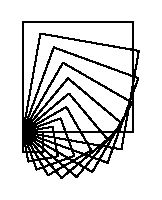}
          \caption{Images for $\code{}_{\text{ref}_1}$ and $\code{}_{\text{ref}_2}$}
      \end{subfigure}
      
      \vspace{1em}
      
      \begin{subfigure}{0.95\columnwidth}
          \centering
          \includegraphics[width=0.2\textwidth]{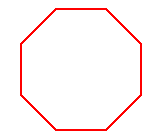}
          \hspace{5em}
          \includegraphics[width=0.15\textwidth]{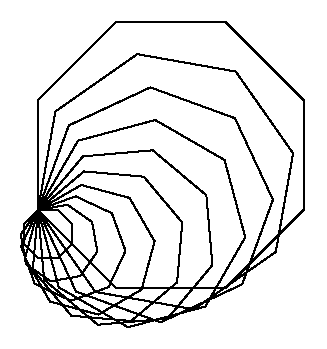}
          \caption{Images for $\code{}_{\text{in}}$ and $\code{}_{\text{out}}$}
      \end{subfigure}

      \vspace{1em}

      \begin{subfigure}{0.95\columnwidth}
          \centering
          \begin{lstlisting}[breaklines=true,breakindent=0pt,xleftmargin=0cm,xrightmargin=0cm,framexleftmargin=0.1cm,framexrightmargin=0.1cm]
- $\textbf{Introduce a loop}$: Add a loop to create multiple instances of a shape with changing parameters.
- $\textbf{Parameterize shape attributes}$: Parameterize attributes such as size, angle, or color to create a dynamic and changing pattern.
- $\textbf{Focus on the pattern}$: Remove or modify existing features to focus on the new pattern created by the loop.
          \end{lstlisting}
          \caption{Inferred mutation pattern $m(\code{}_{\text{ref}_1}, \code{}_{\text{ref}_2}$)}
      \end{subfigure}
    \caption{An illustrative example for the reference-guided code mutation. (a) shows the corresponding images for a pair of sampled reference codes $(\code{}_{\text{ref}_1}, \code{}_{\text{ref}_2})$. (b) shows the corresponding images for $\code{}_{\text{in}}$ and the corresponding mutated code $\code{}_{\text{out}}$ by the LLM's inferred mutation pattern in (c).}
    \label{fig:examples.code_mutation}
\end{figure} 

%% file: appendix-figs/fig_model_version.tex
% !TEX root =  main.tex
%%%%%%%%%%%%%%%%%%%%%%%%%%%%%%%%%%%%%%%%%%%%%%%
%%%%%%%%%%%%%%%%%%%%%%%%%%%%%%%%%%%%%%%%%%%%%%%

\begin{table*}[t!]
    \centering
    \scalebox{0.9}{
    \begin{tabular}{ll}
      \toprule
      \textbf{Model} & \textbf{Version} \\
      \midrule
      o3 & o3-2025-04-16 (\texttt{reasoning\_effort=medium})~\citep{o3_and_o4_mini} \\
      o4-mini & o4-mini-2025-04-16 (\texttt{reasoning\_effort=medium})~\citep{o3_and_o4_mini} \\
      GPT-5 (medium) & gpt-5-2025-08-07 (\texttt{reasoning\_effort=medium})~\citep{gpt5} \\
      \midrule
      \gptfouro{} & gpt-4o-2024-11-20~\citep{gpt4o} \\
      \gptfourv{} & gpt-4-turbo-2024-04-09~\citep{gpt4v} \\
      \midrule
      Qwen3-VL-30B & Qwen/Qwen3-VL-30B-A3B-Instruct~\citep{bai2025qwen3vltechnicalreport} \\
      Qwen2.5-VL-72B & Qwen/Qwen2.5-VL-72B-Instruct~\citep{DBLP:journals/corr/abs-2502-13923} \\
      \qwentwovl{}-72B & Qwen/Qwen2-VL-72B-Instruct~\citep{DBLP:journals/corr/abs-2409-12191} \\
      \qwentwovl{}-7B & Qwen/Qwen2-VL-7B-Instruct~\citep{DBLP:journals/corr/abs-2409-12191} \\
      \qwentwovl{}-72B-\ft{} & Qwen2-VL-72B-Instruct (fine-tuned on \datasetTrain{}) \\
      \qwentwovl{}-7B-\ft{} & Qwen2-VL-7B-Instruct (fine-tuned on \datasetTrain{}) \\
      \midrule
      \molmo{}-72B & allenai/Molmo-72B-0924~\citep{DBLP:journals/corr/abs-2409-17146} \\
      \molmo{}-7B & allenai/Molmo-7B-D-0924~\citep{DBLP:journals/corr/abs-2409-17146} \\
      \midrule
      \llavaov{}-72B & llava-hf/llava-onevision-qwen2-72b-ov-chat-hf~\citep{DBLP:journals/corr/abs-2408-03326} \\
      \llavaov{}-7B & llava-hf/llava-onevision-qwen2-7b-ov-chat-hf~\citep{DBLP:journals/corr/abs-2408-03326} \\
      \midrule
      \nvlmd{} & nvidia/NVLM-D-72B~\citep{DBLP:journals/corr/abs-2409-11402} \\
      \midrule
      \pixtral{}-Large & mistralai/Pixtral-Large-Instruct-2411~\citep{agrawal2024pixtral12b} \\
      \pixtral{}-12B & mistral-community/pixtral-12b~\citep{agrawal2024pixtral12b} \\
      \pixtral{}-12B-\ft{} & Pixtral-12B (fine-tuned on \datasetTrain{}) \\
      \midrule
      InternVL3-76B & OpenGVLab/InternVL3-78B~\citep{DBLP:journals/corr/abs-2504-10479} \\
      \internvltwo{}-76B & OpenGVLab/InternVL2-Llama3-76B~\citep{DBLP:journals/corr/abs-2312-14238} \\
      \internvltwo{}-8B & OpenGVLab/InternVL2-8B~\citep{DBLP:journals/corr/abs-2312-14238} \\
      \midrule
      \glmfourv{}-9B & THUDM/glm-4v-9b~\citep{DBLP:journals/corr/abs-2406-12793} \\
      \bottomrule
    \end{tabular}
    }
    \caption{Evaluated models and their versions.}
    \label{fig:vlm_versions}
\end{table*}

%% file: appendix_case_study.tex
% !TEX root =  main.tex
%%%%%%%%%%%%%%%%%%%%%%%%%%%%%%%%%%%%%%%%%%%%%%%
%%%%%%%%%%%%%%%%%%%%%%%%%%%%%%%%%%%%%%%%%%%%%%%

%%%%%%%%%%%%%%%%%%%%%%%%%%%%%%%%%%%%%%%%% 

\section{Case Study of Failures} \label{sec:appendix:case_study}

We provide a case study of different models' outputs on tasks in the \datasetAll{} dataset. We select five representative VLMs: \gptfouro{}, \pixtral{}-12B, \qwentwovl{}-72B, \pixtral{}-12B-\ft{}, and \qwentwovl{}-72B-\ft{}. To systematically analyze the failure cases, we enumerate all possible failure cases across different types of models and provide examples for each type. 

Specifically, we categorize these $5$ models into $3$ types: proprietary model (\gptfouro{}), open-source base models (i.e., \pixtral{}-12B, \qwentwovl{}-72B), and fine-tuned models (\pixtral{}-12B-\ft{}, \qwentwovl{}-72B-\ft{}). Then we categorize the failure cases into $8$ different possibilities and show examples for each possibility. These possibilities are summarized in \autoref{fig:summary_cases}. \autoref{fig:failure_examples_code_examples1} and \autoref{fig:failure_examples_code_examples2} show example model responses.

\begin{table*}
\centering
\scalebox{1}{
\begin{tabular}{lcccrr}
\toprule
& \gptfouro{} & \begin{tabular}{@{}c@{}} \pixtral{}-12B \\ \qwentwovl{}-72B \end{tabular} & \begin{tabular}{@{}c@{}} \pixtral{}-12B-\ft{} \\ \qwentwovl{}-72B-\ft{} \end{tabular} & \# Cases & Percentage \\
\midrule
\autoref{fig:failure_examples_all_success} & $\successmark$ & $\successmark$ & $\successmark$ & 6 & 0.73\% \\
\midrule
N.A. & $\failmark$ & $\successmark$ & $\successmark$ & 0 & 0.00\% \\
\autoref{fig:failure_examples_all_success_except_base_opensource} & $\successmark$ & $\failmark$ & $\successmark$ & 17 & 2.06\% \\
\autoref{fig:failure_examples_all_success_except_ft} & $\successmark$ & $\successmark$ & $\failmark$ & 3 & 0.36\% \\
\midrule
\autoref{fig:failure_examples_only_gpt4o} & $\successmark$ & $\failmark$ & $\failmark$ & 23 & 2.79\% \\
N.A. & $\failmark$ & $\successmark$ & $\failmark$ & 0 & 0.00\% \\
\autoref{fig:failure_examples_only_ft} & $\failmark$ & $\failmark$ & $\successmark$ & 54 & 6.56\% \\
\midrule
\autoref{fig:failure_examples_all_fail} & $\failmark$ & $\failmark$ & $\failmark$ & 591 & 71.81\% \\
\bottomrule
\end{tabular}
}
\caption{Summary of different possible failure cases across different types of models on our benchmark tasks. We identify eight possible failure cases for the \gptfouro{}, open-source base models (\pixtral{}-12B and \qwentwovl{}-72B), and fine-tuned models (\pixtral{}-12B-\ft{} and \qwentwovl{}-72B-\ft{}). For each possible failure case, we provide the number of occurrences and the corresponding percentage, with references to detailed examples in corresponding figures. Success and failure are indicated by $\successmark$ and $\failmark$, respectively.}
\label{fig:summary_cases}
\end{table*} 

%% file: appendix_prompt.tex
% !TEX root =  main.tex
%%%%%%%%%%%%%%%%%%%%%%%%%%%%%%%%%%%%%%%%%%%%%%%
%%%%%%%%%%%%%%%%%%%%%%%%%%%%%%%%%%%%%%%%%%%%%%%

%%%%%%%%%%%%%%%%%%%%%%%%%%%%%%%%%%%%%%%%% 

\section{Prompts} \label{sec:appendix:prompt}

In this section, we provide the prompts used in \benchmark{} as follows:
\begin{itemize}
    \item \autoref{fig:appendix:prompt_template:code_synthesis} shows the prompt used for guiding VLMs in synthesizing code from a given image input.
    \item \autoref{fig:prompt_template_one_shot_code_generation.example1} shows the prompt for the reference-guided code generation stage in our data synthesis framework \datagenOurs{}.
    \item \autoref{fig:prompt_template_eval_quality} provides the prompt used for the elite selection stage in \datagenOurs{} for scoring the quality of the generated geometric image. 
    \item \autoref{fig:prompt_template_code_relabel} provides the prompt for the CoT labeling stage in \datagenOurs{}.
\end{itemize}

\input{appendix-figs/fig_failure_cases}

\input{appendix-figs/fig_prompt_code_synthesis}

\input{appendix-figs/fig_prompt_one_shot_code_generation}

\input{appendix-figs/fig_prompt_eval_quality}

\input{appendix-figs/fig_prompt_code_relabel}

%% file: appendix-figs/fig_failure_cases.tex
% !TEX root =  main.tex
%%%%%%%%%%%%%%%%%%%%%%%%%%%%%%%%%%%%%%%%%%%%%%%
%%%%%%%%%%%%%%%%%%%%%%%%%%%%%%%%%%%%%%%%%%%%%%%

\input{appendix-figs/fig_failure_cases_all_success.tex}

\input{appendix-figs/fig_failure_cases_all_success_except_base_opensource.tex}

\input{appendix-figs/fig_failure_cases_all_success_except_ft.tex}

\input{appendix-figs/fig_failure_cases_only_gpt4o.tex}

\input{appendix-figs/fig_failure_cases_only_ft.tex}

\input{appendix-figs/fig_failure_cases_all_fail.tex}

\input{appendix-figs/fig_failure_cases_code_examples.tex}

%% file: appendix-figs/fig_failure_cases_all_success.tex
% !TEX root =  main.tex
%%%%%%%%%%%%%%%%%%%%%%%%%%%%%%%%%%%%%%%%%%%%%%%
%%%%%%%%%%%%%%%%%%%%%%%%%%%%%%%%%%%%%%%%%%%%%%%

\begin{table*}[t]
  \centering
  \scalebox{0.8}{
  \begin{tabular}{c|ccccc}
    \toprule
    Input Image & \gptfouro & \pixtral{}-12B & \qwentwovl{}-72B & \pixtral{}-12B-\ft{} & \qwentwovl{}-72B-\ft{} \\
    \midrule
    \midrule
    \begin{tabular}{c}
      \parbox[t][1.8cm][c]{0.1\textwidth}{\includegraphics[width=0.1\textwidth]{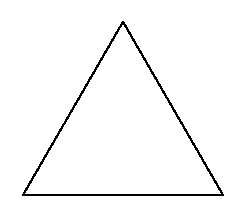}} \\
      $$
    \end{tabular} & 
    \begin{tabular}{c}
      \parbox[t][1.8cm][c]{0.1\textwidth}{\includegraphics[width=0.1\textwidth]{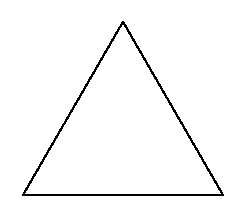}} \\
      $\successmark$
    \end{tabular} &
    \begin{tabular}{c}
      \parbox[t][1.8cm][c]{0.1\textwidth}{\includegraphics[width=0.1\textwidth]{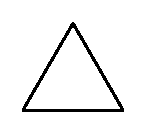}} \\
      $\successmark$
    \end{tabular} &
    \begin{tabular}{c}
      \parbox[t][1.8cm][c]{0.1\textwidth}{\includegraphics[width=0.1\textwidth]{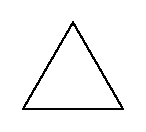}} \\
      $\successmark$
    \end{tabular} &
    \begin{tabular}{c}
      \parbox[t][1.8cm][c]{0.1\textwidth}{\includegraphics[width=0.1\textwidth]{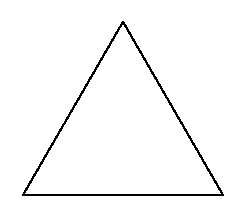}} \\
      $\successmark$
    \end{tabular} &
    \begin{tabular}{c}
      \parbox[t][1.8cm][c]{0.1\textwidth}{\includegraphics[width=0.1\textwidth]{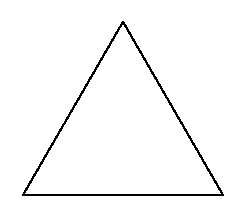}} \\
      $\successmark$
    \end{tabular} \\
    \midrule
    \begin{tabular}{c}
      \parbox[t][1.8cm][c]{0.1\textwidth}{\includegraphics[width=0.1\textwidth]{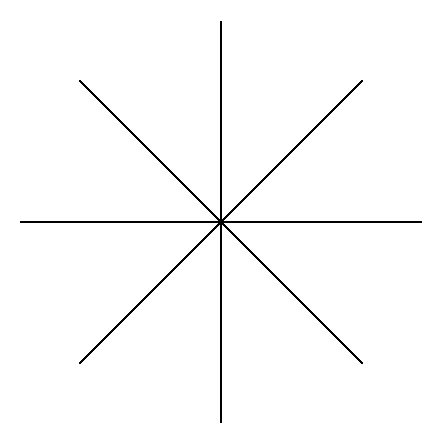}} \\
      $$
    \end{tabular} & 
    \begin{tabular}{c}
      \parbox[t][1.8cm][c]{0.1\textwidth}{\includegraphics[width=0.1\textwidth]{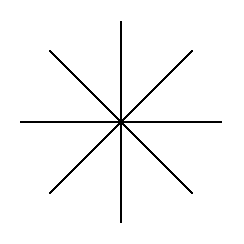}} \\
      $\successmark$
    \end{tabular} &
    \begin{tabular}{c}
      \parbox[t][1.8cm][c]{0.1\textwidth}{\includegraphics[width=0.1\textwidth]{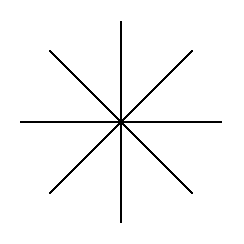}} \\
      $\successmark$
    \end{tabular} &
    \begin{tabular}{c}
      \parbox[t][1.8cm][c]{0.1\textwidth}{\includegraphics[width=0.1\textwidth]{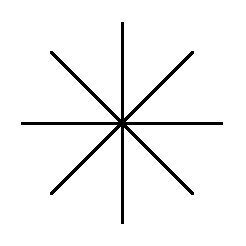}} \\
      $\successmark$
    \end{tabular} &
    \begin{tabular}{c}
      \parbox[t][1.8cm][c]{0.1\textwidth}{\includegraphics[width=0.1\textwidth]{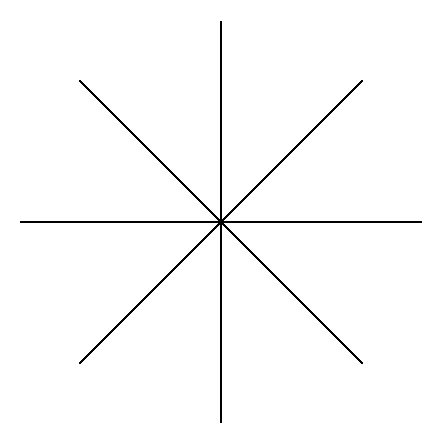}} \\
      $\successmark$
    \end{tabular} &
    \begin{tabular}{c}
      \parbox[t][1.8cm][c]{0.1\textwidth}{\includegraphics[width=0.1\textwidth]{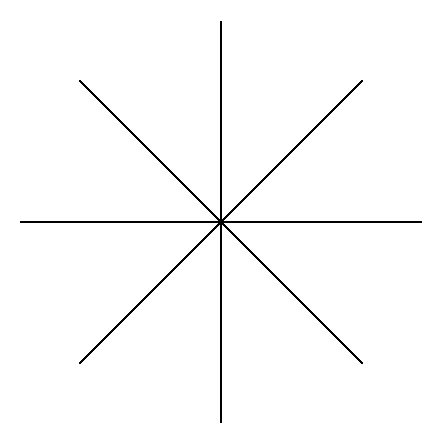}} \\
      $\successmark$
    \end{tabular} \\
    \midrule
    \begin{tabular}{c}
      \parbox[t][1.8cm][c]{0.1\textwidth}{\includegraphics[width=0.1\textwidth]{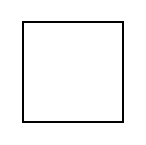}} \\
      $$
    \end{tabular} & 
    \begin{tabular}{c}
      \parbox[t][1.8cm][c]{0.1\textwidth}{\includegraphics[width=0.1\textwidth]{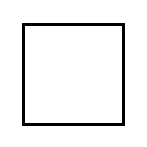}} \\
      $\successmark$
    \end{tabular} &
    \begin{tabular}{c}
      \parbox[t][1.8cm][c]{0.1\textwidth}{\includegraphics[width=0.1\textwidth]{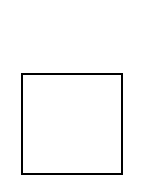}} \\
      $\successmark$
    \end{tabular} &
    \begin{tabular}{c}
      \parbox[t][1.8cm][c]{0.1\textwidth}{\includegraphics[width=0.1\textwidth]{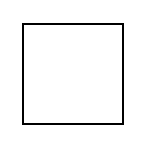}} \\
      $\successmark$
    \end{tabular} &
    \begin{tabular}{c}
      \parbox[t][1.8cm][c]{0.1\textwidth}{\includegraphics[width=0.1\textwidth]{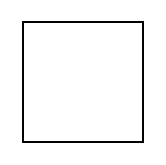}} \\
      $\successmark$
    \end{tabular} &
    \begin{tabular}{c}
      \parbox[t][1.8cm][c]{0.1\textwidth}{\includegraphics[width=0.1\textwidth]{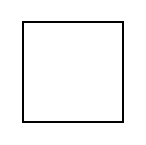}} \\
      $\successmark$
    \end{tabular} \\
    \midrule
    \begin{tabular}{c}
      \parbox[t][1.8cm][c]{0.1\textwidth}{\includegraphics[width=0.1\textwidth]{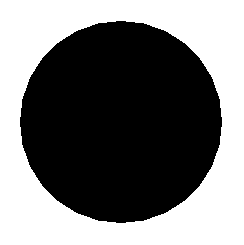}} \\
      $$
    \end{tabular} & 
    \begin{tabular}{c}
      \parbox[t][1.8cm][c]{0.1\textwidth}{\includegraphics[width=0.1\textwidth]{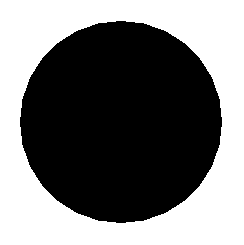}} \\
      $\successmark$
    \end{tabular} &
    \begin{tabular}{c}
      \parbox[t][1.8cm][c]{0.1\textwidth}{\includegraphics[width=0.1\textwidth]{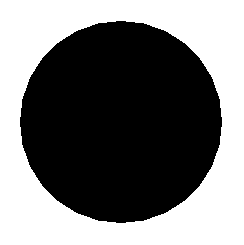}} \\
      $\successmark$
    \end{tabular} &
    \begin{tabular}{c}
      \parbox[t][1.8cm][c]{0.1\textwidth}{\includegraphics[width=0.1\textwidth]{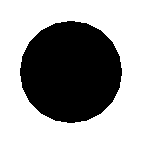}} \\
      $\successmark$
    \end{tabular} &
    \begin{tabular}{c}
      \parbox[t][1.8cm][c]{0.1\textwidth}{\includegraphics[width=0.1\textwidth]{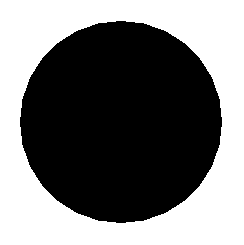}} \\
      $\successmark$
    \end{tabular} &
    \begin{tabular}{c}
      \parbox[t][1.8cm][c]{0.1\textwidth}{\includegraphics[width=0.1\textwidth]{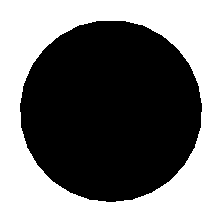}} \\
      $\successmark$
    \end{tabular} \\
    \midrule
    \begin{tabular}{c}
      \parbox[t][1.8cm][c]{0.1\textwidth}{\includegraphics[width=0.1\textwidth]{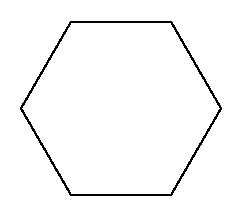}} \\
      $$
    \end{tabular} & 
    \begin{tabular}{c}
      \parbox[t][1.8cm][c]{0.1\textwidth}{\includegraphics[width=0.1\textwidth]{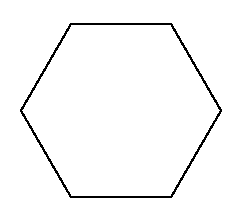}} \\
      $\successmark$
    \end{tabular} &
    \begin{tabular}{c}
      \parbox[t][1.8cm][c]{0.1\textwidth}{\includegraphics[width=0.1\textwidth]{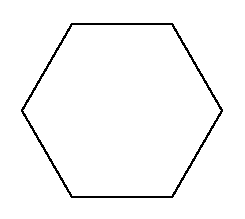}} \\
      $\successmark$
    \end{tabular} &
    \begin{tabular}{c}
      \parbox[t][1.8cm][c]{0.1\textwidth}{\includegraphics[width=0.1\textwidth]{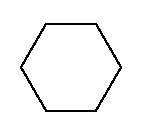}} \\
      $\successmark$
    \end{tabular} &
    \begin{tabular}{c}
      \parbox[t][1.8cm][c]{0.1\textwidth}{\includegraphics[width=0.1\textwidth]{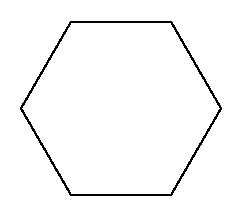}} \\
      $\successmark$
    \end{tabular} &
    \begin{tabular}{c}
      \parbox[t][1.8cm][c]{0.1\textwidth}{\includegraphics[width=0.1\textwidth]{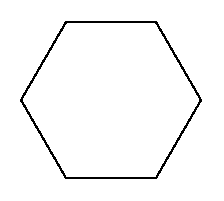}} \\
      $\successmark$
    \end{tabular} \\
    \midrule
    \begin{tabular}{c}
      \parbox[t][1.8cm][c]{0.1\textwidth}{\includegraphics[width=0.1\textwidth]{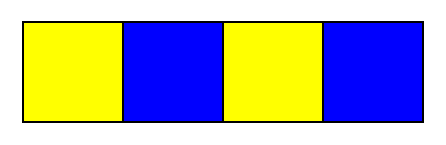}} \\
      $$
    \end{tabular} & 
    \begin{tabular}{c}
      \parbox[t][1.8cm][c]{0.1\textwidth}{\includegraphics[width=0.1\textwidth]{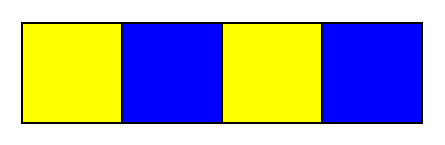}} \\
      $\successmark$
    \end{tabular} &
    \begin{tabular}{c}
      \parbox[t][1.8cm][c]{0.1\textwidth}{\includegraphics[width=0.1\textwidth]{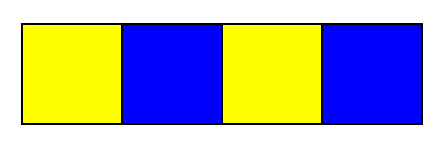}} \\
      $\successmark$
    \end{tabular} &
    \begin{tabular}{c}
      \parbox[t][1.8cm][c]{0.1\textwidth}{\includegraphics[width=0.1\textwidth]{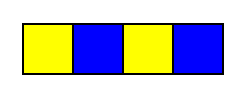}} \\
      $\successmark$
    \end{tabular} &
    \begin{tabular}{c}
      \parbox[t][1.8cm][c]{0.1\textwidth}{\includegraphics[width=0.1\textwidth]{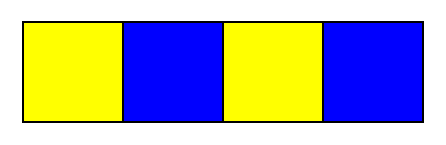}} \\
      $\successmark$
    \end{tabular} &
    \begin{tabular}{c}
      \parbox[t][1.8cm][c]{0.1\textwidth}{\includegraphics[width=0.1\textwidth]{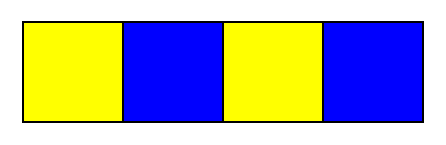}} \\
      $\successmark$
    \end{tabular} \\

    \bottomrule
  \end{tabular}
  }
  \caption{Tasks that are successfully solved by base models (i.e., \gptfouro{}, \pixtral{}-12B, \qwentwovl{}-72B) and our fine-tuned models (i.e., \pixtral{}-12B-\ft{}, \qwentwovl{}-72B-\ft{}) in the \datasetAll{} dataset with $823$ tasks. A total of $6$ tasks (0.73\%) match this criteria.
  Each row shows a ground truth image (leftmost) followed by the corresponding images generated by executing the each model's generated Python code. Success ($\successmark$) and failure ($\failmark$) are determined by our evaluation framework using symbolic comparison.
  }
  \label{fig:failure_examples_all_success}
\end{table*}

%% file: appendix-figs/fig_failure_cases_all_success_except_base_opensource.tex
% !TEX root =  main.tex
%%%%%%%%%%%%%%%%%%%%%%%%%%%%%%%%%%%%%%%%%%%%%%%
%%%%%%%%%%%%%%%%%%%%%%%%%%%%%%%%%%%%%%%%%%%%%%%

\begin{table*}[t]
  \centering
  \scalebox{0.8}{
  \begin{tabular}{c|ccccc}
    \toprule
    Input Image & \gptfouro & \pixtral{}-12B & \qwentwovl{}-72B & \pixtral{}-12B-\ft{} & \qwentwovl{}-72B-\ft{} \\
    \midrule
    \midrule
    \begin{tabular}{c}
      \parbox[t][1.8cm][c]{0.1\textwidth}{\includegraphics[width=0.1\textwidth]{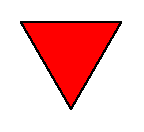}} \\
      $$
    \end{tabular} & 
    \begin{tabular}{c}
      \parbox[t][1.8cm][c]{0.1\textwidth}{\includegraphics[width=0.1\textwidth]{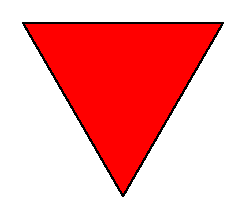}} \\
      $\successmark$
    \end{tabular} &
    \begin{tabular}{c}
      \parbox[t][1.8cm][c]{0.1\textwidth}{\includegraphics[width=0.1\textwidth]{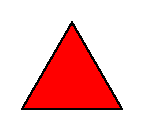}} \\
      $\failmark$
    \end{tabular} &
    \begin{tabular}{c}
      \parbox[t][1.8cm][c]{0.1\textwidth}{\includegraphics[width=0.1\textwidth]{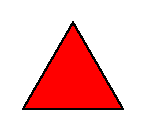}} \\
      $\failmark$
    \end{tabular} &
    \begin{tabular}{c}
      \parbox[t][1.8cm][c]{0.1\textwidth}{\includegraphics[width=0.1\textwidth]{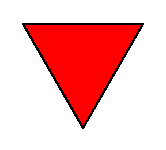}} \\
      $\successmark$
    \end{tabular} &
    \begin{tabular}{c}
      \parbox[t][1.8cm][c]{0.1\textwidth}{\includegraphics[width=0.1\textwidth]{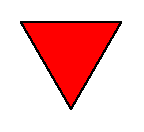}} \\
      $\successmark$
    \end{tabular} \\
    \midrule
    \begin{tabular}{c}
      \parbox[t][1.8cm][c]{0.1\textwidth}{\includegraphics[width=0.1\textwidth]{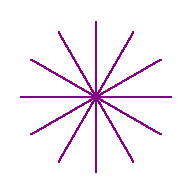}} \\
      $$
    \end{tabular} & 
    \begin{tabular}{c}
      \parbox[t][1.8cm][c]{0.1\textwidth}{\includegraphics[width=0.1\textwidth]{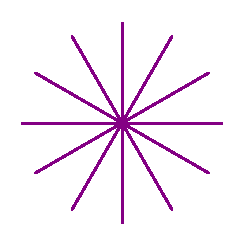}} \\
      $\successmark$
    \end{tabular} &
    \begin{tabular}{c}
      \parbox[t][1.8cm][c]{0.1\textwidth}{\includegraphics[width=0.1\textwidth]{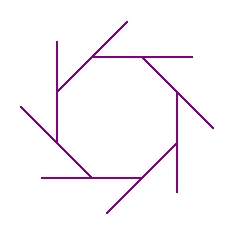}} \\
      $\failmark$
    \end{tabular} &
    \begin{tabular}{c}
      \parbox[t][1.8cm][c]{0.1\textwidth}{\includegraphics[width=0.1\textwidth]{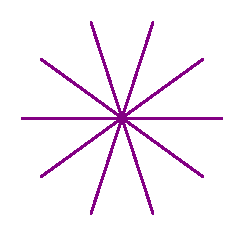}} \\
      $\failmark$
    \end{tabular} &
    \begin{tabular}{c}
      \parbox[t][1.8cm][c]{0.1\textwidth}{\includegraphics[width=0.1\textwidth]{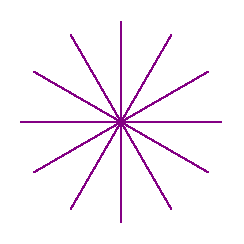}} \\
      $\successmark$
    \end{tabular} &
    \begin{tabular}{c}
      \parbox[t][1.8cm][c]{0.1\textwidth}{\includegraphics[width=0.1\textwidth]{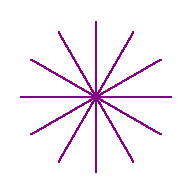}} \\
      $\successmark$
    \end{tabular} \\
    \midrule
    \begin{tabular}{c}
      \parbox[t][1.8cm][c]{0.1\textwidth}{\includegraphics[width=0.1\textwidth]{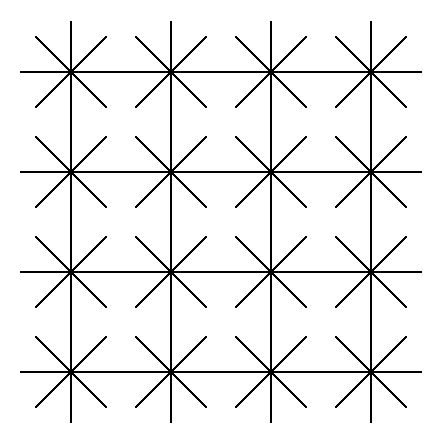}} \\
      $$
    \end{tabular} & 
    \begin{tabular}{c}
      \parbox[t][1.8cm][c]{0.1\textwidth}{\includegraphics[width=0.1\textwidth]{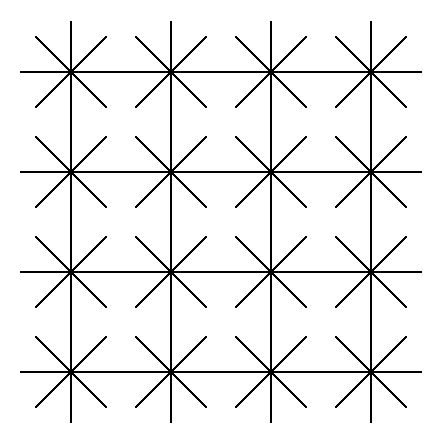}} \\
      $\successmark$
    \end{tabular} &
    \begin{tabular}{c}
      \parbox[t][1.8cm][c]{0.1\textwidth}{\includegraphics[width=0.1\textwidth]{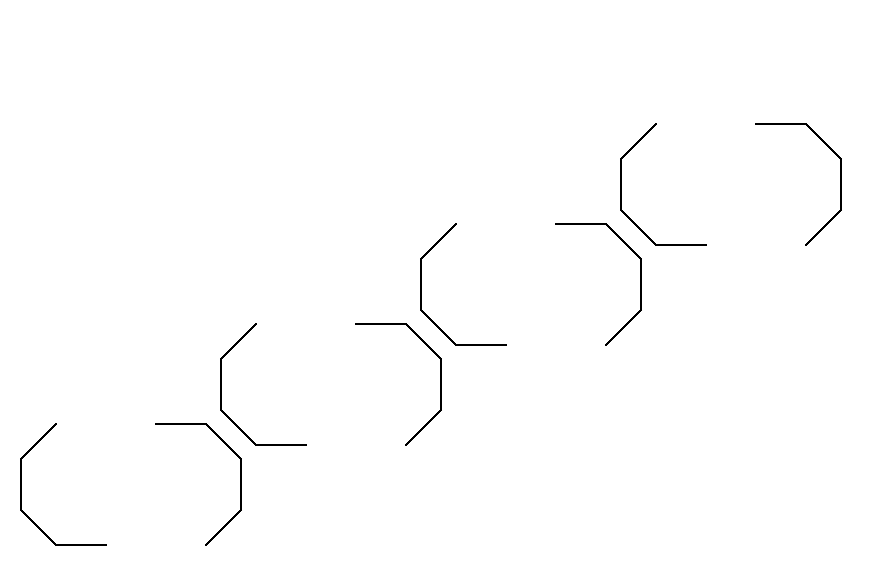}} \\
      $\failmark$
    \end{tabular} &
    \begin{tabular}{c}
      \parbox[t][1.8cm][c]{0.1\textwidth}{\includegraphics[width=0.1\textwidth]{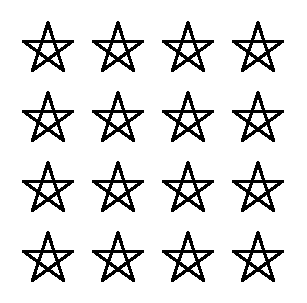}} \\
      $\failmark$
    \end{tabular} &
    \begin{tabular}{c}
      \parbox[t][1.8cm][c]{0.1\textwidth}{\includegraphics[width=0.1\textwidth]{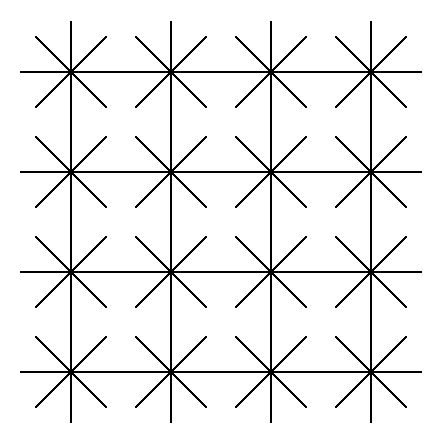}} \\
      $\successmark$
    \end{tabular} &
    \begin{tabular}{c}
      \parbox[t][1.8cm][c]{0.1\textwidth}{\includegraphics[width=0.1\textwidth]{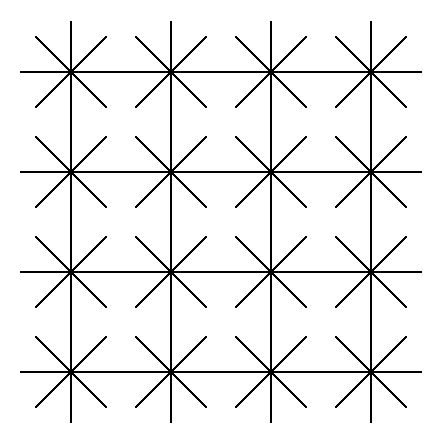}} \\
      $\successmark$
    \end{tabular} \\
    \midrule
    \begin{tabular}{c}
      \parbox[t][1.8cm][c]{0.1\textwidth}{\includegraphics[width=0.1\textwidth]{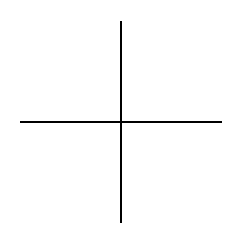}} \\
      $$
    \end{tabular} & 
    \begin{tabular}{c}
      \parbox[t][1.8cm][c]{0.1\textwidth}{\includegraphics[width=0.1\textwidth]{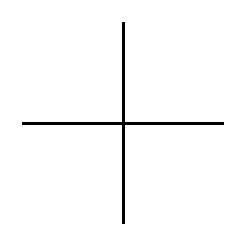}} \\
      $\successmark$
    \end{tabular} &
    \begin{tabular}{c}
      \parbox[t][1.8cm][c]{0.1\textwidth}{\includegraphics[width=0.1\textwidth]{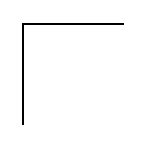}} \\
      $\failmark$
    \end{tabular} &
    \begin{tabular}{c}
      \parbox[t][1.8cm][c]{0.1\textwidth}{\includegraphics[width=0.1\textwidth]{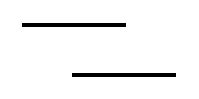}} \\
      $\failmark$
    \end{tabular} &
    \begin{tabular}{c}
      \parbox[t][1.8cm][c]{0.1\textwidth}{\includegraphics[width=0.1\textwidth]{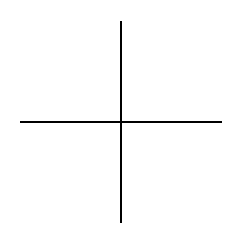}} \\
      $\successmark$
    \end{tabular} &
    \begin{tabular}{c}
      \parbox[t][1.8cm][c]{0.1\textwidth}{\includegraphics[width=0.1\textwidth]{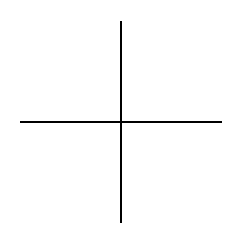}} \\
      $\successmark$
    \end{tabular} \\
    \midrule
    \begin{tabular}{c}
      \parbox[t][1.8cm][c]{0.1\textwidth}{\includegraphics[width=0.1\textwidth]{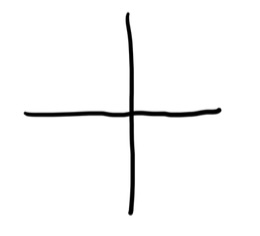}} \\
      $$
    \end{tabular} & 
    \begin{tabular}{c}
      \parbox[t][1.8cm][c]{0.1\textwidth}{\includegraphics[width=0.1\textwidth]{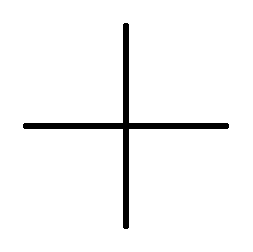}} \\
      $\successmark$
    \end{tabular} &
    \begin{tabular}{c}
      \parbox[t][1.8cm][c]{0.1\textwidth}{\includegraphics[width=0.1\textwidth]{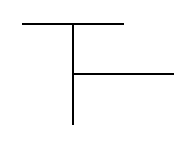}} \\
      $\failmark$
    \end{tabular} &
    \begin{tabular}{c}
      \parbox[t][1.8cm][c]{0.1\textwidth}{\includegraphics[width=0.1\textwidth]{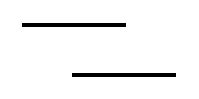}} \\
      $\failmark$
    \end{tabular} &
    \begin{tabular}{c}
      \parbox[t][1.8cm][c]{0.1\textwidth}{\includegraphics[width=0.1\textwidth]{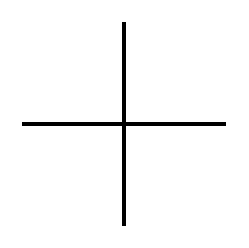}} \\
      $\successmark$
    \end{tabular} &
    \begin{tabular}{c}
      \parbox[t][1.8cm][c]{0.1\textwidth}{\includegraphics[width=0.1\textwidth]{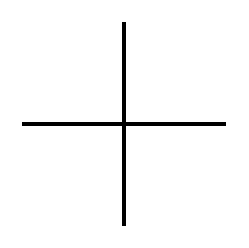}} \\
      $\successmark$
    \end{tabular} \\
    \midrule
    \begin{tabular}{c}
      \parbox[t][1.8cm][c]{0.1\textwidth}{\includegraphics[width=0.1\textwidth]{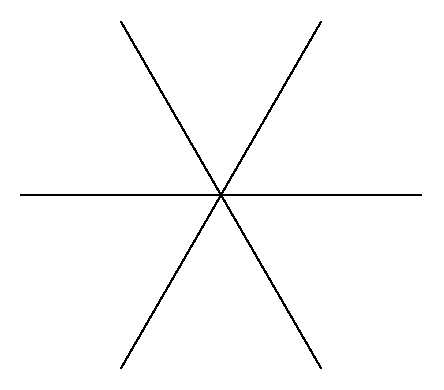}} \\
      $$
    \end{tabular} & 
    \begin{tabular}{c}
      \parbox[t][1.8cm][c]{0.1\textwidth}{\includegraphics[width=0.1\textwidth]{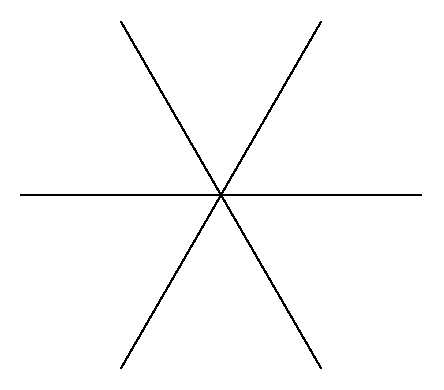}} \\
      $\successmark$
    \end{tabular} &
    \begin{tabular}{c}
      \parbox[t][1.8cm][c]{0.1\textwidth}{\includegraphics[width=0.1\textwidth]{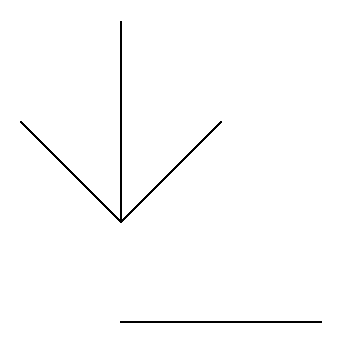}} \\
      $\failmark$
    \end{tabular} &
    \begin{tabular}{c}
      \parbox[t][1.8cm][c]{0.1\textwidth}{\includegraphics[width=0.1\textwidth]{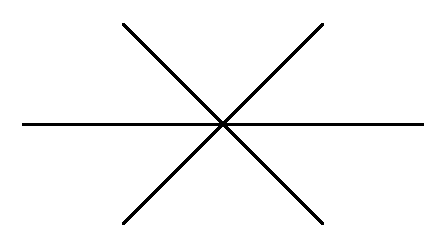}} \\
      $\failmark$
    \end{tabular} &
    \begin{tabular}{c}
      \parbox[t][1.8cm][c]{0.1\textwidth}{\includegraphics[width=0.1\textwidth]{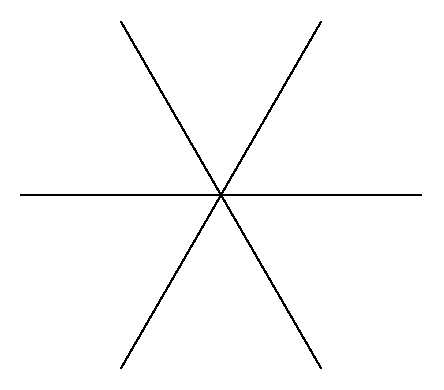}} \\
      $\successmark$
    \end{tabular} &
    \begin{tabular}{c}
      \parbox[t][1.8cm][c]{0.1\textwidth}{\includegraphics[width=0.1\textwidth]{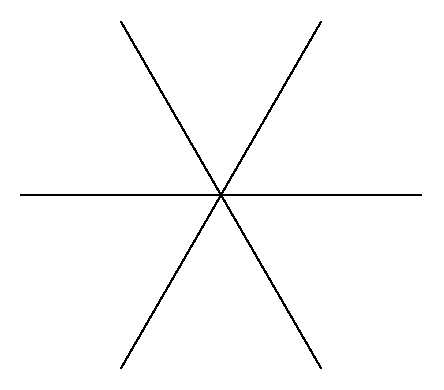}} \\
      $\successmark$
    \end{tabular} \\
    \midrule
    \begin{tabular}{c}
      \parbox[t][1.8cm][c]{0.1\textwidth}{\includegraphics[width=0.1\textwidth]{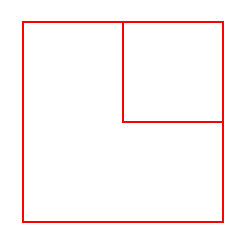}} \\
      $$
    \end{tabular} & 
    \begin{tabular}{c}
      \parbox[t][1.8cm][c]{0.1\textwidth}{\includegraphics[width=0.1\textwidth]{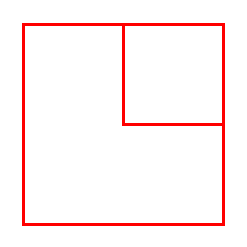}} \\
      $\successmark$
    \end{tabular} &
    \begin{tabular}{c}
      \parbox[t][1.8cm][c]{0.1\textwidth}{\includegraphics[width=0.08\textwidth]{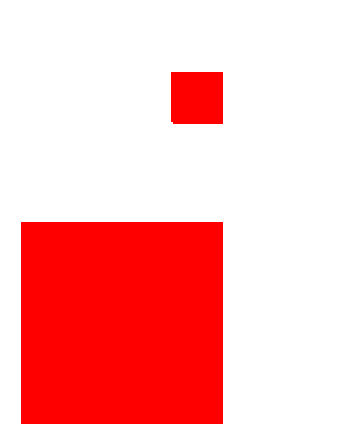}} \\
      $\failmark$
    \end{tabular} &
    \begin{tabular}{c}
      \parbox[t][1.8cm][c]{0.1\textwidth}{\includegraphics[width=0.1\textwidth]{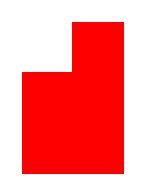}} \\
      $\failmark$
    \end{tabular} &
    \begin{tabular}{c}
      \parbox[t][1.8cm][c]{0.1\textwidth}{\includegraphics[width=0.1\textwidth]{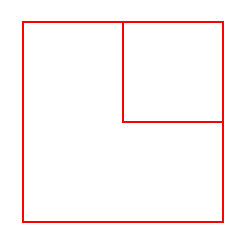}} \\
      $\successmark$
    \end{tabular} &
    \begin{tabular}{c}
      \parbox[t][1.8cm][c]{0.1\textwidth}{\includegraphics[width=0.1\textwidth]{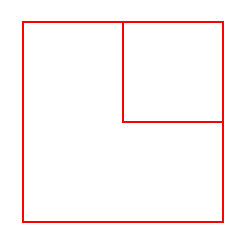}} \\
      $\successmark$
    \end{tabular} \\
    \midrule
    \begin{tabular}{c}
      \parbox[t][1.8cm][c]{0.1\textwidth}{\includegraphics[width=0.1\textwidth]{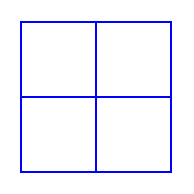}} \\
      $$
    \end{tabular} & 
    \begin{tabular}{c}
      \parbox[t][1.8cm][c]{0.1\textwidth}{\includegraphics[width=0.1\textwidth]{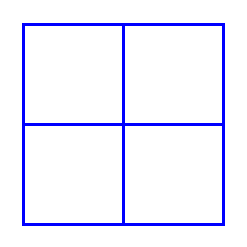}} \\
      $\successmark$
    \end{tabular} &
    \begin{tabular}{c}
      \parbox[t][1.8cm][c]{0.1\textwidth}{\includegraphics[width=0.06\textwidth]{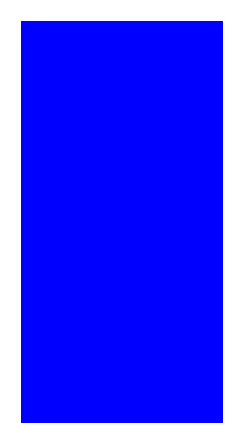}} \\
      $\failmark$
    \end{tabular} &
    \begin{tabular}{c}
      \parbox[t][1.8cm][c]{0.1\textwidth}{\includegraphics[width=0.1\textwidth]{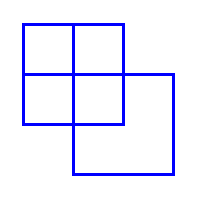}} \\
      $\failmark$
    \end{tabular} &
    \begin{tabular}{c}
      \parbox[t][1.8cm][c]{0.1\textwidth}{\includegraphics[width=0.1\textwidth]{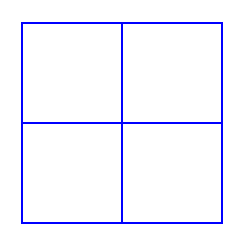}} \\
      $\successmark$
    \end{tabular} &
    \begin{tabular}{c}
      \parbox[t][1.8cm][c]{0.1\textwidth}{\includegraphics[width=0.1\textwidth]{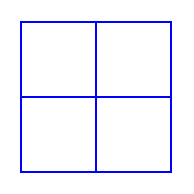}} \\
      $\successmark$
    \end{tabular} \\
    \midrule
    \begin{tabular}{c}
      \parbox[t][1.8cm][c]{0.1\textwidth}{\includegraphics[width=0.1\textwidth]{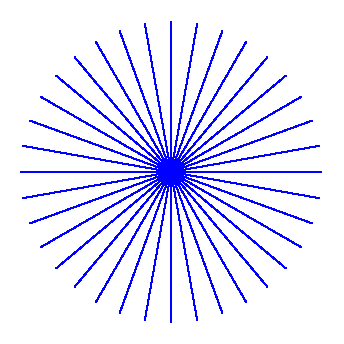}} \\
      $$
    \end{tabular} & 
    \begin{tabular}{c}
      \parbox[t][1.8cm][c]{0.1\textwidth}{\includegraphics[width=0.1\textwidth]{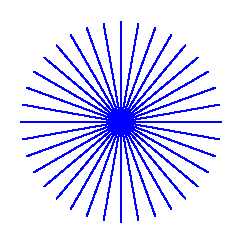}} \\
      $\successmark$
    \end{tabular} &
    \begin{tabular}{c}
      \parbox[t][1.8cm][c]{0.1\textwidth}{\includegraphics[width=0.1\textwidth]{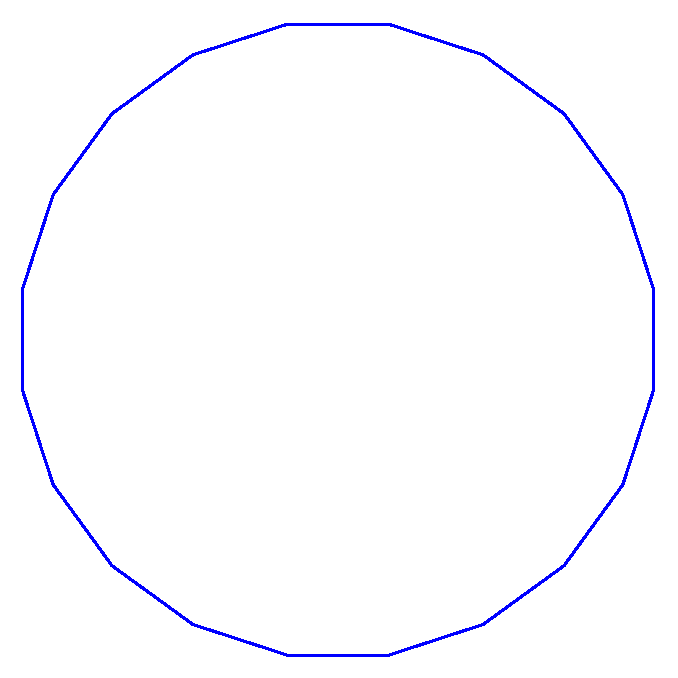}} \\
      $\failmark$
    \end{tabular} &
    \begin{tabular}{c}
      \parbox[t][1.8cm][c]{0.1\textwidth}{\includegraphics[width=0.1\textwidth]{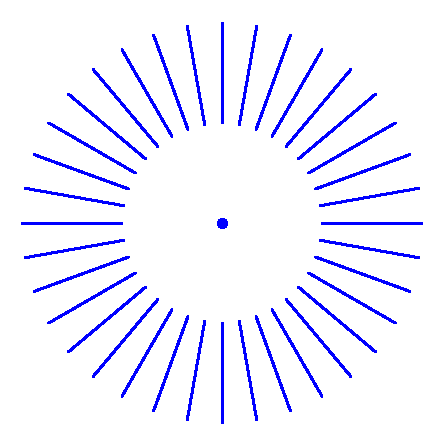}} \\
      $\failmark$
    \end{tabular} &
    \begin{tabular}{c}
      \parbox[t][1.8cm][c]{0.1\textwidth}{\includegraphics[width=0.1\textwidth]{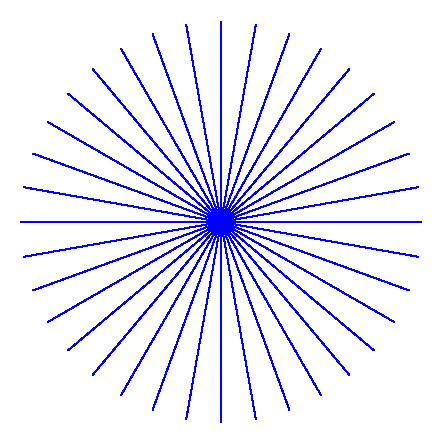}} \\
      $\successmark$
    \end{tabular} &
    \begin{tabular}{c}
      \parbox[t][1.8cm][c]{0.1\textwidth}{\includegraphics[width=0.1\textwidth]{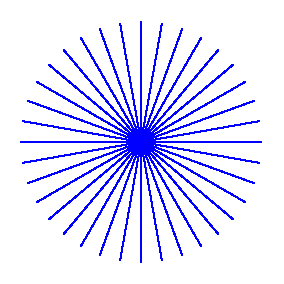}} \\
      $\successmark$
    \end{tabular} \\
    \bottomrule
  \end{tabular}
  }
  \caption{Example tasks that are successfully solved by \gptfouro{} and our fine-tuned models (i.e., \pixtral{}-12B-\ft{}, \qwentwovl{}-72B-\ft{}), but not solved by the base models (\pixtral{}-12B, \qwentwovl{}-72B) in the \datasetAll{} dataset. A total of $17$ tasks (2.06\%) match this criteria.
  Each row shows a ground truth image (leftmost) followed by the corresponding images generated by executing the each model's generated Python code. Success ($\successmark$) and failure ($\failmark$) are determined by our evaluation framework using symbolic comparison. 
  }

  \label{fig:failure_examples_all_success_except_base_opensource}
\end{table*}

%% file: appendix-figs/fig_failure_cases_all_success_except_ft.tex
% !TEX root =  main.tex
%%%%%%%%%%%%%%%%%%%%%%%%%%%%%%%%%%%%%%%%%%%%%%%
%%%%%%%%%%%%%%%%%%%%%%%%%%%%%%%%%%%%%%%%%%%%%%%

\begin{table*}[t]
  \centering
  \scalebox{0.8}{
  \begin{tabular}{c|ccccc}
    \toprule
    Input Image & \gptfouro & \pixtral{}-12B & \qwentwovl{}-72B & \pixtral{}-12B-\ft{} & \qwentwovl{}-72B-\ft{} \\
    \midrule
    \midrule
    \begin{tabular}{c}
      \parbox[t][1.8cm][c]{0.1\textwidth}{\includegraphics[width=0.1\textwidth]{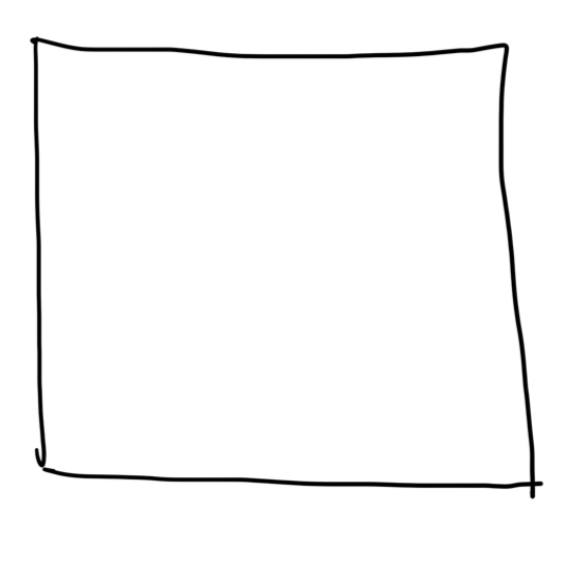}} \\
      $$
    \end{tabular} & 
    \begin{tabular}{c}
      \parbox[t][1.8cm][c]{0.1\textwidth}{\includegraphics[width=0.1\textwidth]{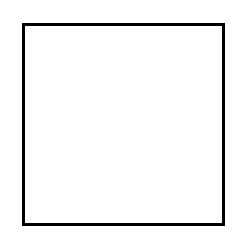}} \\
      $\successmark$
    \end{tabular} &
    \begin{tabular}{c}
      \parbox[t][1.8cm][c]{0.1\textwidth}{\includegraphics[width=0.1\textwidth]{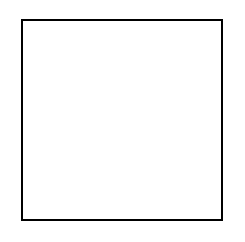}} \\
      $\successmark$
    \end{tabular} &
    \begin{tabular}{c}
      \parbox[t][1.8cm][c]{0.1\textwidth}{\includegraphics[width=0.1\textwidth]{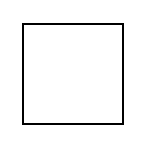}} \\
      $\successmark$
    \end{tabular} &
    \begin{tabular}{c}
      \parbox[t][1.8cm][c]{0.1\textwidth}{\includegraphics[width=0.1\textwidth]{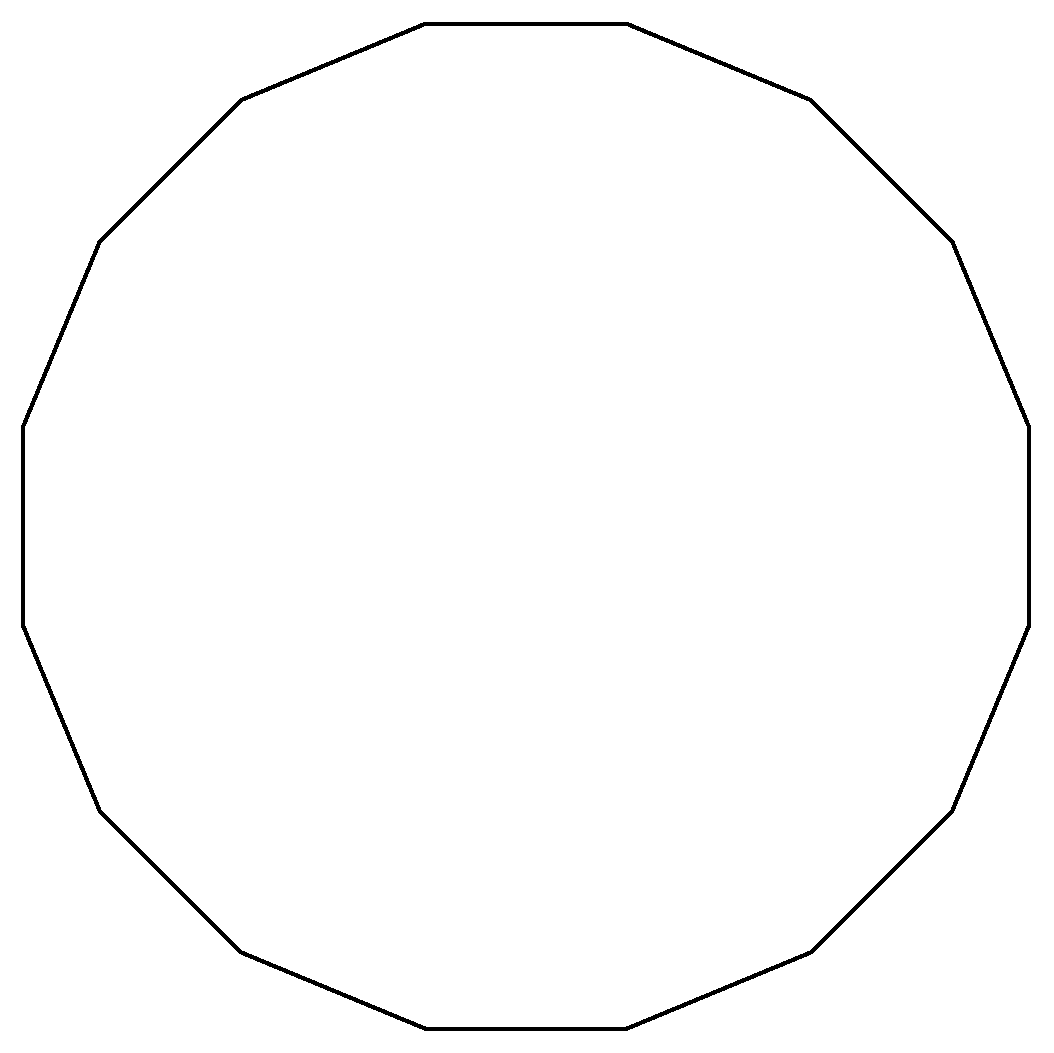}} \\
      $\failmark$
    \end{tabular} &
    \begin{tabular}{c}
      \parbox[t][1.8cm][c]{0.1\textwidth}{\includegraphics[width=0.1\textwidth]{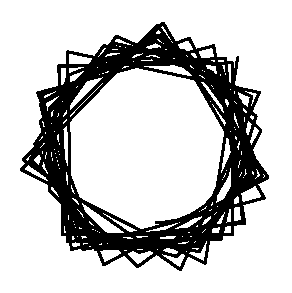}} \\
      $\failmark$
    \end{tabular} \\
    \midrule
    \begin{tabular}{c}
      \parbox[t][1.8cm][c]{0.1\textwidth}{\includegraphics[width=0.1\textwidth]{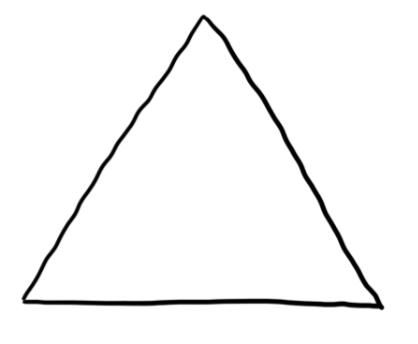}} \\
      $$
    \end{tabular} & 
    \begin{tabular}{c}
      \parbox[t][1.8cm][c]{0.1\textwidth}{\includegraphics[width=0.1\textwidth]{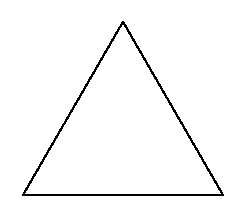}} \\
      $\successmark$
    \end{tabular} &
    \begin{tabular}{c}
      \parbox[t][1.8cm][c]{0.1\textwidth}{\includegraphics[width=0.1\textwidth]{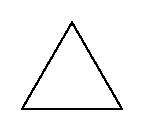}} \\
      $\successmark$
    \end{tabular} &
    \begin{tabular}{c}
      \parbox[t][1.8cm][c]{0.1\textwidth}{\includegraphics[width=0.1\textwidth]{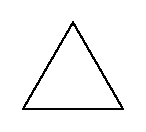}} \\
      $\successmark$
    \end{tabular} &
    \begin{tabular}{c}
      \parbox[t][1.8cm][c]{0.1\textwidth}{\includegraphics[width=0.1\textwidth]{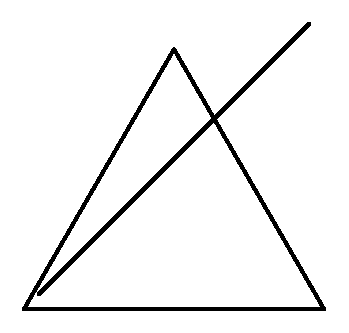}} \\
      $\failmark$
    \end{tabular} &
    \begin{tabular}{c}
      \parbox[t][1.8cm][c]{0.1\textwidth}{\includegraphics[width=0.1\textwidth]{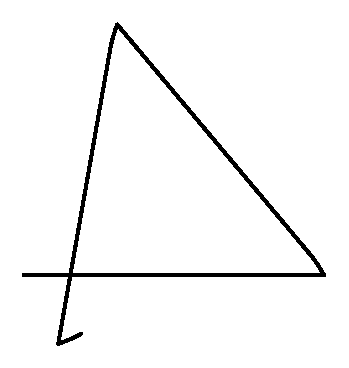}} \\
      $\failmark$
    \end{tabular} \\
    \midrule
    \begin{tabular}{c}
      \parbox[t][1.8cm][c]{0.1\textwidth}{\includegraphics[width=0.1\textwidth]{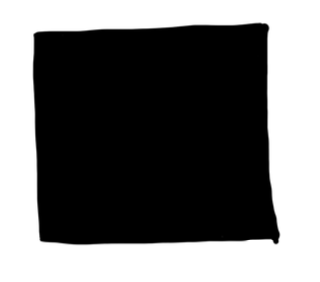}} \\
      $$
    \end{tabular} & 
    \begin{tabular}{c}
      \parbox[t][1.8cm][c]{0.1\textwidth}{\includegraphics[width=0.1\textwidth]{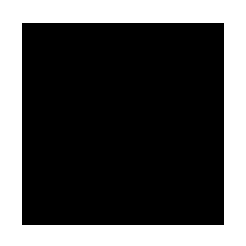}} \\
      $\successmark$
    \end{tabular} &
    \begin{tabular}{c}
      \parbox[t][1.8cm][c]{0.1\textwidth}{\includegraphics[width=0.1\textwidth]{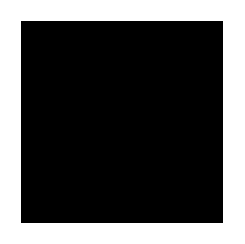}} \\
      $\successmark$
    \end{tabular} &
    \begin{tabular}{c}
      \parbox[t][1.8cm][c]{0.1\textwidth}{\includegraphics[width=0.1\textwidth]{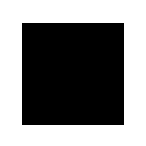}} \\
      $\successmark$
    \end{tabular} &
    \begin{tabular}{c}
      \parbox[t][1.8cm][c]{0.1\textwidth}{\includegraphics[width=0.1\textwidth]{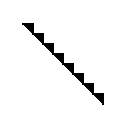}} \\
      $\failmark$
    \end{tabular} &
    \begin{tabular}{c}
      \parbox[t][1.8cm][c]{0.1\textwidth}{\includegraphics[width=0.1\textwidth]{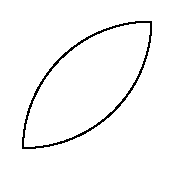}} \\
      $\failmark$
    \end{tabular} \\
    \bottomrule
  \end{tabular}
  }
  \caption{Tasks that are successfully solved by base models (i.e., \gptfouro{}, \pixtral{}-12B, \qwentwovl{}-72B), but not solved by our fine-tuned models (i.e., \pixtral{}-12B-\ft{}, \qwentwovl{}-72B-\ft{}) in the \datasetAll{} dataset. A total of $3$ tasks (0.36\%) match this criteria.
  Each row shows a ground truth image (leftmost) followed by the corresponding images generated by executing the each model's generated Python code. Success ($\successmark$) and failure ($\failmark$) are determined by our evaluation framework using symbolic comparison.
  }
  \label{fig:failure_examples_all_success_except_ft}
\end{table*}

%% file: appendix-figs/fig_failure_cases_only_gpt4o.tex
% !TEX root =  main.tex
%%%%%%%%%%%%%%%%%%%%%%%%%%%%%%%%%%%%%%%%%%%%%%%
%%%%%%%%%%%%%%%%%%%%%%%%%%%%%%%%%%%%%%%%%%%%%%%

% first table
\begin{table*}[t]
  \centering
  \scalebox{0.75}{
  \begin{tabular}{c|ccccc}
    \toprule
    Input Image & \gptfouro & \pixtral{}-12B & \qwentwovl{}-72B & \pixtral{}-12B-\ft{} & \qwentwovl{}-72B-\ft{} \\
    \midrule
    \midrule
    \begin{tabular}{c}
      \parbox[t][1.8cm][c]{0.1\textwidth}{\includegraphics[width=0.1\textwidth]{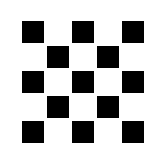}} \\
      $$
    \end{tabular} & 
    \begin{tabular}{c}
      \parbox[t][1.8cm][c]{0.12\textwidth}{\centering \includegraphics[width=0.1\textwidth]{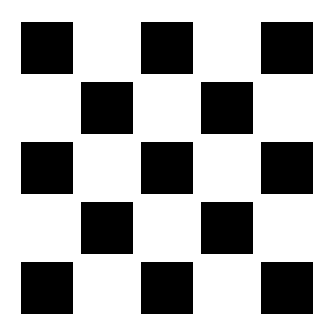}} \\
      $\successmark$
    \end{tabular} &
    \begin{tabular}{c}
      \parbox[t][1.8cm][c]{0.12\textwidth}{\centering \includegraphics[width=0.08\textwidth]{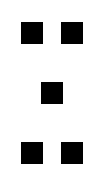}} \\
      $\failmark$
    \end{tabular} &
    \begin{tabular}{c}
      \parbox[t][1.8cm][c]{0.12\textwidth}{\centering \includegraphics[width=0.1\textwidth]{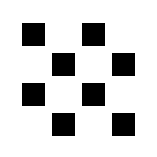}} \\
      $\failmark$
    \end{tabular} &
    \begin{tabular}{c}
      \parbox[t][1.8cm][c]{0.12\textwidth}{\centering \includegraphics[width=0.1\textwidth]{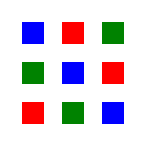}} \\
      $\failmark$
    \end{tabular} &
    \begin{tabular}{c}
      \parbox[t][1.8cm][c]{0.12\textwidth}{\centering \includegraphics[width=0.1\textwidth]{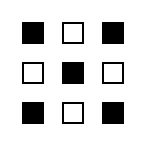}} \\
      $\failmark$
    \end{tabular} \\
    \midrule
    \begin{tabular}{c}
      \parbox[t][1.8cm][c]{0.1\textwidth}{\includegraphics[width=0.1\textwidth]{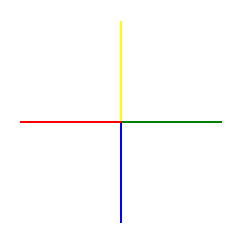}} \\
      $$
    \end{tabular} & 
    \begin{tabular}{c}
      \parbox[t][1.8cm][c]{0.12\textwidth}{\centering \includegraphics[width=0.1\textwidth]{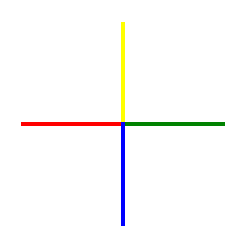}} \\
      $\successmark$
    \end{tabular} &
    \begin{tabular}{c}
      \parbox[t][1.8cm][c]{0.12\textwidth}{\centering \includegraphics[width=0.1\textwidth]{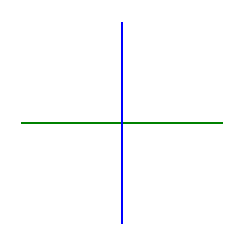}} \\
      $\failmark$
    \end{tabular} &
    \begin{tabular}{c}
      \parbox[t][1.8cm][c]{0.12\textwidth}{\centering \includegraphics[width=0.1\textwidth]{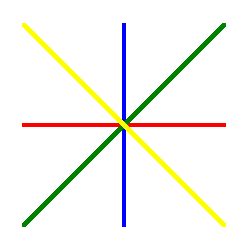}} \\
      $\failmark$
    \end{tabular} &
    \begin{tabular}{c}
      \parbox[t][1.8cm][c]{0.12\textwidth}{\centering \includegraphics[width=0.1\textwidth]{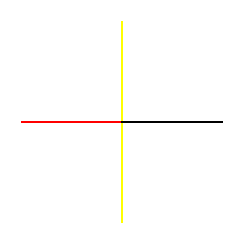}} \\
      $\failmark$
    \end{tabular} &
    \begin{tabular}{c}
      \parbox[t][1.8cm][c]{0.12\textwidth}{\centering \includegraphics[width=0.1\textwidth]{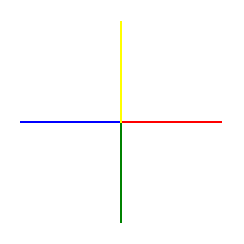}} \\
      $\failmark$
    \end{tabular} \\
    \midrule
    \begin{tabular}{c}
      \parbox[t][1.8cm][c]{0.1\textwidth}{\includegraphics[width=0.1\textwidth]{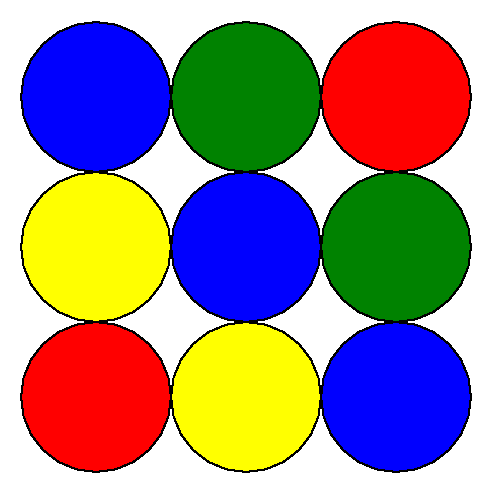}} \\
      $$
    \end{tabular} & 
    \begin{tabular}{c}
      \parbox[t][1.8cm][c]{0.12\textwidth}{\centering \includegraphics[width=0.1\textwidth]{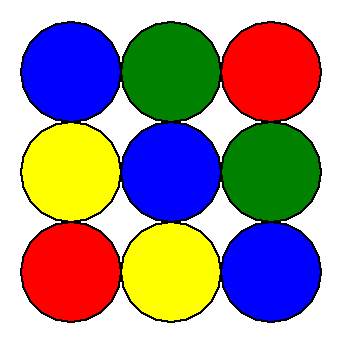}} \\
      $\successmark$
    \end{tabular} &
    \begin{tabular}{c}
      \parbox[t][1.8cm][c]{0.12\textwidth}{\centering \includegraphics[width=0.1\textwidth]{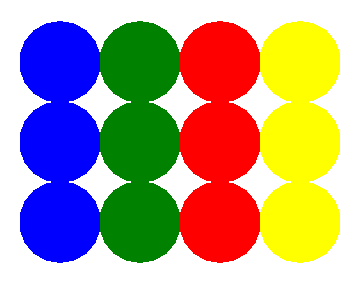}} \\
      $\failmark$
    \end{tabular} &
    \begin{tabular}{c}
      \parbox[t][1.8cm][c]{0.12\textwidth}{\centering \includegraphics[width=0.1\textwidth]{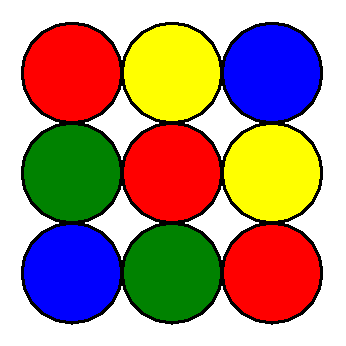}} \\
      $\failmark$
    \end{tabular} &
    \begin{tabular}{c}
      \parbox[t][1.8cm][c]{0.12\textwidth}{\centering \includegraphics[width=0.1\textwidth]{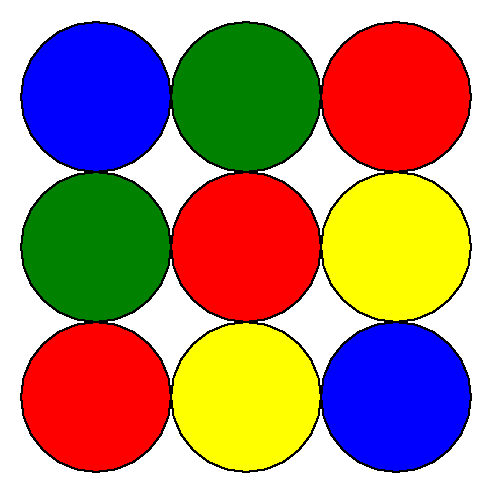}} \\
      $\failmark$
    \end{tabular} &
    \begin{tabular}{c}
      \parbox[t][1.8cm][c]{0.12\textwidth}{\centering \includegraphics[width=0.1\textwidth]{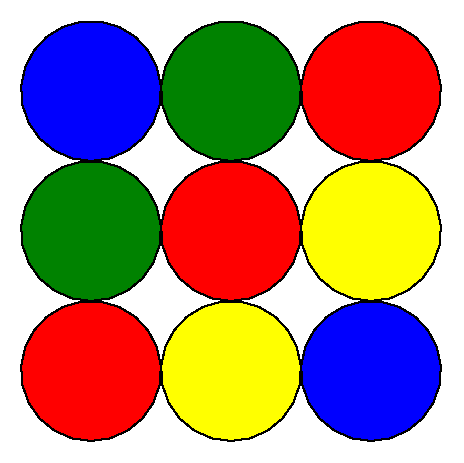}} \\
      $\failmark$
    \end{tabular} \\
    \midrule
    \begin{tabular}{c}
      \parbox[t][1.8cm][c]{0.1\textwidth}{\includegraphics[width=0.1\textwidth]{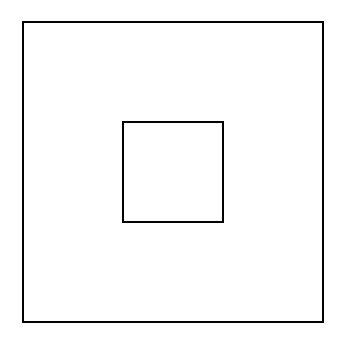}} \\
      $$
    \end{tabular} & 
    \begin{tabular}{c}
      \parbox[t][1.8cm][c]{0.12\textwidth}{\centering \includegraphics[width=0.1\textwidth]{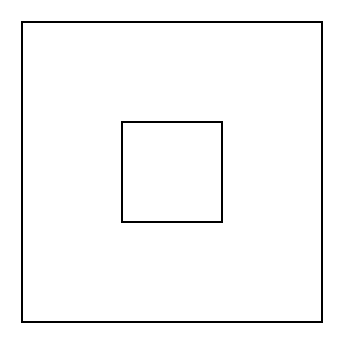}} \\
      $\successmark$
    \end{tabular} &
    \begin{tabular}{c}
      \parbox[t][1.8cm][c]{0.12\textwidth}{\centering \includegraphics[width=0.08\textwidth]{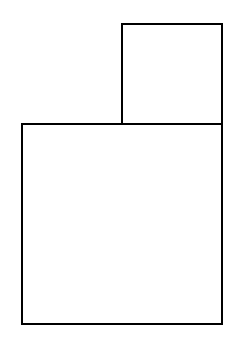}} \\
      $\failmark$
    \end{tabular} &
    \begin{tabular}{c}
      \parbox[t][1.8cm][c]{0.12\textwidth}{\centering \includegraphics[width=0.1\textwidth]{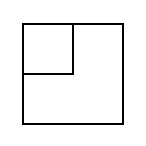}} \\
      $\failmark$
    \end{tabular} &
    \begin{tabular}{c}
      \parbox[t][1.8cm][c]{0.12\textwidth}{\centering \includegraphics[width=0.1\textwidth]{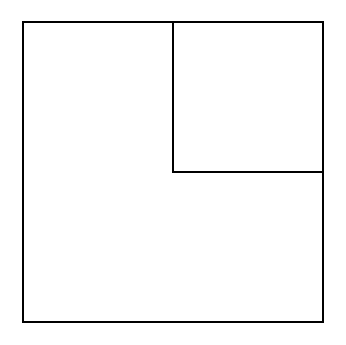}} \\
      $\failmark$
    \end{tabular} &
    \begin{tabular}{c}
      \parbox[t][1.8cm][c]{0.12\textwidth}{\centering \includegraphics[width=0.1\textwidth]{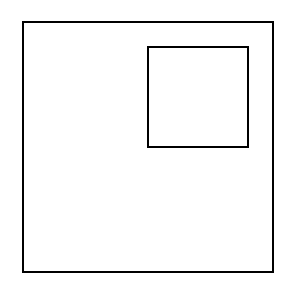}} \\
      $\failmark$
    \end{tabular} \\
    \midrule
    \begin{tabular}{c}
      \parbox[t][1.8cm][c]{0.1\textwidth}{\includegraphics[width=0.1\textwidth]{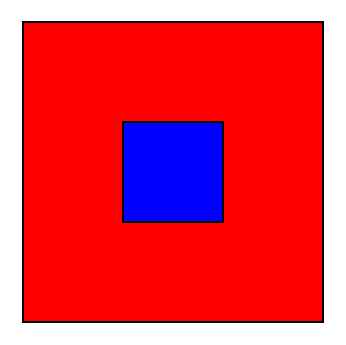}} \\
      $$
    \end{tabular} & 
    \begin{tabular}{c}
      \parbox[t][1.8cm][c]{0.12\textwidth}{\centering \includegraphics[width=0.1\textwidth]{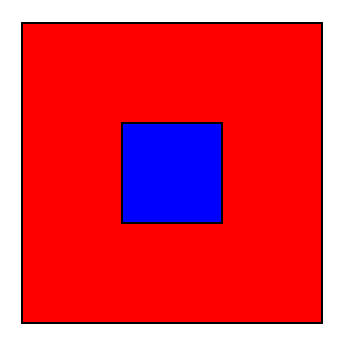}} \\
      $\successmark$
    \end{tabular} &
    \begin{tabular}{c}
      \parbox[t][1.8cm][c]{0.12\textwidth}{\centering \includegraphics[width=0.1\textwidth]{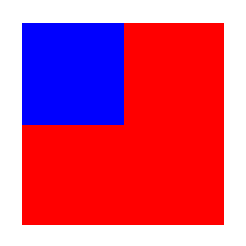}} \\
      $\failmark$
    \end{tabular} &
    \begin{tabular}{c}
      \parbox[t][1.8cm][c]{0.12\textwidth}{\centering \includegraphics[width=0.1\textwidth]{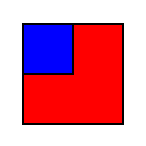}} \\
      $\failmark$
    \end{tabular} &
    \begin{tabular}{c}
      \parbox[t][1.8cm][c]{0.12\textwidth}{\centering \includegraphics[width=0.1\textwidth]{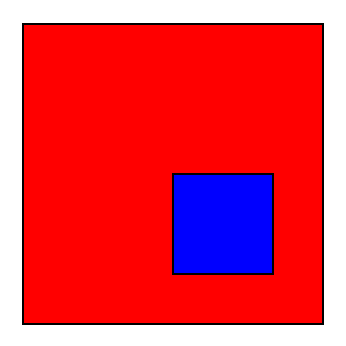}} \\
      $\failmark$
    \end{tabular} &
    \begin{tabular}{c}
      \parbox[t][1.8cm][c]{0.12\textwidth}{\centering \includegraphics[width=0.1\textwidth]{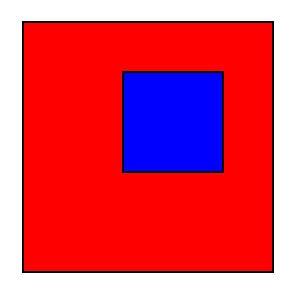}} \\
      $\failmark$
    \end{tabular} \\
    \midrule
    \begin{tabular}{c}
      \parbox[t][1.8cm][c]{0.1\textwidth}{\includegraphics[width=0.1\textwidth]{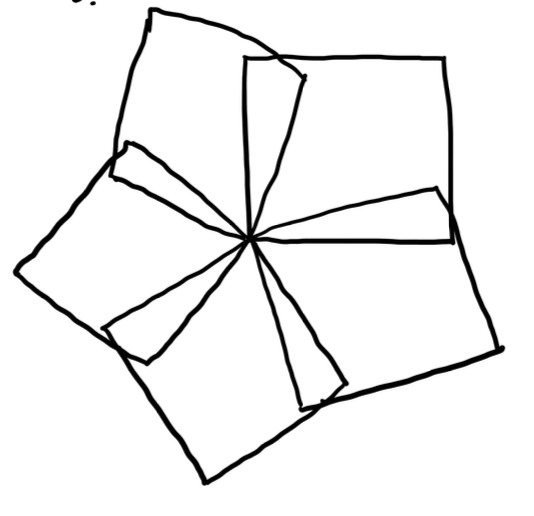}} \\
      $$
    \end{tabular} & 
    \begin{tabular}{c}
      \parbox[t][1.8cm][c]{0.12\textwidth}{\centering \includegraphics[width=0.1\textwidth]{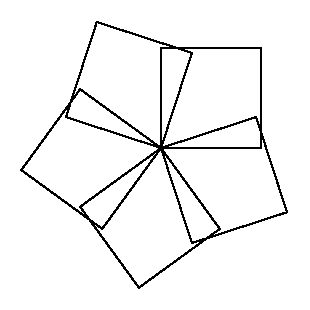}} \\
      $\successmark$
    \end{tabular} &
    \begin{tabular}{c}
      \parbox[t][1.8cm][c]{0.12\textwidth}{\centering \includegraphics[width=0.1\textwidth]{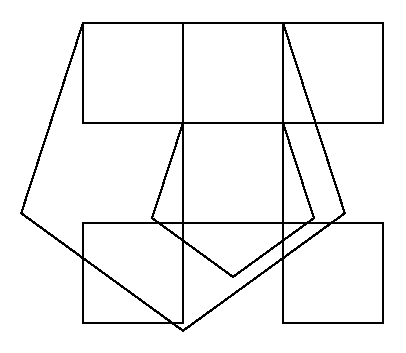}} \\
      $\failmark$
    \end{tabular} &
    \begin{tabular}{c}
      \parbox[t][1.8cm][c]{0.12\textwidth}{\centering \includegraphics[width=0.1\textwidth]{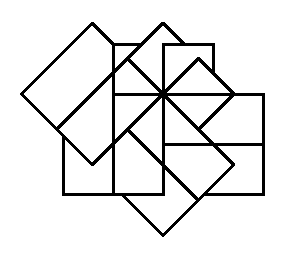}} \\
      $\failmark$
    \end{tabular} &
    \begin{tabular}{c}
      \parbox[t][1.8cm][c]{0.12\textwidth}{\centering \includegraphics[width=0.1\textwidth]{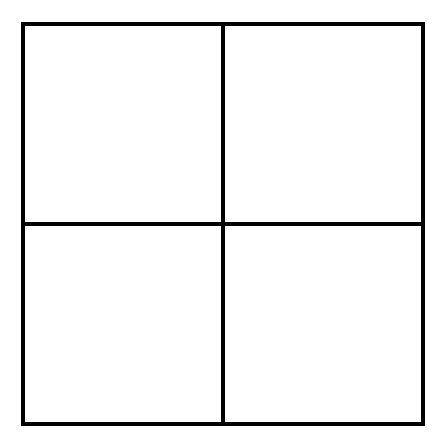}} \\
      $\failmark$
    \end{tabular} &
    \begin{tabular}{c}
      \parbox[t][1.8cm][c]{0.12\textwidth}{\centering \includegraphics[width=0.1\textwidth]{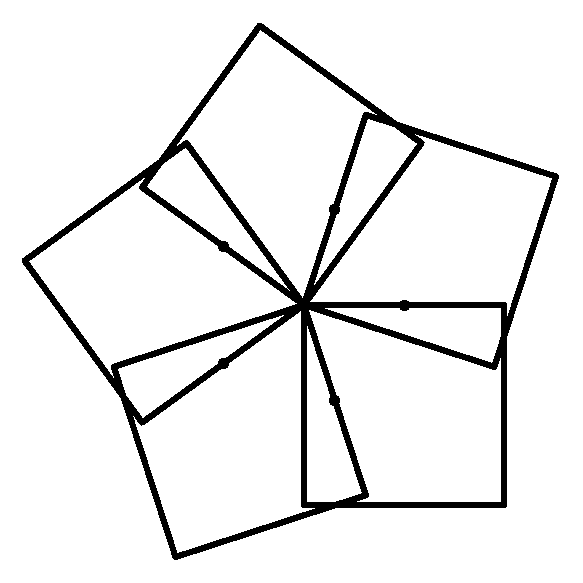}} \\
      $\failmark$
    \end{tabular} \\
    \midrule
    \begin{tabular}{c}
      \parbox[t][1.8cm][c]{0.1\textwidth}{\includegraphics[width=0.1\textwidth]{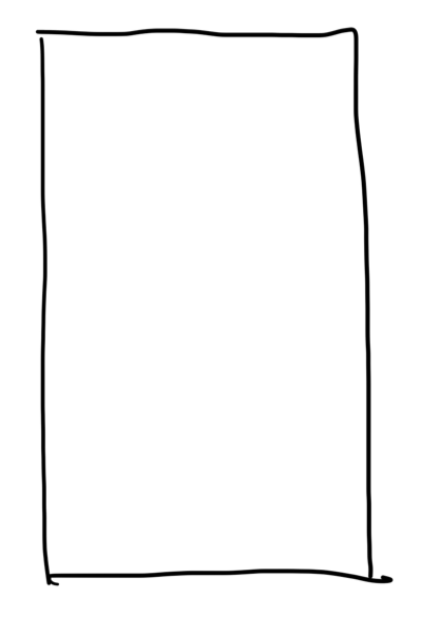}} \\
      $$
    \end{tabular} & 
    \begin{tabular}{c}
      \parbox[t][1.8cm][c]{0.12\textwidth}{\centering \includegraphics[width=0.06\textwidth]{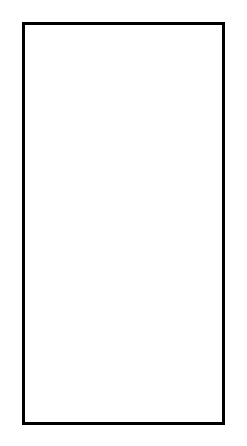}} \\
      $\successmark$
    \end{tabular} &
    \begin{tabular}{c}
      \parbox[t][1.8cm][c]{0.12\textwidth}{\centering \includegraphics[width=0.08\textwidth]{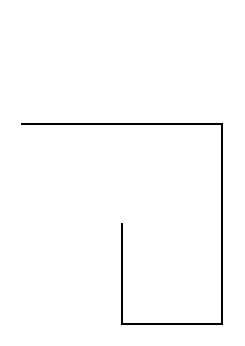}} \\
      $\failmark$
    \end{tabular} &
    \begin{tabular}{c}
      \parbox[t][1.8cm][c]{0.12\textwidth}{\centering \includegraphics[width=0.1\textwidth]{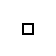}} \\
      $\failmark$
    \end{tabular} &
    \begin{tabular}{c}
      \parbox[t][1.8cm][c]{0.12\textwidth}{\centering \includegraphics[width=0.1\textwidth]{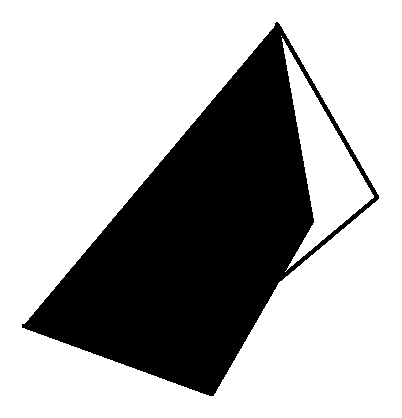}} \\
      $\failmark$
    \end{tabular} &
    \begin{tabular}{c}
      \parbox[t][1.8cm][c]{0.12\textwidth}{\centering \includegraphics[width=0.1\textwidth]{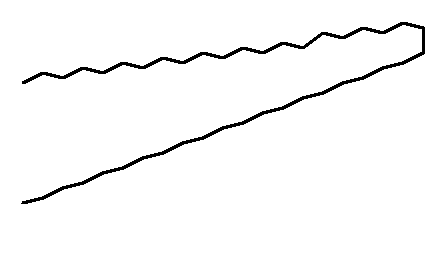}} \\
      $\failmark$
    \end{tabular} \\
    \midrule
    \begin{tabular}{c}
      \parbox[t][1.8cm][c]{0.1\textwidth}{\includegraphics[width=0.1\textwidth]{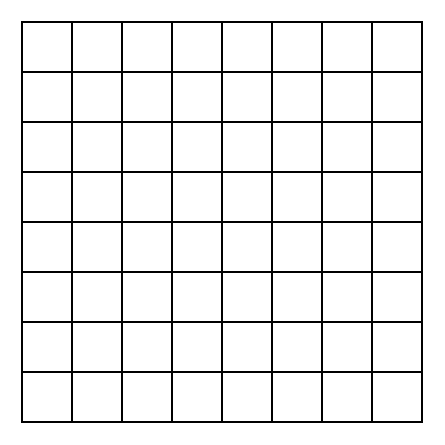}} \\
      $$
    \end{tabular} & 
    \begin{tabular}{c}
      \parbox[t][1.8cm][c]{0.12\textwidth}{\centering \includegraphics[width=0.1\textwidth]{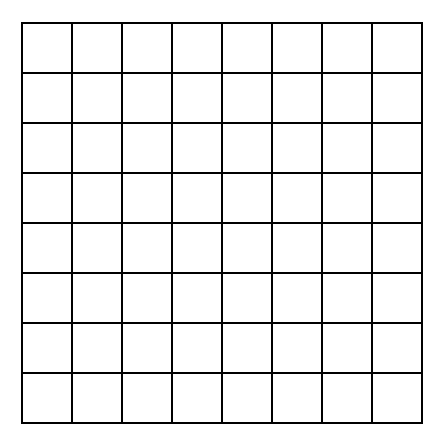}} \\
      $\successmark$
    \end{tabular} &
    \begin{tabular}{c}
      \parbox[t][1.8cm][c]{0.12\textwidth}{\centering \includegraphics[width=0.1\textwidth]{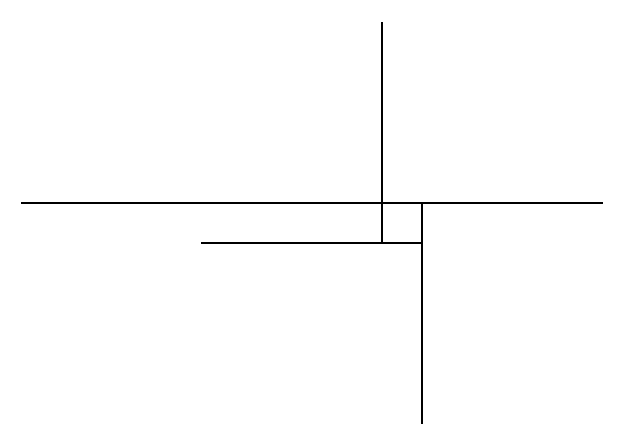}} \\
      $\failmark$
    \end{tabular} &
    \begin{tabular}{c}
      \parbox[t][1.8cm][c]{0.12\textwidth}{\centering \includegraphics[width=0.1\textwidth]{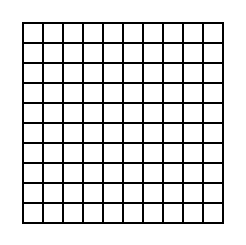}} \\
      $\failmark$
    \end{tabular} &
    \begin{tabular}{c}
      \parbox[t][1.8cm][c]{0.12\textwidth}{\centering \includegraphics[width=0.1\textwidth]{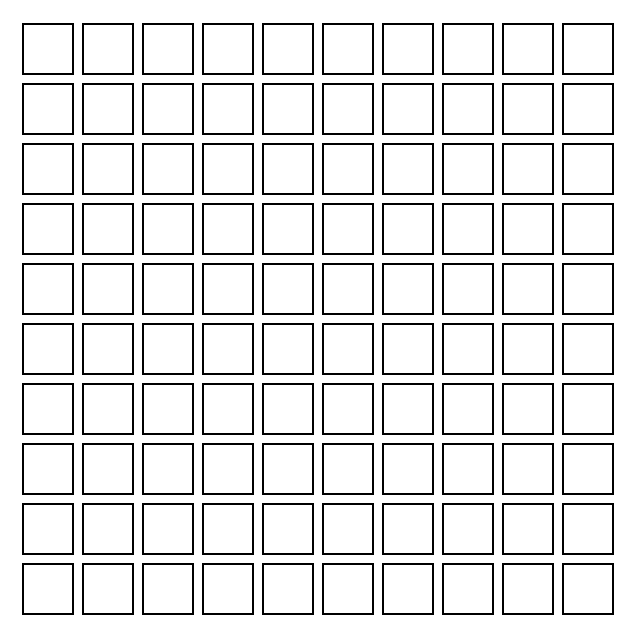}} \\
      $\failmark$
    \end{tabular} &
    \begin{tabular}{c}
      \parbox[t][1.8cm][c]{0.12\textwidth}{\centering \includegraphics[width=0.1\textwidth]{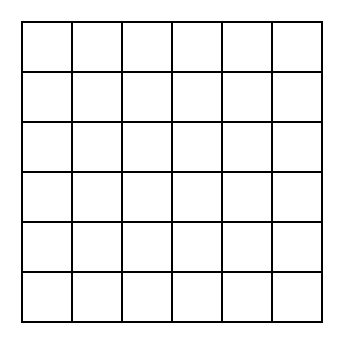}} \\
      $\failmark$
    \end{tabular} \\
    \midrule
    \begin{tabular}{c}
      \parbox[t][1.8cm][c]{0.1\textwidth}{\includegraphics[width=0.1\textwidth]{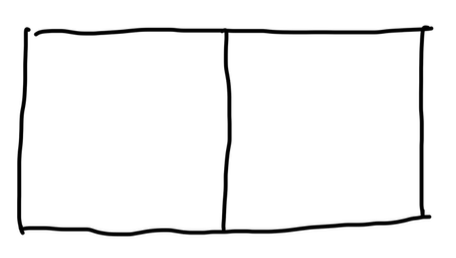}} \\
      $$
    \end{tabular} & 
    \begin{tabular}{c}
      \parbox[t][1.8cm][c]{0.12\textwidth}{\centering \includegraphics[width=0.1\textwidth]{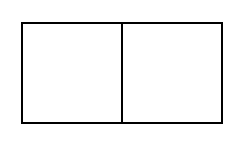}} \\
      $\successmark$
    \end{tabular} &
    \begin{tabular}{c}
      \parbox[t][1.8cm][c]{0.12\textwidth}{\centering \includegraphics[width=0.1\textwidth]{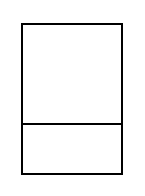}} \\
      $\failmark$
    \end{tabular} &
    \begin{tabular}{c}
      \parbox[t][1.8cm][c]{0.12\textwidth}{\centering \includegraphics[width=0.1\textwidth]{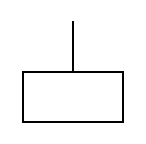}} \\
      $\failmark$
    \end{tabular} &
    \begin{tabular}{c}
      \parbox[t][1.8cm][c]{0.12\textwidth}{\centering \includegraphics[width=0.1\textwidth]{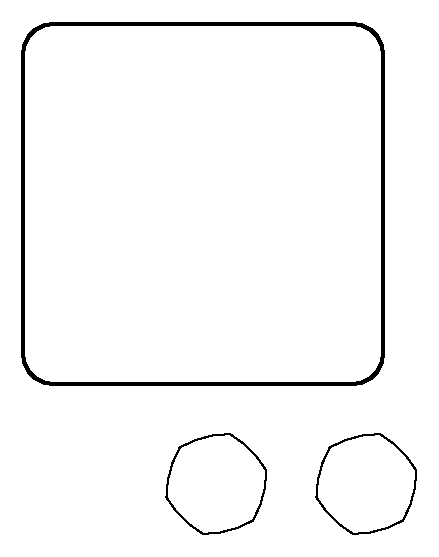}} \\
      $\failmark$
    \end{tabular} &
    \begin{tabular}{c}
      \parbox[t][1.8cm][c]{0.12\textwidth}{\centering \includegraphics[width=0.1\textwidth]{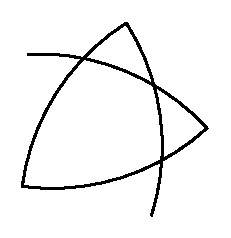}} \\
      $\failmark$
    \end{tabular} \\
    \midrule
    \begin{tabular}{c}
      \parbox[t][1.8cm][c]{0.1\textwidth}{\includegraphics[width=0.1\textwidth]{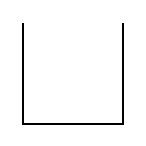}} \\
      $$
    \end{tabular} & 
    \begin{tabular}{c}
      \parbox[t][1.8cm][c]{0.12\textwidth}{\centering \includegraphics[width=0.1\textwidth]{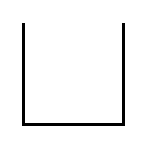}} \\
      $\successmark$
    \end{tabular} &
    \begin{tabular}{c}
      \parbox[t][1.8cm][c]{0.12\textwidth}{\centering \includegraphics[width=0.1\textwidth]{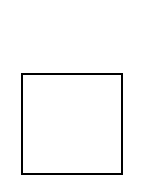}} \\
      $\failmark$
    \end{tabular} &
    \begin{tabular}{c}
      \parbox[t][1.8cm][c]{0.12\textwidth}{\centering \includegraphics[width=0.1\textwidth]{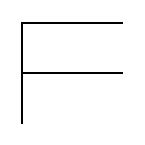}} \\
      $\failmark$
    \end{tabular} &
    \begin{tabular}{c}
      \parbox[t][1.8cm][c]{0.12\textwidth}{\centering \includegraphics[width=0.1\textwidth]{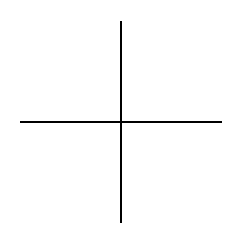}} \\
      $\failmark$
    \end{tabular} &
    \begin{tabular}{c}
      \parbox[t][1.8cm][c]{0.12\textwidth}{\centering \includegraphics[width=0.1\textwidth]{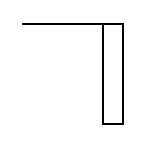}} \\
      $\failmark$
    \end{tabular} \\
    \bottomrule
  \end{tabular}
  }
  \caption{
  Example tasks that are only successfully solved by \gptfouro{} and not by \pixtral{}-12B, \qwentwovl{}-72B, \pixtral{}-12B-\ft{}, or \qwentwovl{}-72B-\ft{} models in the \datasetAll{} dataset. A total of $23$ tasks ($2.79\%$) match this criterion.
  Each row shows a ground truth image (leftmost) followed by the corresponding images generated by executing the each model's generated Python code. Success ($\successmark$) and failure ($\failmark$) are determined by our evaluation framework using symbolic comparison.
  }
  \label{fig:failure_examples_only_gpt4o}
\end{table*}

%% file: appendix-figs/fig_failure_cases_only_ft.tex
% !TEX root =  main.tex
%%%%%%%%%%%%%%%%%%%%%%%%%%%%%%%%%%%%%%%%%%%%%%%
%%%%%%%%%%%%%%%%%%%%%%%%%%%%%%%%%%%%%%%%%%%%%%%

% first table
\begin{table*}[t]
  \centering
  \scalebox{0.75}{
  \begin{tabular}{c|ccccc}
    \toprule
    Input Image & \gptfouro & \pixtral{}-12B & \qwentwovl{}-72B & \pixtral{}-12B-\ft{} & \qwentwovl{}-72B-\ft{} \\
    \midrule
    \midrule
    \begin{tabular}{c}
      \parbox[t][1.8cm][c]{0.1\textwidth}{\centering \includegraphics[width=0.1\textwidth]{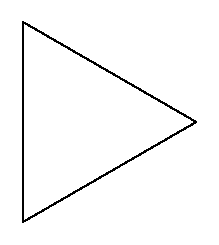}} \\
      $$
    \end{tabular} & 
    \begin{tabular}{c}
      \parbox[t][1.8cm][c]{0.1\textwidth}{\centering \includegraphics[width=0.1\textwidth]{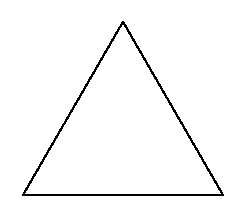}} \\
      $\failmark$
    \end{tabular} &
    \begin{tabular}{c}
      \parbox[t][1.8cm][c]{0.1\textwidth}{\centering \includegraphics[width=0.1\textwidth]{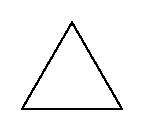}} \\
      $\failmark$
    \end{tabular} &
    \begin{tabular}{c}
      \parbox[t][1.8cm][c]{0.1\textwidth}{\centering \includegraphics[width=0.1\textwidth]{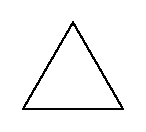}} \\
      $\failmark$
    \end{tabular} &
    \begin{tabular}{c}
      \parbox[t][1.8cm][c]{0.1\textwidth}{\centering \includegraphics[width=0.1\textwidth]{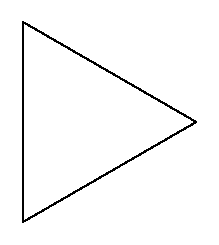}} \\
      $\successmark$
    \end{tabular} &
    \begin{tabular}{c}
      \parbox[t][1.8cm][c]{0.1\textwidth}{\centering \includegraphics[width=0.1\textwidth]{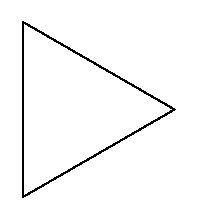}} \\
      $\successmark$
    \end{tabular} \\
    \midrule
    \begin{tabular}{c}
      \parbox[t][1.8cm][c]{0.1\textwidth}{\centering \includegraphics[width=0.1\textwidth]{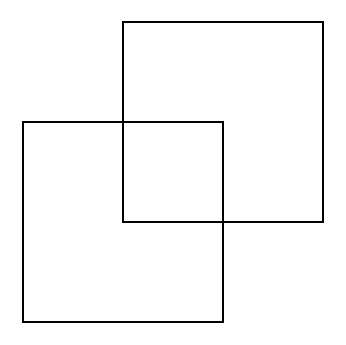}} \\
      $$
    \end{tabular} & 
    \begin{tabular}{c}
      \parbox[t][1.8cm][c]{0.1\textwidth}{\centering \includegraphics[width=0.1\textwidth]{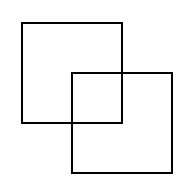}} \\
      $\failmark$
    \end{tabular} &
    \begin{tabular}{c}
      \parbox[t][1.8cm][c]{0.1\textwidth}{\centering \includegraphics[width=0.1\textwidth]{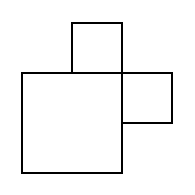}} \\
      $\failmark$
    \end{tabular} &
    \begin{tabular}{c}
      \parbox[t][1.8cm][c]{0.1\textwidth}{\centering \includegraphics[width=0.1\textwidth]{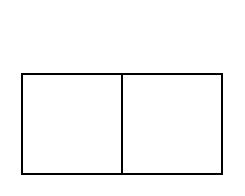}} \\
      $\failmark$
    \end{tabular} &
    \begin{tabular}{c}
      \parbox[t][1.8cm][c]{0.1\textwidth}{\centering \includegraphics[width=0.1\textwidth]{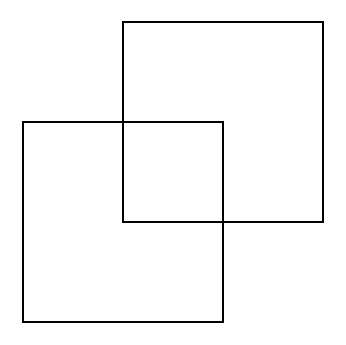}} \\
      $\successmark$
    \end{tabular} &
    \begin{tabular}{c}
      \parbox[t][1.8cm][c]{0.1\textwidth}{\centering \includegraphics[width=0.1\textwidth]{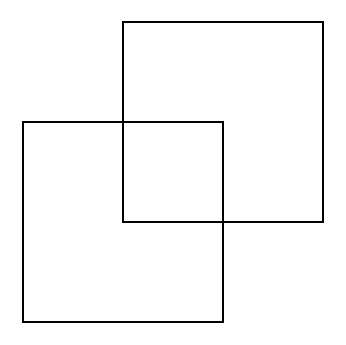}} \\
      $\successmark$
    \end{tabular} \\
    \midrule
    \begin{tabular}{c}
      \parbox[t][1.8cm][c]{0.1\textwidth}{\centering \includegraphics[width=0.1\textwidth]{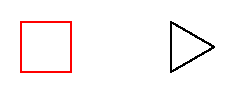}} \\
      $$
    \end{tabular} & 
    \begin{tabular}{c}
      \parbox[t][1.8cm][c]{0.12\textwidth}{\centering \includegraphics[width=0.1\textwidth]{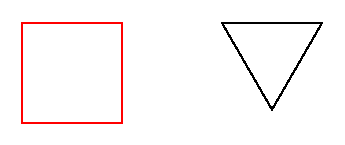}} \\
      $\failmark$
    \end{tabular} &
    \begin{tabular}{c}
      \parbox[t][1.8cm][c]{0.12\textwidth}{\centering \includegraphics[width=0.1\textwidth]{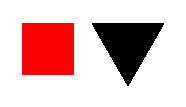}} \\
      $\failmark$
    \end{tabular} &
    \begin{tabular}{c}
      \parbox[t][1.8cm][c]{0.12\textwidth}{\centering \includegraphics[width=0.1\textwidth]{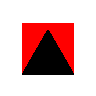}} \\
      $\failmark$
    \end{tabular} &
    \begin{tabular}{c}
      \parbox[t][1.8cm][c]{0.12\textwidth}{\centering \includegraphics[width=0.1\textwidth]{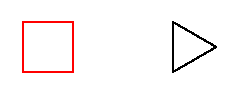}} \\
      $\successmark$
    \end{tabular} &
    \begin{tabular}{c}
      \parbox[t][1.8cm][c]{0.12\textwidth}{\centering \includegraphics[width=0.1\textwidth]{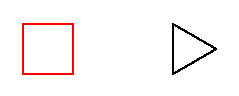}} \\
      $\successmark$
    \end{tabular} \\
    \midrule
    \begin{tabular}{c}
      \parbox[t][1.8cm][c]{0.1\textwidth}{\centering \includegraphics[width=0.1\textwidth]{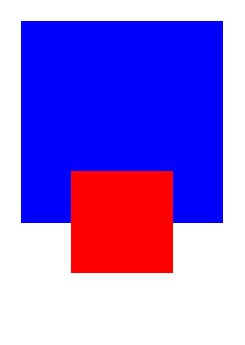}} \\
      $$
    \end{tabular} & 
    \begin{tabular}{c}
      \parbox[t][1.8cm][c]{0.12\textwidth}{\centering \includegraphics[width=0.1\textwidth]{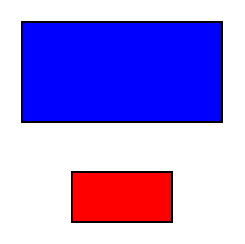}} \\
      $\failmark$
    \end{tabular} &
    \begin{tabular}{c}
      \parbox[t][1.8cm][c]{0.12\textwidth}{\centering \includegraphics[width=0.1\textwidth]{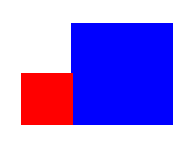}} \\
      $\failmark$
    \end{tabular} &
    \begin{tabular}{c}
      \parbox[t][1.8cm][c]{0.12\textwidth}{\centering \includegraphics[width=0.06\textwidth]{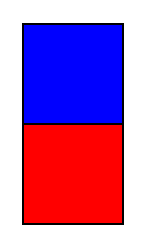}} \\
      $\failmark$
    \end{tabular} &
    \begin{tabular}{c}
      \parbox[t][1.8cm][c]{0.12\textwidth}{\centering \includegraphics[width=0.08\textwidth]{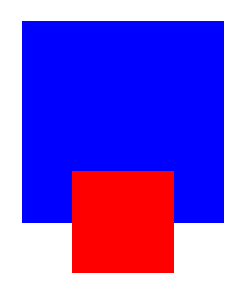}} \\
      $\successmark$
    \end{tabular} &
    \begin{tabular}{c}
      \parbox[t][1.8cm][c]{0.12\textwidth}{\centering \includegraphics[width=0.1\textwidth]{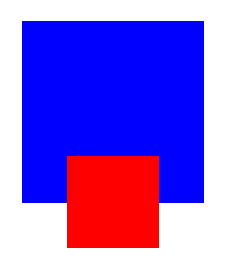}} \\
      $\successmark$
    \end{tabular} \\
    \midrule
    \begin{tabular}{c}
      \parbox[t][1.8cm][c]{0.1\textwidth}{\centering \includegraphics[width=0.1\textwidth]{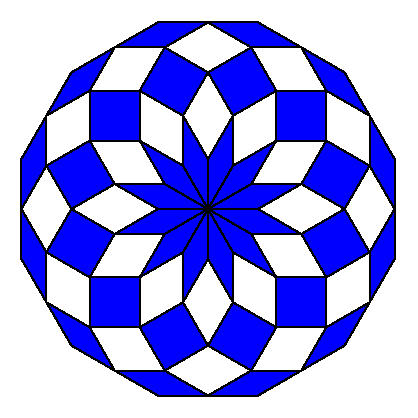}} \\
      $$
    \end{tabular} & 
    \begin{tabular}{c}
      \parbox[t][1.8cm][c]{0.1\textwidth}{\centering \includegraphics[width=0.1\textwidth]{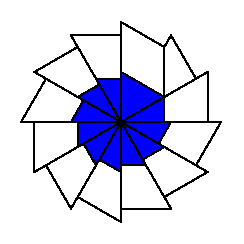}} \\
      $\failmark$
    \end{tabular} &
    \begin{tabular}{c}
      \parbox[t][1.8cm][c]{0.1\textwidth}{\centering \includegraphics[width=0.07\textwidth]{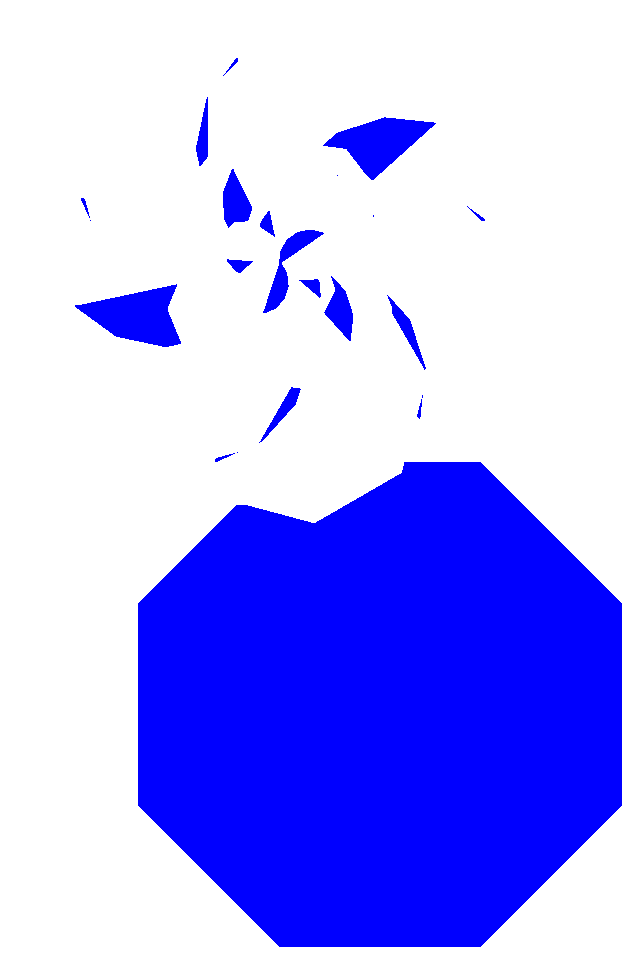}} \\
      $\failmark$
    \end{tabular} &
    \begin{tabular}{c}
      \parbox[t][1.8cm][c]{0.1\textwidth}{\centering N.A.} \\
      $\failmark$
    \end{tabular} &
    \begin{tabular}{c}
      \parbox[t][1.8cm][c]{0.1\textwidth}{\centering \includegraphics[width=0.1\textwidth]{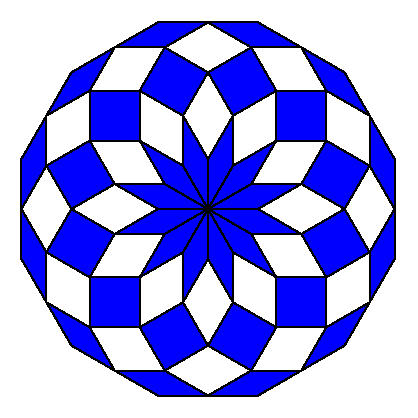}} \\
      $\successmark$
    \end{tabular} &
    \begin{tabular}{c}
      \parbox[t][1.8cm][c]{0.1\textwidth}{\centering \includegraphics[width=0.1\textwidth]{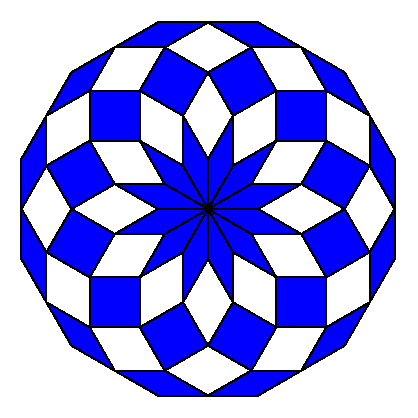}} \\
      $\successmark$
    \end{tabular} \\
    \midrule
    \begin{tabular}{c}
      \parbox[t][1.8cm][c]{0.1\textwidth}{\centering \includegraphics[width=0.1\textwidth]{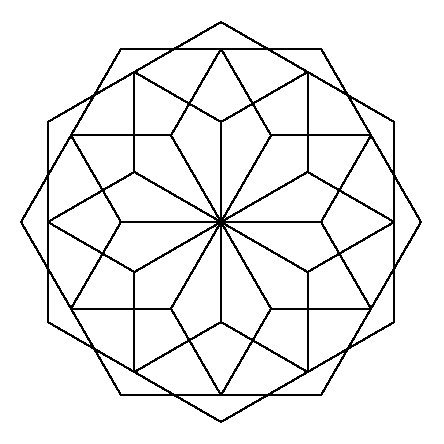}} \\
      $$
    \end{tabular} & 
    \begin{tabular}{c}
      \parbox[t][1.8cm][c]{0.1\textwidth}{\centering \includegraphics[width=0.1\textwidth]{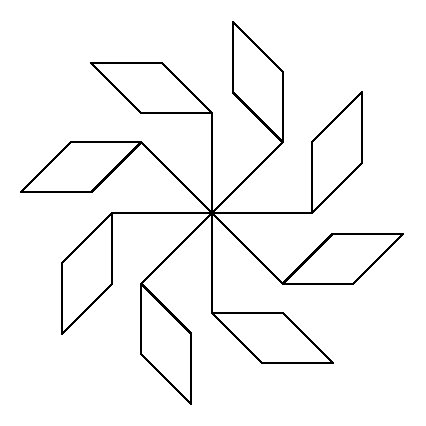}} \\
      $\failmark$
    \end{tabular} &
    \begin{tabular}{c}
      \parbox[t][1.8cm][c]{0.1\textwidth}{\centering N.A.} \\
      $\failmark$
    \end{tabular} &
    \begin{tabular}{c}
      \parbox[t][1.8cm][c]{0.1\textwidth}{\centering N.A.} \\
      $\failmark$
    \end{tabular} &
    \begin{tabular}{c}
      \parbox[t][1.8cm][c]{0.1\textwidth}{\centering \includegraphics[width=0.1\textwidth]{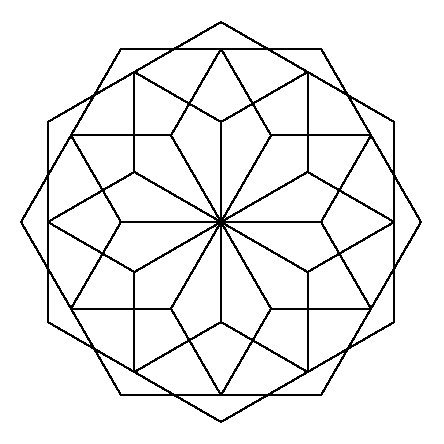}} \\
      $\successmark$
    \end{tabular} &
    \begin{tabular}{c}
      \parbox[t][1.8cm][c]{0.1\textwidth}{\centering \includegraphics[width=0.1\textwidth]{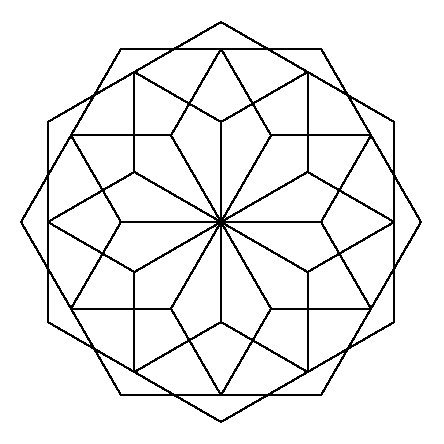}} \\
      $\successmark$
    \end{tabular} \\
    \midrule
    \begin{tabular}{c}
      \parbox[t][1.8cm][c]{0.1\textwidth}{\centering \includegraphics[width=0.1\textwidth]{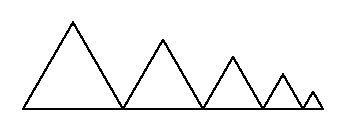}} \\
      $$
    \end{tabular} & 
    \begin{tabular}{c}
      \parbox[t][1.8cm][c]{0.1\textwidth}{\centering \includegraphics[width=0.1\textwidth]{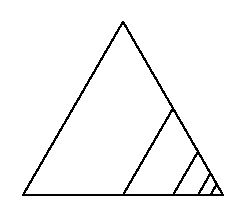}} \\
      $\failmark$
    \end{tabular} &
    \begin{tabular}{c}
      \parbox[t][1.8cm][c]{0.1\textwidth}{\centering \includegraphics[width=0.06\textwidth]{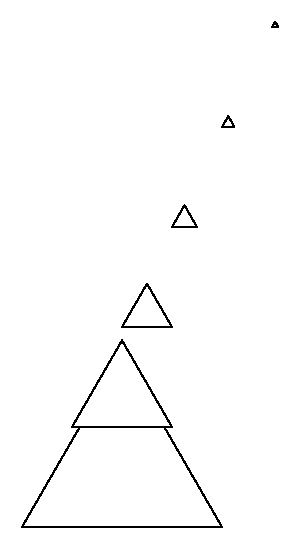}} \\
      $\failmark$
    \end{tabular} &
    \begin{tabular}{c}
      \parbox[t][1.8cm][c]{0.1\textwidth}{\centering \includegraphics[width=0.1\textwidth]{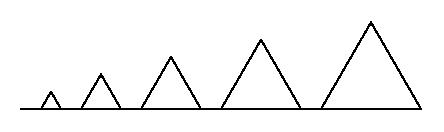}} \\
      $\failmark$
    \end{tabular} &
    \begin{tabular}{c}
      \parbox[t][1.8cm][c]{0.1\textwidth}{\centering \includegraphics[width=0.1\textwidth]{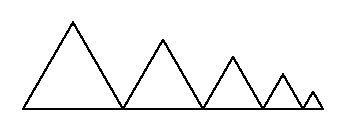}} \\
      $\successmark$
    \end{tabular} &
    \begin{tabular}{c}
      \parbox[t][1.8cm][c]{0.1\textwidth}{\centering \includegraphics[width=0.1\textwidth]{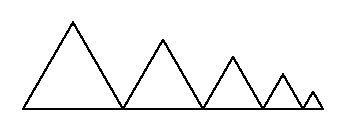}} \\
      $\successmark$
    \end{tabular} \\
    \midrule
    \begin{tabular}{c}
      \parbox[t][1.8cm][c]{0.1\textwidth}{\centering \includegraphics[width=0.1\textwidth]{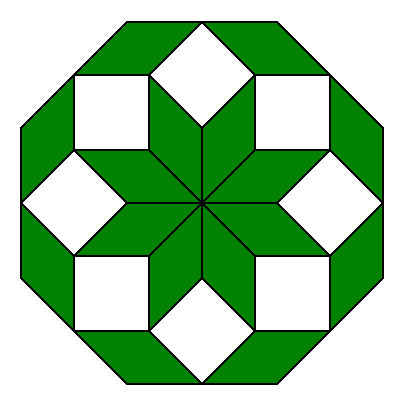}} \\
      $$
    \end{tabular} & 
    \begin{tabular}{c}
      \parbox[t][1.8cm][c]{0.1\textwidth}{\centering \includegraphics[width=0.1\textwidth]{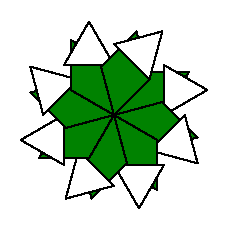}} \\
      $\failmark$
    \end{tabular} &
    \begin{tabular}{c}
      \parbox[t][1.8cm][c]{0.1\textwidth}{\centering N.A.} \\
      $\failmark$
    \end{tabular} &
    \begin{tabular}{c}
      \parbox[t][1.8cm][c]{0.1\textwidth}{\centering N.A.} \\
      $\failmark$
    \end{tabular} &
    \begin{tabular}{c}
      \parbox[t][1.8cm][c]{0.1\textwidth}{\centering \includegraphics[width=0.1\textwidth]{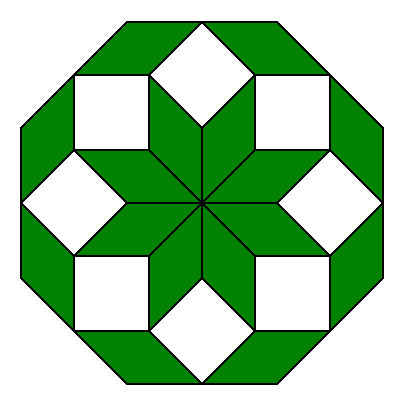}} \\
      $\successmark$
    \end{tabular} &
    \begin{tabular}{c}
      \parbox[t][1.8cm][c]{0.1\textwidth}{\centering \includegraphics[width=0.1\textwidth]{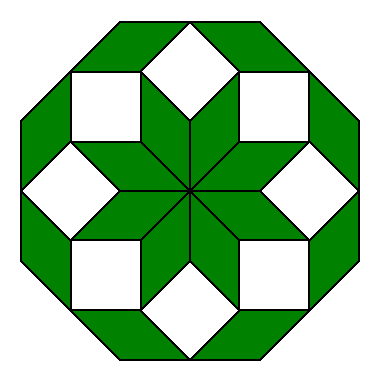}} \\
      $\successmark$
    \end{tabular} \\
    \midrule
    \begin{tabular}{c}
      \parbox[t][1.8cm][c]{0.1\textwidth}{\centering \includegraphics[width=0.1\textwidth]{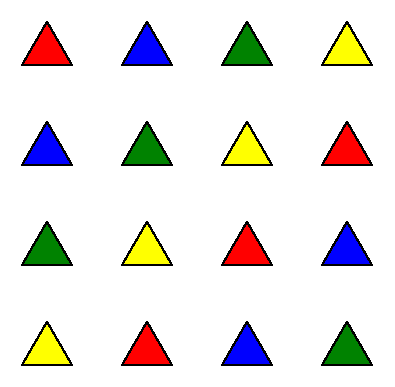}} \\
      $$
    \end{tabular} & 
    \begin{tabular}{c}
      \parbox[t][1.8cm][c]{0.1\textwidth}{\centering \includegraphics[width=0.1\textwidth]{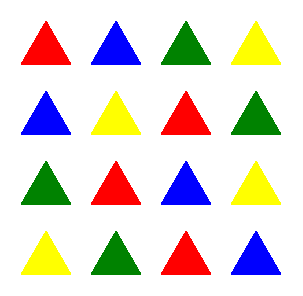}} \\
      $\failmark$
    \end{tabular} &
    \begin{tabular}{c}
      \parbox[t][1.8cm][c]{0.1\textwidth}{\centering \includegraphics[width=0.1\textwidth]{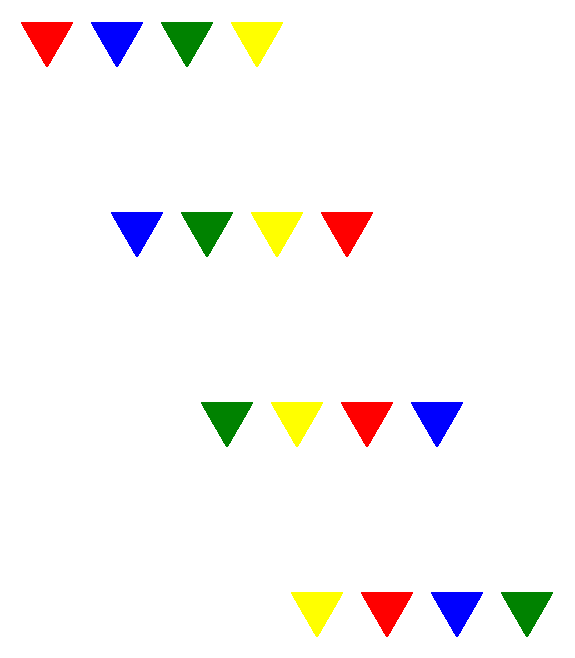}} \\
      $\failmark$
    \end{tabular} &
    \begin{tabular}{c}
      \parbox[t][1.8cm][c]{0.1\textwidth}{\centering \includegraphics[width=0.1\textwidth]{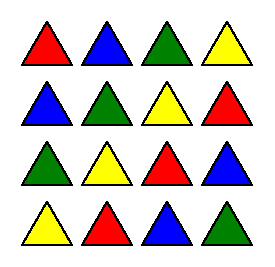}} \\
      $\failmark$
    \end{tabular} &
    \begin{tabular}{c}
      \parbox[t][1.8cm][c]{0.1\textwidth}{\centering \includegraphics[width=0.1\textwidth]{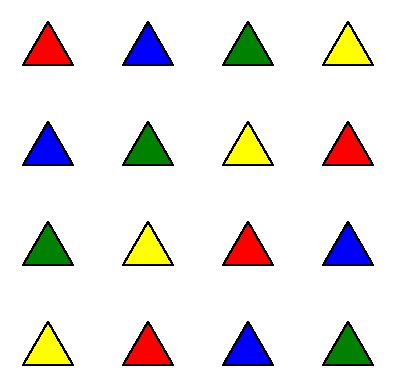}} \\
      $\successmark$
    \end{tabular} &
    \begin{tabular}{c}
      \parbox[t][1.8cm][c]{0.1\textwidth}{\centering \includegraphics[width=0.1\textwidth]{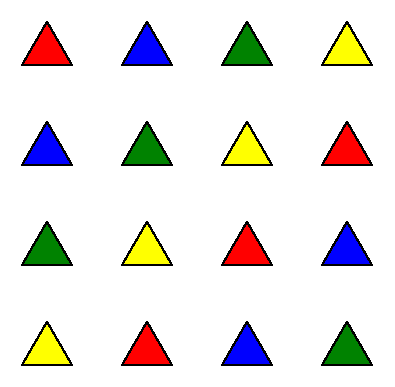}} \\
      $\successmark$
    \end{tabular} \\
    \midrule
    \begin{tabular}{c}
      \parbox[t][1.8cm][c]{0.1\textwidth}{\centering \includegraphics[width=0.1\textwidth]{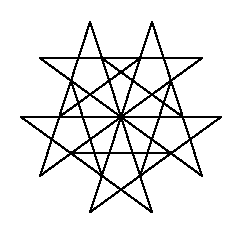}} \\
      $$
    \end{tabular} & 
    \begin{tabular}{c}
      \parbox[t][1.8cm][c]{0.1\textwidth}{\centering \includegraphics[width=0.1\textwidth]{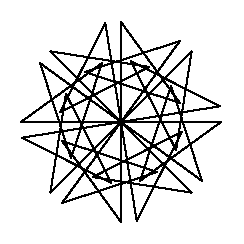}} \\
      $\failmark$
    \end{tabular} &
    \begin{tabular}{c}
      \parbox[t][1.8cm][c]{0.1\textwidth}{\centering \includegraphics[width=0.1\textwidth]{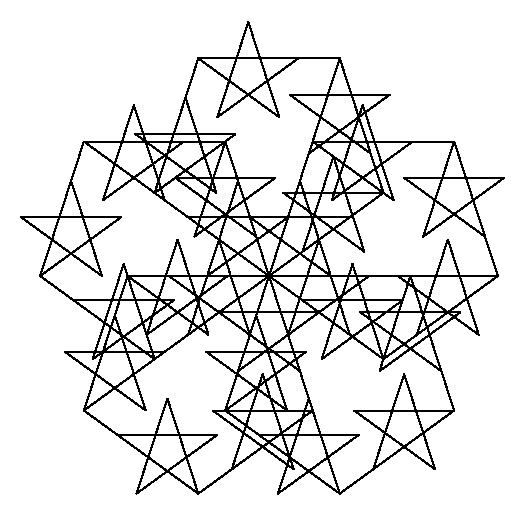}} \\
      $\failmark$
    \end{tabular} &
    \begin{tabular}{c}
      \parbox[t][1.8cm][c]{0.1\textwidth}{\centering \includegraphics[width=0.1\textwidth]{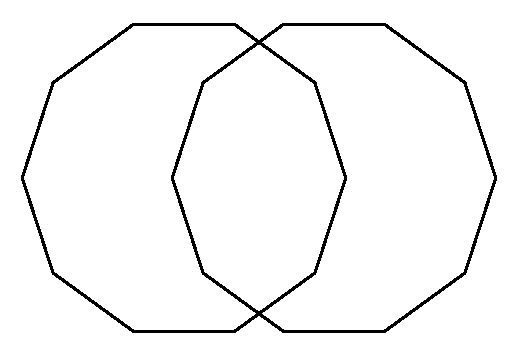}} \\
      $\failmark$
    \end{tabular} &
    \begin{tabular}{c}
      \parbox[t][1.8cm][c]{0.1\textwidth}{\centering \includegraphics[width=0.1\textwidth]{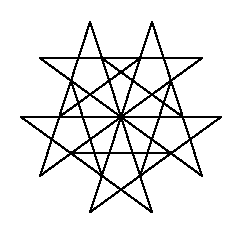}} \\
      $\successmark$
    \end{tabular} &
    \begin{tabular}{c}
      \parbox[t][1.8cm][c]{0.1\textwidth}{\centering \includegraphics[width=0.1\textwidth]{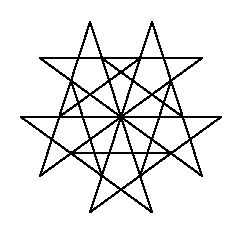}} \\
      $\successmark$
    \end{tabular} \\
    \bottomrule
  \end{tabular}
  }
  \caption{Example tasks that are only successfully solved by our fine-tuned models (i.e., \pixtral{}-12B-\ft{}, \qwentwovl{}-72B-\ft{}) and not by \pixtral{}-12B, \qwentwovl{}-72B, or \gptfouro{} models in the \datasetAll{} dataset. A total of $54$ tasks ($6.56\%$) match this criterion.
  Each row shows a ground truth image (leftmost) followed by the corresponding images generated by executing the each model's generated Python code. Success ($\successmark$) and failure ($\failmark$) are determined by our evaluation framework using symbolic comparison.
  }
  \label{fig:failure_examples_only_ft}
\end{table*}

%% file: appendix-figs/fig_failure_cases_all_fail.tex
% !TEX root =  main.tex
%%%%%%%%%%%%%%%%%%%%%%%%%%%%%%%%%%%%%%%%%%%%%%%
%%%%%%%%%%%%%%%%%%%%%%%%%%%%%%%%%%%%%%%%%%%%%%%

% first table
\begin{table*}[t]
  \centering
  \scalebox{0.75}{
  \begin{tabular}{l|ccccc}
    \toprule
    Input Image & \gptfouro & \pixtral{}-12B & \qwentwovl{}-72B & \pixtral{}-12B-\ft{} & \qwentwovl{}-72B-\ft{} \\
    \midrule
    \midrule
    \begin{tabular}{c}
      \parbox[t][1.8cm][c]{0.1\textwidth}{\centering \includegraphics[width=0.1\textwidth]{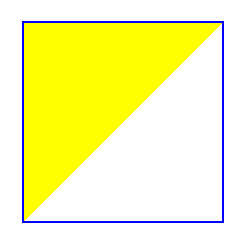}} \\
      $$
    \end{tabular} & 
    \begin{tabular}{c}
      \parbox[t][1.8cm][c]{0.12\textwidth}{\centering \includegraphics[width=0.1\textwidth]{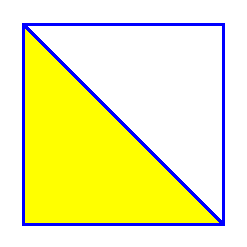}} \\
      $\failmark$
    \end{tabular} &
    \begin{tabular}{c}
      \parbox[t][1.8cm][c]{0.12\textwidth}{\centering \includegraphics[width=0.1\textwidth]{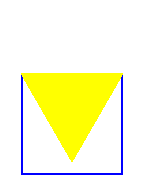}} \\
      $\failmark$
    \end{tabular} &
    \begin{tabular}{c}
      \parbox[t][1.8cm][c]{0.12\textwidth}{\centering \includegraphics[width=0.07\textwidth]{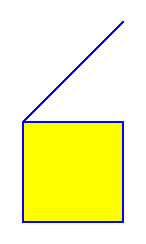}} \\
      $\failmark$
    \end{tabular} &
    \begin{tabular}{c}
      \parbox[t][1.8cm][c]{0.12\textwidth}{\centering \includegraphics[width=0.1\textwidth]{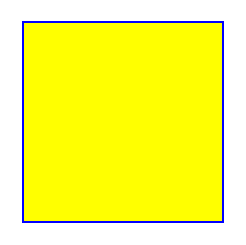}} \\
      $\failmark$
    \end{tabular} &
    \begin{tabular}{c}
      \parbox[t][1.8cm][c]{0.12\textwidth}{\centering \includegraphics[width=0.1\textwidth]{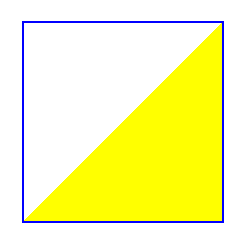}} \\
      $\failmark$
    \end{tabular} \\
    \midrule
    \begin{tabular}{c}
      \parbox[t][1.8cm][c]{0.1\textwidth}{\centering \includegraphics[width=0.1\textwidth]{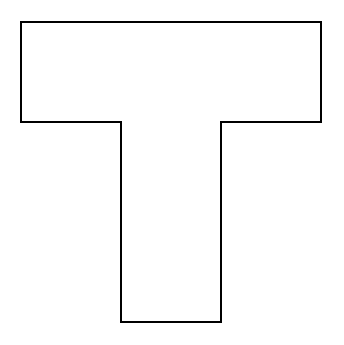}} \\
      $$
    \end{tabular} & 
    \begin{tabular}{c}
      \parbox[t][1.8cm][c]{0.12\textwidth}{\centering \includegraphics[width=0.1\textwidth]{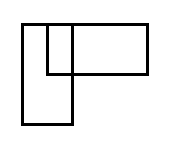}} \\
      $\failmark$
    \end{tabular} &
    \begin{tabular}{c}
      \parbox[t][1.8cm][c]{0.12\textwidth}{\centering \includegraphics[width=0.1\textwidth]{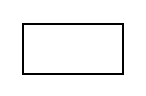}} \\
      $\failmark$
    \end{tabular} &
    \begin{tabular}{c}
      \parbox[t][1.8cm][c]{0.12\textwidth}{\centering N.A.} \\
      $\failmark$
    \end{tabular} &
    \begin{tabular}{c}
      \parbox[t][1.8cm][c]{0.12\textwidth}{\centering \includegraphics[width=0.1\textwidth]{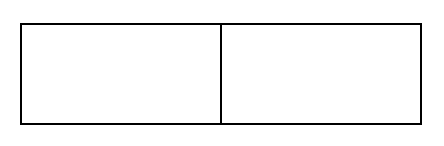}} \\
      $\failmark$
    \end{tabular} &
    \begin{tabular}{c}
      \parbox[t][1.8cm][c]{0.12\textwidth}{\centering \includegraphics[width=0.1\textwidth]{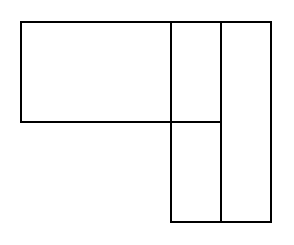}} \\
      $\failmark$
    \end{tabular} \\
    \midrule
    \begin{tabular}{c}
      \parbox[t][1.8cm][c]{0.1\textwidth}{\centering \includegraphics[width=0.1\textwidth]{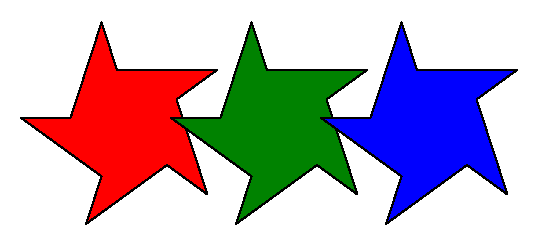}} \\
      $$
    \end{tabular} & 
    \begin{tabular}{c}
      \parbox[t][1.8cm][c]{0.12\textwidth}{\centering \includegraphics[width=0.1\textwidth]{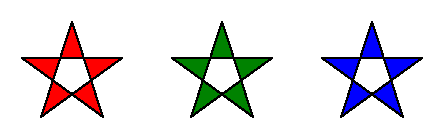}} \\
      $\failmark$
    \end{tabular} &
    \begin{tabular}{c}
      \parbox[t][1.8cm][c]{0.12\textwidth}{\centering \includegraphics[width=0.1\textwidth]{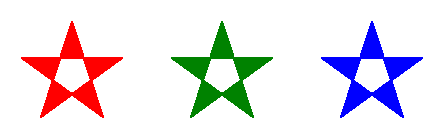}} \\
      $\failmark$
    \end{tabular} &
    \begin{tabular}{c}
      \parbox[t][1.8cm][c]{0.12\textwidth}{\centering \includegraphics[width=0.1\textwidth]{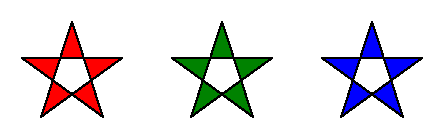}} \\
      $\failmark$
    \end{tabular} &
    \begin{tabular}{c}
      \parbox[t][1.8cm][c]{0.12\textwidth}{\centering \includegraphics[width=0.1\textwidth]{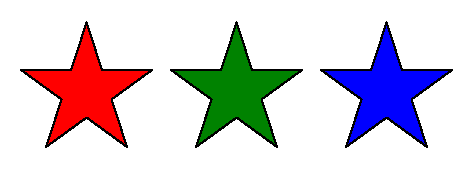}} \\
      $\failmark$
    \end{tabular} &
    \begin{tabular}{c}
      \parbox[t][1.8cm][c]{0.12\textwidth}{\centering \includegraphics[width=0.1\textwidth]{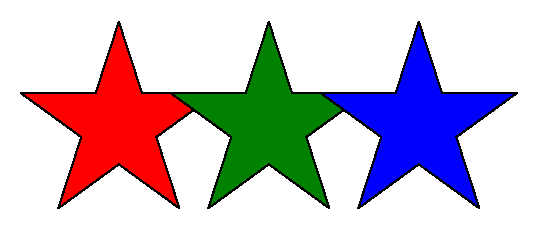}} \\
      $\failmark$
    \end{tabular} \\
    \midrule
    \begin{tabular}{c}
      \parbox[t][1.8cm][c]{0.1\textwidth}{\centering \includegraphics[width=0.1\textwidth]{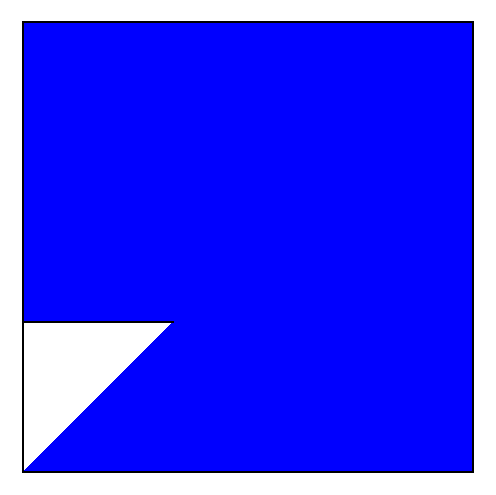}} \\
      $$
    \end{tabular} & 
    \begin{tabular}{c}
      \parbox[t][1.8cm][c]{0.12\textwidth}{\centering \includegraphics[width=0.1\textwidth]{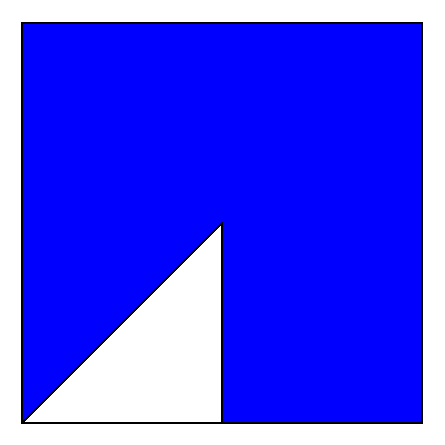}} \\
      $\failmark$
    \end{tabular} &
    \begin{tabular}{c}
      \parbox[t][1.8cm][c]{0.12\textwidth}{\centering \includegraphics[width=0.1\textwidth]{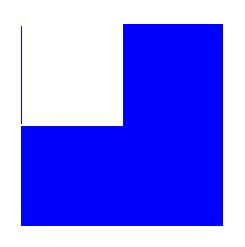}} \\
      $\failmark$
    \end{tabular} &
    \begin{tabular}{c}
      \parbox[t][1.8cm][c]{0.12\textwidth}{\centering \includegraphics[width=0.1\textwidth]{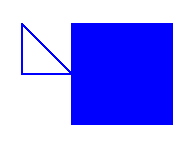}} \\
      $\failmark$
    \end{tabular} &
    \begin{tabular}{c}
      \parbox[t][1.8cm][c]{0.12\textwidth}{\centering \includegraphics[width=0.1\textwidth]{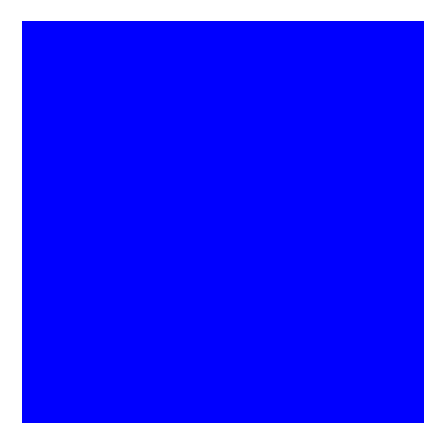}} \\
      $\failmark$
    \end{tabular} &
    \begin{tabular}{c}
      \parbox[t][1.8cm][c]{0.12\textwidth}{\centering \includegraphics[width=0.1\textwidth]{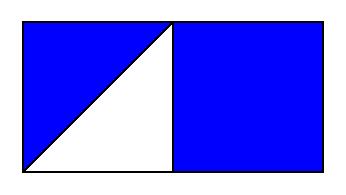}} \\
      $\failmark$
    \end{tabular} \\
    \midrule
    \begin{tabular}{c}
      \parbox[t][1.8cm][c]{0.1\textwidth}{\centering \includegraphics[width=0.02\textwidth]{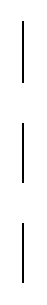}} \\
      $$
    \end{tabular} & 
    \begin{tabular}{c}
      \parbox[t][1.8cm][c]{0.12\textwidth}{\centering \includegraphics[width=0.02\textwidth]{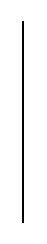}} \\
      $\failmark$
    \end{tabular} &
    \begin{tabular}{c}
      \parbox[t][1.8cm][c]{0.12\textwidth}{\centering \includegraphics[width=0.1\textwidth]{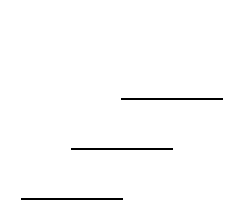}} \\
      $\failmark$
    \end{tabular} &
    \begin{tabular}{c}
      \parbox[t][1.8cm][c]{0.12\textwidth}{\centering \includegraphics[width=0.1\textwidth]{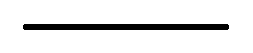}} \\
      $\failmark$
    \end{tabular} &
    \begin{tabular}{c}
      \parbox[t][1.8cm][c]{0.12\textwidth}{\centering \includegraphics[width=0.1\textwidth]{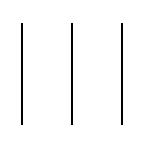}} \\
      $\failmark$
    \end{tabular} &
    \begin{tabular}{c}
      \parbox[t][1.8cm][c]{0.12\textwidth}{\centering \includegraphics[width=0.1\textwidth]{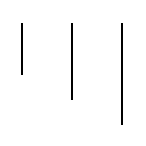}} \\
      $\failmark$
    \end{tabular} \\
    \midrule
    \begin{tabular}{c}
      \parbox[t][1.8cm][c]{0.1\textwidth}{\centering \includegraphics[width=0.04\textwidth]{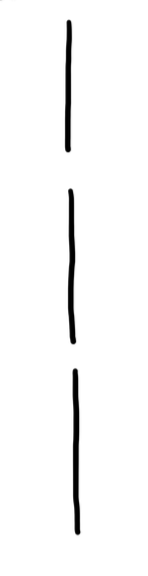}} \\
      $$
    \end{tabular} & 
    \begin{tabular}{c}
      \parbox[t][1.8cm][c]{0.12\textwidth}{\centering \includegraphics[width=0.01\textwidth]{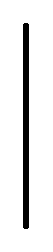}} \\
      $\failmark$
    \end{tabular} &
    \begin{tabular}{c}
      \parbox[t][1.8cm][c]{0.12\textwidth}{\centering \includegraphics[width=0.1\textwidth]{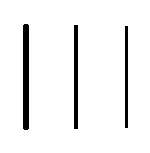}} \\
      $\failmark$
    \end{tabular} &
    \begin{tabular}{c}
      \parbox[t][1.8cm][c]{0.12\textwidth}{\centering \includegraphics[width=0.1\textwidth]{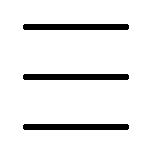}} \\
      $\failmark$
    \end{tabular} &
    \begin{tabular}{c}
      \parbox[t][1.8cm][c]{0.12\textwidth}{\centering \includegraphics[width=0.1\textwidth]{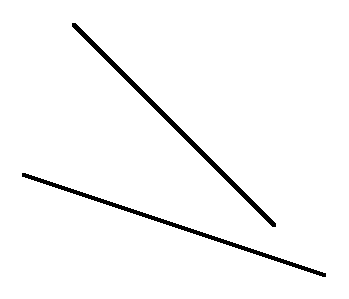}} \\
      $\failmark$
    \end{tabular} &
    \begin{tabular}{c}
      \parbox[t][1.8cm][c]{0.12\textwidth}{\centering \includegraphics[width=0.1\textwidth]{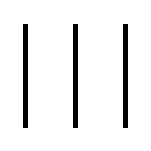}} \\
      $\failmark$
    \end{tabular} \\
    \midrule
    \begin{tabular}{c}
      \parbox[t][1.8cm][c]{0.1\textwidth}{\centering \includegraphics[width=0.1\textwidth]{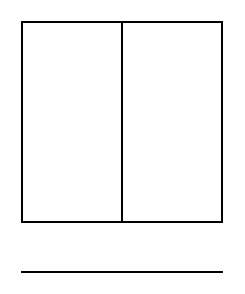}} \\
      $$
    \end{tabular} & 
    \begin{tabular}{c}
      \parbox[t][1.8cm][c]{0.12\textwidth}{\centering \includegraphics[width=0.05\textwidth]{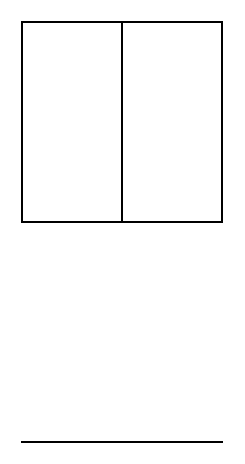}} \\
      $\failmark$
    \end{tabular} &
    \begin{tabular}{c}
      \parbox[t][1.8cm][c]{0.12\textwidth}{\centering \includegraphics[width=0.1\textwidth]{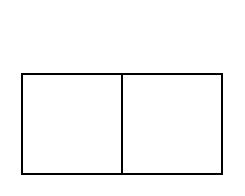}} \\
      $\failmark$
    \end{tabular} &
    \begin{tabular}{c}
      \parbox[t][1.8cm][c]{0.12\textwidth}{\centering \includegraphics[width=0.1\textwidth]{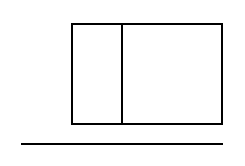}} \\
      $\failmark$
    \end{tabular} &
    \begin{tabular}{c}
      \parbox[t][1.8cm][c]{0.12\textwidth}{\centering \includegraphics[width=0.1\textwidth]{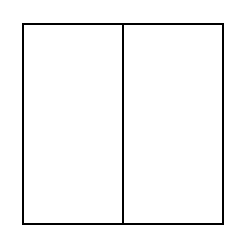}} \\
      $\failmark$
    \end{tabular} &
    \begin{tabular}{c}
      \parbox[t][1.8cm][c]{0.12\textwidth}{\centering \includegraphics[width=0.07\textwidth]{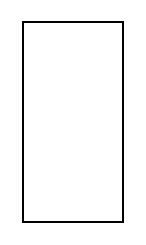}} \\
      $\failmark$
    \end{tabular} \\
    \midrule
    \begin{tabular}{c}
      \parbox[t][1.8cm][c]{0.1\textwidth}{\centering \includegraphics[width=0.1\textwidth]{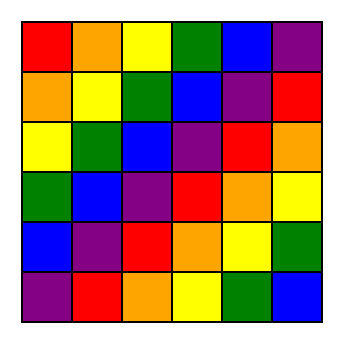}} \\
      $$
    \end{tabular} & 
    \begin{tabular}{c}
      \parbox[t][1.8cm][c]{0.12\textwidth}{\centering \includegraphics[width=0.1\textwidth]{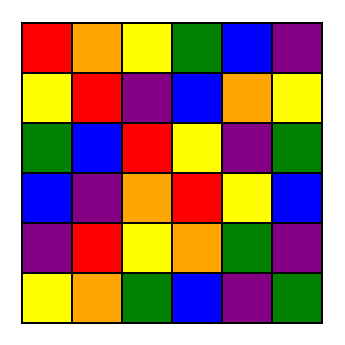}} \\
      $\failmark$
    \end{tabular} &
    \begin{tabular}{c}
      \parbox[t][1.8cm][c]{0.12\textwidth}{\centering N.A.} \\
      $\failmark$
    \end{tabular} &
    \begin{tabular}{c}
      \parbox[t][1.8cm][c]{0.12\textwidth}{\centering \includegraphics[width=0.1\textwidth]{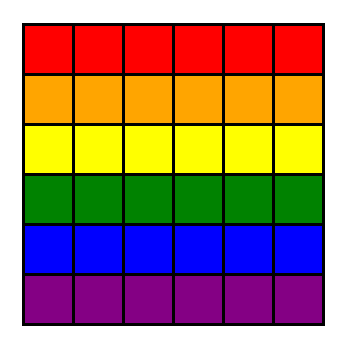}} \\
      $\failmark$
    \end{tabular} &
    \begin{tabular}{c}
      \parbox[t][1.8cm][c]{0.12\textwidth}{\centering \includegraphics[width=0.1\textwidth]{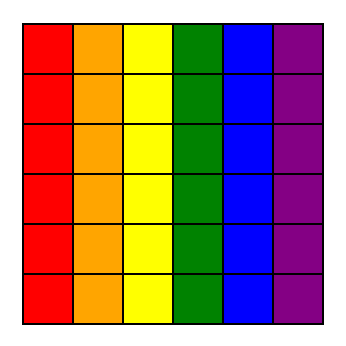}} \\
      $\failmark$
    \end{tabular} &
    \begin{tabular}{c}
      \parbox[t][1.8cm][c]{0.12\textwidth}{\centering \includegraphics[width=0.1\textwidth]{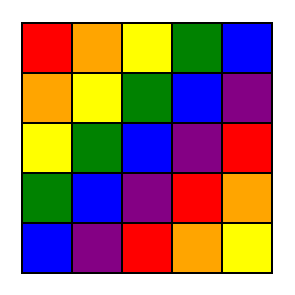}} \\
      $\failmark$
    \end{tabular} \\
    \midrule
    \begin{tabular}{c}
      \parbox[t][1.8cm][c]{0.1\textwidth}{\centering \includegraphics[width=0.1\textwidth]{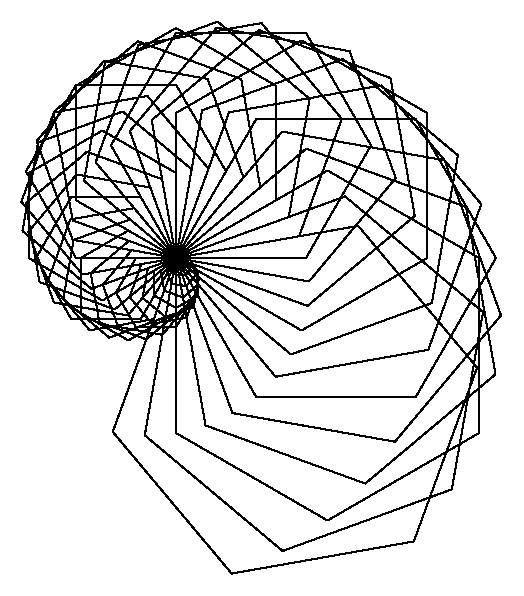}} \\
      $$
    \end{tabular} & 
    \begin{tabular}{c}
      \parbox[t][1.8cm][c]{0.12\textwidth}{\centering \includegraphics[width=0.1\textwidth]{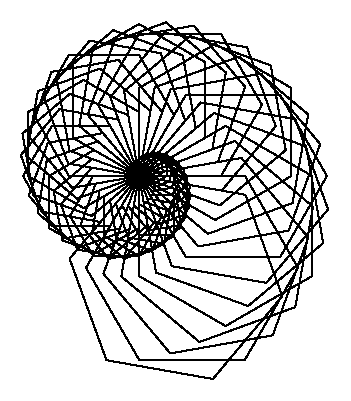}} \\
      $\failmark$
    \end{tabular} &
    \begin{tabular}{c}
      \parbox[t][1.8cm][c]{0.12\textwidth}{\centering \includegraphics[width=0.1\textwidth]{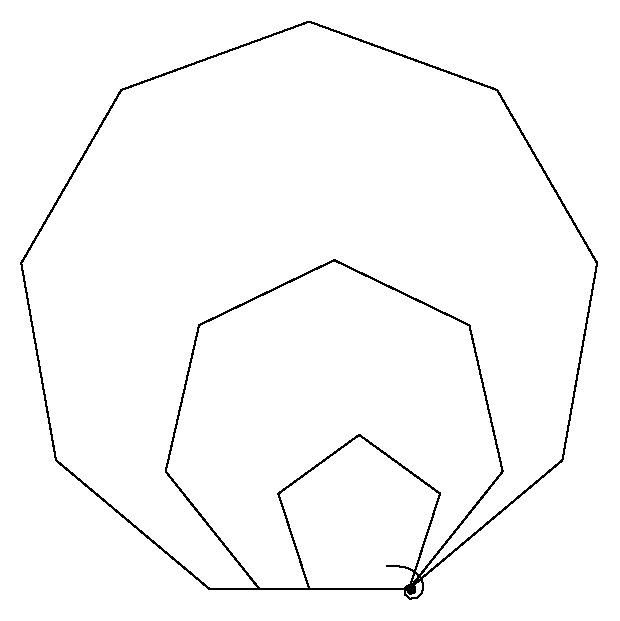}} \\
      $\failmark$
    \end{tabular} &
    \begin{tabular}{c}
      \parbox[t][1.8cm][c]{0.12\textwidth}{\centering \includegraphics[width=0.1\textwidth]{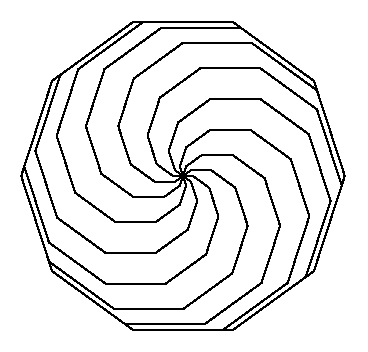}} \\
      $\failmark$
    \end{tabular} &
    \begin{tabular}{c}
      \parbox[t][1.8cm][c]{0.12\textwidth}{\centering \includegraphics[width=0.1\textwidth]{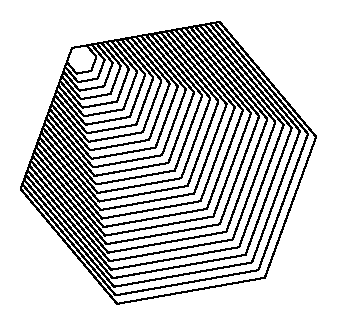}} \\
      $\failmark$
    \end{tabular} &
    \begin{tabular}{c}
      \parbox[t][1.8cm][c]{0.12\textwidth}{\centering \includegraphics[width=0.1\textwidth]{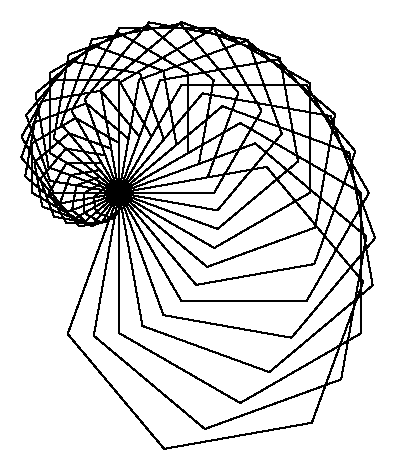}} \\
      $\failmark$
    \end{tabular} \\
    \midrule
    \begin{tabular}{c}
      \parbox[t][1.8cm][c]{0.1\textwidth}{\centering \includegraphics[width=0.1\textwidth]{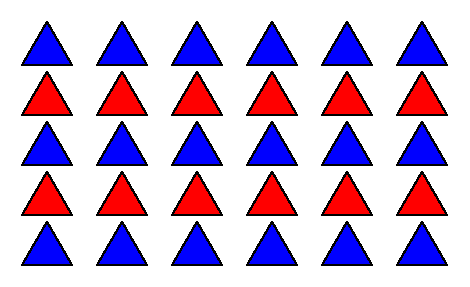}} \\
      $$
    \end{tabular} & 
    \begin{tabular}{c}
      \parbox[t][1.8cm][c]{0.12\textwidth}{\centering \includegraphics[width=0.1\textwidth]{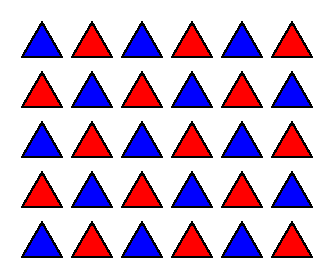}} \\
      $\failmark$
    \end{tabular} &
    \begin{tabular}{c}
      \parbox[t][1.8cm][c]{0.12\textwidth}{\centering \includegraphics[width=0.05\textwidth]{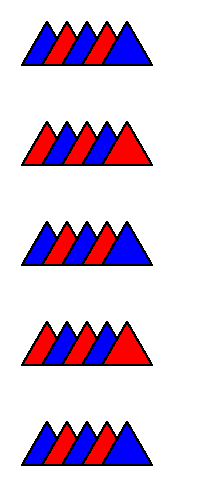}} \\
      $\failmark$
    \end{tabular} &
    \begin{tabular}{c}
      \parbox[t][1.8cm][c]{0.12\textwidth}{\centering \includegraphics[width=0.1\textwidth]{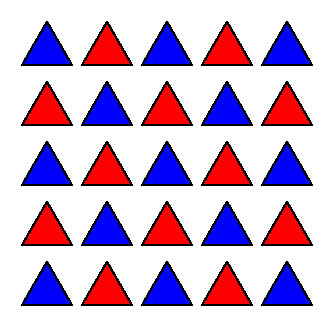}} \\
      $\failmark$
    \end{tabular} &
    \begin{tabular}{c}
      \parbox[t][1.8cm][c]{0.12\textwidth}{\centering \includegraphics[width=0.1\textwidth]{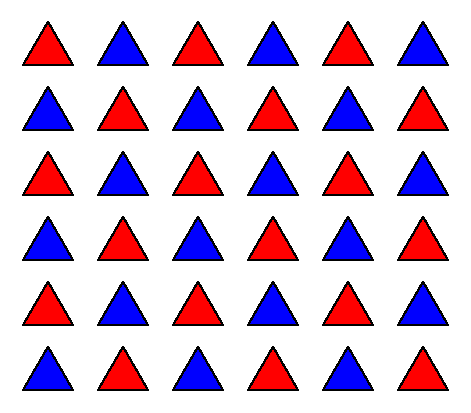}} \\
      $\failmark$
    \end{tabular} &
    \begin{tabular}{c}
      \parbox[t][1.8cm][c]{0.12\textwidth}{\centering \includegraphics[width=0.1\textwidth]{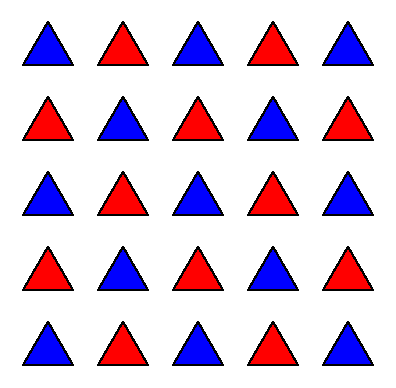}} \\
      $\failmark$
    \end{tabular} \\
    \bottomrule
  \end{tabular}
  }
  \caption{Tasks that are not solved by any of the models (i.e., \gptfouro{}, \pixtral{}-12B, \qwentwovl{}-72B, \pixtral{}-12B-\ft{}, \qwentwovl{}-72B-\ft{}) in the \datasetAll{} dataset. A total of $591$ tasks ($71.81\%$) match this criteria.
  Each row shows a ground truth image (leftmost) followed by the corresponding images generated by executing the each model's generated Python code. Success ($\successmark$) and failure ($\failmark$) are determined by our evaluation framework using symbolic comparison.
  }
  \label{fig:failure_examples_all_fail}
\end{table*}

%% file: appendix-figs/fig_failure_cases_code_examples.tex
% !TEX root =  main.tex
%%%%%%%%%%%%%%%%%%%%%%%%%%%%%%%%%%%%%%%%%%%%%%%
%%%%%%%%%%%%%%%%%%%%%%%%%%%%%%%%%%%%%%%%%%%%%%%

\begin{figure*}[t]
  \centering
  \begin{subfigure}{0.9\textwidth}
    \centering
    \scalebox{0.7}{
    \begin{tabular}{c|ccc}
      \toprule
    Input Image & \gptfouro & \qwentwovl{}-72B & \qwentwovl{}-72B-\ft{} \\
    \midrule
    \midrule
    \begin{tabular}{c}
      \parbox[t][1.8cm][c]{0.1\textwidth}{\centering \includegraphics[width=0.1\textwidth]{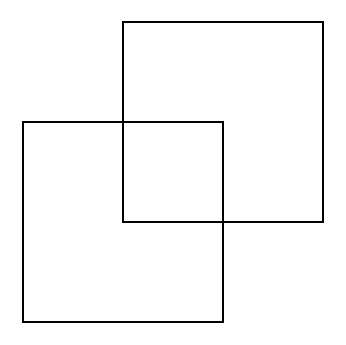}} \\
      $$
    \end{tabular} & 
    \begin{tabular}{c}
      \parbox[t][1.8cm][c]{0.12\textwidth}{\centering \includegraphics[width=0.1\textwidth]{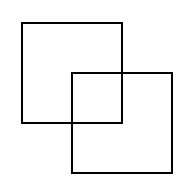}} \\
      \failmark
    \end{tabular} &
    \begin{tabular}{c}
      \parbox[t][1.8cm][c]{0.12\textwidth}{\centering \includegraphics[width=0.1\textwidth]{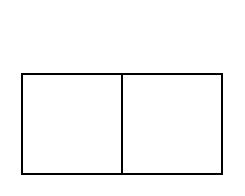}} \\
      \failmark
    \end{tabular} &
    \begin{tabular}{c}
      \parbox[t][1.8cm][c]{0.12\textwidth}{\centering \includegraphics[width=0.1\textwidth]{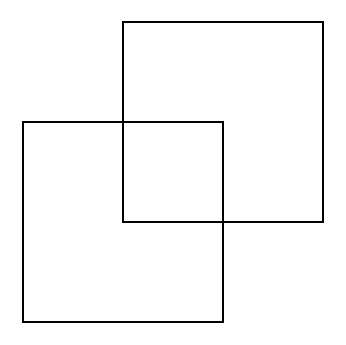}} \\
      \successmark
    \end{tabular} \\
    \bottomrule
  \end{tabular}
    }
    \caption{Input images and the generated images by different models.}
  \end{subfigure}

  \vspace{0.2cm}
  
  \begin{subfigure}{0.31\textwidth}
    \centering
    \begin{lstlisting}[breaklines=true,breakindent=0pt,xleftmargin=0cm,xrightmargin=0cm,framexleftmargin=0.1cm,framexrightmargin=0.1cm]
Here is the Python code to replicate the given geometric pattern using the Turtle Graphics library. The pattern consists of two overlapping squares.

```python
def draw(t):
    """Draw two overlapping squares."""
    def draw_square(t, size):
        """Helper function to draw a square of a given size."""
        for _ in range(4):
            t.forward(size)
            t.right(90)

    # Set up the turtle
    t.penup()
    t.goto(-50, 50)  # Move to the starting position for the first square
    t.pendown()

    # Draw the first square
    draw_square(t, 100)

    # Move to the starting position for the second square
    t.penup()
    t.goto(0, 0)
    t.pendown()
    # Draw the second square
    draw_square(t, 100)
```

### Explanation:
1. **`draw_square` Function**:
    - A helper function to draw a square of a given size.
    - The turtle moves forward by the specified size and turns 90 degrees to the right four times to complete the square.

2. **Positioning**:
    - The turtle starts at `(-50, 50)` to draw the first square.
    - After completing the first square, the turtle moves to `(0, 0)` to draw the second square, ensuring the overlap.

3. **Overlapping**:
    - The two squares overlap at their bottom-left and top-right corners, as shown in the image.

### Usage:
To execute the code, use the following:
```python
import turtle
# Create a turtle object
t = turtle.Turtle()
# Call the draw function
draw(t)
# Keep the window open
turtle.done()
```
    \end{lstlisting}
    \caption{\gptfouro{}}
  \end{subfigure}%
  \hfill%
  \begin{subfigure}{0.35\textwidth}
    \centering
    \begin{lstlisting}[breaklines=true,breakindent=0pt,xleftmargin=0cm,xrightmargin=0cm,framexleftmargin=0.1cm,framexrightmargin=0.1cm]
```python
import turtle

def draw(t):
    """Draw the specified geometric pattern."""
    # Set the turtle's speed
    t.speed(0)
    
    # Define the side length of the squares
    side_length = 100
    
    # Draw the first square
    t.penup()
    t.goto(-side_length / 2, -side_length / 2)
    t.pendown()
    for _ in range(4):
        t.forward(side_length)
        t.right(90)
    
    # Draw the second square
    t.penup()
    t.goto(side_length / 2, -side_length / 2)
    t.pendown()
    for _ in range(4):
        t.forward(side_length)
        t.right(90)

# Create a turtle object
t = turtle.Turtle()
draw(t)

# Keep the window open
turtle.done()
```

This code defines a function `draw(t)` that uses a turtle object `t` to draw the specified geometric pattern. The pattern consists of two overlapping squares, each with a side length of 100 units. The turtle starts at the center of the screen and moves to the starting position of each square before drawing it. The `turtle.done()` function is used to keep the window open after the drawing is complete.

$$
    \end{lstlisting}
    \caption{\qwentwovl{}-72B}
  \end{subfigure}%
  \hfill%
  \begin{subfigure}{0.31\textwidth}
    \centering
    \begin{lstlisting}[breaklines=true,breakindent=0pt,xleftmargin=0cm,xrightmargin=0cm,framexleftmargin=0.1cm,framexrightmargin=0.1cm]
## Image Description

The image displays two overlapping squares. The squares are of equal size and are positioned such that one square is shifted diagonally relative to the other. The overlapping area forms a smaller square in the center, while the non-overlapping parts of each square extend outward, creating a visually symmetrical pattern. The lines are black, and the background is white.

## Analysis & Solution Code

To create the pattern shown in the image using Turtle Graphics, we need to draw two squares of the same size. The first square is drawn starting from the initial position. After completing the first square, the turtle is moved diagonally to a new position to draw the second square. This diagonal movement ensures that the two squares overlap, creating the desired pattern.

Here is the code with comprehensive docstrings and comments to create the pattern:

```python
def draw(t):
    """
    Draws two overlapping squares of the same size (200 units) with a diagonal offset.

    Args:
        t: Turtle graphics object
    """
    def draw_square(size):
        """
        Draws a square with the given side length.

        Args:
            size: The length of each side of the square.
        """
        for _ in range(4):
            t.forward(size)  # Move the turtle forward by the side length
            t.left(90)       # Turn the turtle left by 90 degrees to form a corner

    # Draw the first square
    draw_square(200)
    # Move the turtle diagonally to the starting position of the second square
    t.penup()               # Lift the pen to move without drawing
    t.goto(100, 100)        # Move to the new starting position
    t.pendown()             # Put the pen down to start drawing again
    # Draw the second square
    draw_square(200)
```
    \end{lstlisting}
    \caption{\qwentwovl{}-72B-\ft{}}
  \end{subfigure}
  \vspace{-0.3em}
  \caption{Example outputs generated by \gptfouro{}, \qwentwovl{}-72B, and \qwentwovl{}-72B-\ft{}. (a) shows the input image and the generated images by executing the code generated by models. (b-d) show the outputs generated by different models.}
  \label{fig:failure_examples_code_examples1}
\end{figure*}

\clearpage

\begin{figure*}[t]
  \centering
  \begin{subfigure}{0.9\textwidth}
    \centering
    \scalebox{0.8}{
    \begin{tabular}{c|ccc}
      \toprule
    Input Image & \gptfouro & \qwentwovl{}-72B & \qwentwovl{}-72B-\ft{} \\
    \midrule
    \midrule
    \begin{tabular}{c}
      \parbox[t][1.8cm][c]{0.1\textwidth}{\centering \includegraphics[width=0.1\textwidth]{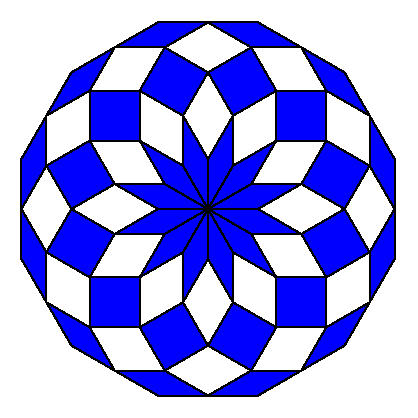}} \\
      $$
    \end{tabular} & 
    \begin{tabular}{c}
      \parbox[t][1.8cm][c]{0.1\textwidth}{\centering \includegraphics[width=0.1\textwidth]{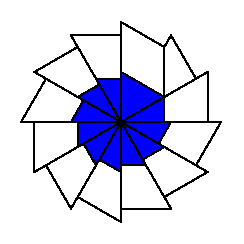}} \\
      \failmark
    \end{tabular} &
    \begin{tabular}{c}
      \parbox[t][1.8cm][c]{0.1\textwidth}{\centering N.A.} \\
      \failmark
    \end{tabular} &
    \begin{tabular}{c}
      \parbox[t][1.8cm][c]{0.1\textwidth}{\centering \includegraphics[width=0.1\textwidth]{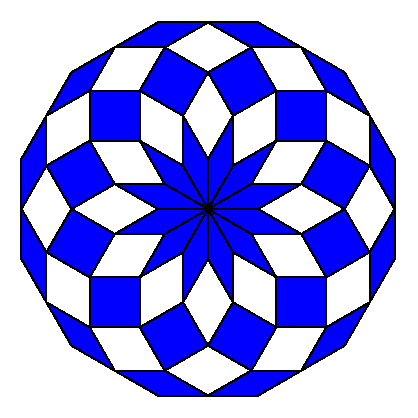}} \\
      \successmark
    \end{tabular} \\
    \bottomrule
  \end{tabular}
    }
    \caption{Input images and the generated images by different models.}
  \end{subfigure}

  \vspace{0.2cm}
  
  \begin{subfigure}{0.572\textwidth}
    \centering
    \begin{lstlisting}[breaklines=true,breakindent=0pt,xleftmargin=0cm,xrightmargin=0cm,framexleftmargin=0.1cm,framexrightmargin=0.1cm]
To replicate the geometric pattern shown in the image using Python's Turtle Graphics library, we need to break the pattern into smaller components. The pattern consists of a circular arrangement of blue and white diamond shapes, forming a star-like structure.

Here is the Python code to recreate the pattern:

```python
def draw(t):
    import turtle
    import math

    # Helper function to draw a diamond shape
    def draw_diamond(t, size, color):
        t.fillcolor(color)
        t.begin_fill()
        for _ in range(2):
            t.forward(size)
            t.left(60)
            t.forward(size)
            t.left(120)
        t.end_fill()

    # Helper function to draw a single segment of the pattern
    def draw_segment(t, size):
        # Draw the blue diamond
        draw_diamond(t, size, "blue")
        # Position for the white diamond
        t.forward(size)
        t.left(60)
        # Draw the white diamond
        draw_diamond(t, size, "white")
        # Return to the center
        t.right(60)
        t.backward(size)

    # Main function to draw the full pattern
    def draw_pattern(t, size, segments):
        for _ in range(segments):
            draw_segment(t, size)
            t.right(360 / segments)

    # Set up the turtle
    t.speed(0)
    t.penup()
    t.goto(0, 0)
    t.pendown()

    # Draw the pattern
    draw_pattern(t, 50, 12)  # Adjust size and number of segments as needed

    # Hide the turtle
    t.hideturtle()
```

### Explanation:

$\textcolor{blue}{\text{<some text is omitted here to save space>}}$

### Usage:

$\textcolor{blue}{\text{<some text is omitted here to save space>}}$

This code will produce the desired geometric pattern. Adjust the `size` and `segments` parameters in the `draw_pattern` function to modify the pattern's appearance.
    \end{lstlisting}
    \caption{\gptfouro{}}
  \end{subfigure}
  \hfill
  \begin{subfigure}{0.415\textwidth}
    \centering
    \begin{lstlisting}[breaklines=true,breakindent=0pt,xleftmargin=0cm,xrightmargin=0cm,framexleftmargin=0.1cm,framexrightmargin=0.1cm]
## Image Description

The image displays a symmetrical, flower-like pattern composed of geometric shapes. The primary shape is a dodecagon (a 12-sided polygon) repeated in a circular arrangement. Each dodecagon is filled with a blue color, and the spaces between the dodecagons are white, creating a checkerboard-like effect. The pattern exhibits rotational symmetry, with each dodecagon rotated slightly relative to its neighbors, forming a visually appealing star-like structure at the center. The overall design is intricate yet balanced, with the blue and white sections contrasting sharply against each other.

## Analysis & Solution Code

To create the pattern shown in the image using Turtle Graphics, we need to follow these steps:

1. **Set Up the Turtle**: Initialize the turtle graphics and set the fill color to blue.
2. **Define the Dodecagon**: Create a function to draw a single dodecagon. A dodecagon has 12 sides, and each internal angle is 150 degrees (calculated as \( \frac{180 \times (12-2)}{12} \)).
3. **Draw the Pattern**: Use a loop to draw 12 dodecagons, each rotated by 30 degrees (360 degrees / 12) relative to the previous one. This rotation ensures the symmetrical arrangement of the dodecagons.
4. **Fill the Shape**: Use the `begin_fill` and `end_fill` methods to fill the dodecagons with the specified color.

Here is the code with comprehensive docstrings and comments to create the pattern:

```python
def draw(t):
    """
    Draws a symmetrical, flower-like pattern composed of 12 blue-filled dodecagons.
    Each dodecagon is rotated by 30 degrees relative to the previous one, creating a
    visually appealing star-like structure at the center.

    Args:
        t: Turtle graphics object
    """
    t.fillcolor('blue')  # Set the fill color to blue

    def draw_dodecagon():
        """
        Draws a single dodecagon with 12 sides.
        Each side is 50 units long, and the internal angle is 150 degrees.
        """
        for _ in range(12):
            t.forward(50)  # Move the turtle forward by 50 units
            t.left(30)     # Turn the turtle left by 30 degrees

    t.begin_fill()  # Start filling the shape
    for _ in range(12):
        draw_dodecagon()  # Draw a dodecagon
        t.left(30)        # Rotate the turtle by 30 degrees for the next dodecagon
    t.end_fill()    # End filling the shape
```
    \end{lstlisting}
    \caption{\qwentwovl{}-72B-\ft{}}
  \end{subfigure}
  \caption{Example outputs generated by \gptfouro{} and \qwentwovl{}-72B-\ft{}. (a) shows the input image and the generated images by executing the code generated by models. (b) and (c) show the outputs generated by different models. \qwentwovl{}-72B's output is not shown since it generates repetitive text.}
  \label{fig:failure_examples_code_examples2}
\end{figure*}

%% file: appendix-figs/fig_prompt_code_synthesis.tex
% !TEX root =  main.tex
%%%%%%%%%%%%%%%%%%%%%%%%%%%%%%%%%%%%%%%
%%%%%%%%%%%%%%%%%%%%%%%%%%%%%%%%%%%%%%%

\begin{figure*}[h!]
    \centering
    \setlength\tabcolsep{5pt}
    \renewcommand{\arraystretch}{1.2}
    \input{appendix-figs/fig_prompt_code_synthesis_text}
    \caption{Prompt template for code synthesis from visual input.}
    \label{fig:appendix:prompt_template:code_synthesis}
\end{figure*}
%%%%%%%%%%%%%%%%%%%%%%%%%%%%%%%%%%%%%%% 

%% file: appendix-figs/fig_prompt_code_synthesis_text.tex
% !TEX root =  main.tex
%%%%%%%%%%%%%%%%%%%%%%%%%%%%%%%%%%%%%%%
%%%%%%%%%%%%%%%%%%%%%%%%%%%%%%%%%%%%%%%
\tiny
\begin{tabular}{|p{0.9\linewidth}|}
    \hline
    \\[-6pt]
    \multicolumn{1}{|c|}{\promptheader{Prompt for Generating Code from Image}} \\

    \begin{alltt}
\textbf{\textcolor{red}{<image>}} You are a Turtle Graphics programmer tasked with creating Python code to replicate a specified geometric 
pattern using the Turtle Graphics library.

### Task Overview:

Analyze the provided image of a geometric pattern. Carefully break down the pattern into individual shapes, colors, 
angles, and layout components. Using this information, write Python code within a function called `draw(t)`, 
where `t` is a Turtle object. Assume:
- The turtle starts at the center of the screen at coordinates `(0, 0)`.
- The turtle initially faces east (to the right).

The goal is for your `draw(t)` function to accurately recreate the pattern shown in the image, including its positioning, 
angles, colors, and details.

### Requirements:

1. \textbf{Code Structure}:
    - Place all code inside the `draw(t)` function.
    - The function takes a turtle object `t` as input.
    - Format your code using triple backticks with the 'python' language specifier, i.e., ```python```.

    Example format:
    ```python
    def draw(t):
        # Your code here
    ```
2. \textbf{Color Accuracy}:
    - Match colors in the image exactly, both for fills and outlines.
3. \textbf{Pattern Precision}:
    - Reproduce the pattern as accurately as possible, maintaining symmetry, shapes, and angles.
4. \textbf{Self-Contained}:
    - Do not include code outside the `draw(t)` function.
    - All necessary imports, variables, and helper functions should be inside `draw(t)`.

### Execution Context:

Your `draw(t)` function will be called in the following manner:

```python
import turtle

def draw(t):
    # Describe the drawing steps here
    pass

t = turtle.Turtle()
draw(t)
```

### Example Outputs:

- \textbf{Example 1 - Drawing a Rectangle}:
```python
def draw(t):
    """Draw a rectangle."""
    def draw_rectangle(t):
        # Draw a rectangle with side length 10
        for _ in range(4):
            t.forward(10)
            t.right(90)
    draw_rectangle(t)
```

- \textbf{Example 2 - Drawing a circle}:
```python
def draw(t):
    """Draw a circle."""
    import math

    def draw_circle(t, radius):
        circumference = 2 * math.pi * radius
        step_length = circumference / 360
        step_angle = 1      

        for _ in range(360):
            t.forward(step_length)
            t.left(step_angle)

    draw_circle(t, 50)
```

\emph{Note:} The examples are simplified. Your final code may require nested loops or additional logic to fully replicate 
complex patterns.

Now, write the code for the `draw(t)` function to recreate the pattern shown in the image as closely as possible in terms 
of shape, color, and structure.
    \end{alltt} \\
    \hline
\end{tabular}
%%%%%%%%%%%%%%%%%%%%%%%%%%%%%%%%%%%%%%% 

%% file: appendix-figs/fig_prompt_one_shot_code_generation.tex
% !TEX root =  main.tex
%%%%%%%%%%%%%%%%%%%%%%%%%%%%%%%%%%%%%%%
%%%%%%%%%%%%%%%%%%%%%%%%%%%%%%%%%%%%%%%

\begin{figure*}[h!]
    \centering
    \setlength\tabcolsep{5pt}
    \renewcommand{\arraystretch}{1.2}
    \input{appendix-figs/fig_prompt_one_shot_code_generation_text}
    \caption{Prompt template for reference-guided code generation.}
    \label{fig:prompt_template_one_shot_code_generation.example1}
\end{figure*}
%%%%%%%%%%%%%%%%%%%%%%%%%%%%%%%%%%%%%%%

%% file: appendix-figs/fig_prompt_one_shot_code_generation_text.tex
% !TEX root =  main.tex
%%%%%%%%%%%%%%%%%%%%%%%%%%%%%%%%%%%%%%%
%%%%%%%%%%%%%%%%%%%%%%%%%%%%%%%%%%%%%%%
\tiny
\begin{tabular}{|p{0.9\linewidth}|}
    \hline
    \\[-6pt]
    \multicolumn{1}{|c|}{\promptheader{Prompt for the Reference-guided Code Mutation Stage of \datagenOurs{}}} \\

    \begin{alltt}
You are a turtle graphics programmer tasked with \textbf{analyzing and applying code adaptations} in Python using the Turtle 
Graphics library. You are given \textbf{two reference codes} that perform a certain drawing task. Your mission is to:

1. \textbf{Identify how the adaptation is done} from the first code to the second code.
2. \textbf{Summarize the adaptation} in a \textbf{high-level way}, so it can be applied to any other code.
3. \textbf{Apply the core idea of the adaptation} to a new piece of code provided.

### \textbf{Key Requirements for Code Adaptation}:

1. Syntactic Correctness:
   - The adapted code must be \textbf{syntactically correct} and free of errors.

2. Structural and Logical Consistency:
   - Maintain the \textbf{structural integrity} and \textbf{logical flow} of the original code.
   - Ensure that no unintended behavior is introduced by the adaptation.

3. Geometric Structure & Symmetry (if applicable):
   - Ensure that all drawings consist of \textbf{clear geometric shapes} with \textbf{symmetry} and \textbf{geometric accuracy}.

4. Visual Clarity & Simplicity:
   - The output should be \textbf{visually clear} and \textbf{simple}.
   - Avoid overly complex designs that may confuse or clutter the output.

5. Function and Code Requirements:
   - Define the function `draw(t)` that contains all the drawing code.
   - Use appropriate Turtle Graphics library commands within the `draw(t)` function.
   - Only provide the `draw(t)` function. \textbf{Do not include import statements} or other code outside of the `draw()` function.

6. Different Output:
- The \textbf{adapted code must generate a different drawing} compared to the original new code.
- The drawing must be a different shape or have a distinct pattern to clearly show the adaptation's impact.

### Your Task:

\textbf{Reference Code 1}:
```python
\textbf{\textcolor{blue}{\{reference_code_1\}}}
```

\textbf{Reference Code 2}:
```python
\textbf{\textcolor{blue}{\{reference_code_2\}}}
```

\textbf{New Code to Adapt}:
```python
\textbf{\textcolor{blue}{\{code_to_adapt\}}}
```

Now, follow these steps:

1. Analyze the Adaptation:
   - Examine how \textbf{Reference Code 1} is adapted into \textbf{Reference Code 2}.
   - Summarize the adaptation in a \textbf{high-level way} that can be applied to other codes.

2. Apply the Adaptation:
   - Apply the core idea of the adaptation to the \textbf{New Code to Adapt}.
   - Provide the \textbf{Adapted Code} that reflects this adaptation.
   - Ensure the adapted code is \textbf{syntactically correct} and that the resulting drawing after execution meets all the 
   specified requirements (geometric structure, symmetry, visual clarity, simplicity, etc.).

\textbf{Adapted Code}:

Provide your adapted code here. Ensure it meets all the specified requirements, especially that it must generate a 
different drawing compared to the original new code. Use the following Python code block format:

```python
def draw(t):
   # Your adapted code here
```
    \end{alltt} \\
    \hline
\end{tabular}
%%%%%%%%%%%%%%%%%%%%%%%%%%%%%%%%%%%%%%%

%% file: appendix-figs/fig_prompt_eval_quality.tex
% !TEX root =  main.tex
%%%%%%%%%%%%%%%%%%%%%%%%%%%%%%%%%%%%%%%
%%%%%%%%%%%%%%%%%%%%%%%%%%%%%%%%%%%%%%%

\begin{figure*}[h!]
    \centering
    \setlength\tabcolsep{5pt}
    \renewcommand{\arraystretch}{1.2}
    \input{appendix-figs/fig_prompt_eval_quality_text}
    \caption{Prompt template for the elite selection stage in \datagenOurs{}.}
    \label{fig:prompt_template_eval_quality}
\end{figure*}
%%%%%%%%%%%%%%%%%%%%%%%%%%%%%%%%%%%%%%% 

%% file: appendix-figs/fig_prompt_eval_quality_text.tex
% !TEX root =  main.tex
%%%%%%%%%%%%%%%%%%%%%%%%%%%%%%%%%%%%%%%
%%%%%%%%%%%%%%%%%%%%%%%%%%%%%%%%%%%%%%%
\tiny
\begin{tabular}{|p{0.9\linewidth}|}
    \hline
    \\[-6pt]
    \multicolumn{1}{|c|}{\promptheader{Prompt for the Elite Selection Stage of \datagenOurs{}}} \\
    \begin{alltt}
\textbf{\textcolor{red}{<image>}} You are an evaluator responsible for automatically assessing the quality of a turtle graphics programming task 
using the following rubrics. Each rubric evaluates different aspects of the task, including its clarity, difficulty, 
alignment with programming concepts, and creativity. Please assign a score from 0 to 10 for each rubric and provide an 
explanation for your scoring. Each rubric has equal weight, and the rubrics are as follows:

### Rubrics Breakdown:

1. \textbf{Geometric Structure & Symmetry} 
- Score Breakdown (0-10):
  - 9-10: Perfect geometric accuracy and symmetry - all elements are precisely aligned and balanced.
  - 6-8: Mostly symmetric with minor imperfections - slight deviations that do not detract from overall symmetry.
  - 3-5: Some geometric or symmetry issues - noticeable asymmetries or inaccuracies in shape.
  - 0-2: Significant asymmetry and inaccuracies - major deviations from expected geometric forms.
2. \textbf{Visual Appeal, Clarity & Simplicity}  
- Score Breakdown (0-10):
  - 9-10: Clear, simple design with purposeful aesthetics - easily understood and visually harmonious. No unnecessary 
  complexity.
  - 6-8: Generally clear design with good balance, but has minor complexity or visual elements that could be simplified.
  - 3-5: Either overly complex, lacks visual harmony, or has clarity issues - may have unnecessary elements or 
  confusing design choices.
  - 0-2: Significant issues with clarity or complexity - cluttered, difficult to interpret, or contains many 
  unnecessary elements.
3. \textbf{Structural Coherence}  
- Score Breakdown (0-10):
  - 9-10: Strong structural integrity - design is cohesive, whether through repeated patterns, basic geometric 
  shapes, or a purposeful unique design.
  - 6-8: Good structure with minor imperfections - mostly coherent with slight inconsistencies.
  - 3-5: Basic structure present but with noticeable flaws - some elements may seem out of place or poorly integrated.
  - 0-2: Weak or unclear structure - lacks a clear organizational pattern or design logic.
4. \textbf{Alignment & Positioning}  
- Score Breakdown (0-10):
  - 9-10: Excellent alignment and positioning - all elements are precisely placed and aligned.
  - 6-8: Good alignment with minor issues - generally well-positioned with slight misalignments.
  - 3-5: Some misalignment - noticeable but not critical positioning errors.
  - 0-2: Noticeable misalignment - significant positioning errors that affect the overall design.
5. \textbf{Educational Value & Solvability}  
- Score Breakdown (0-10):
  - 9-10: Excellent educational value - pattern complexity is appropriate for learning, clear objectives, and perfectly 
  balanced difficulty that students can reasonably solve.
  - 6-8: Strong educational value - complexity is manageable for students with some guidance, mostly clear and 
  appropriately challenging without being overwhelming.
  - 3-5: Moderate educational value - either too simple to be educational or too complex for students to reasonably 
  solve. Would require significant modifications to be classroom-ready.
  - 0-2: Poor educational value - not suitable for classroom use due to excessive complexity, confusing structure, 
  or contains sensitive/inappropriate imagery (e.g., Swastika, Confederate flag, etc.).
6. \textbf{Color Usage & Necessity}  
- Score Breakdown (0-10):
  - 9-10: Excellent use of minimal colors - either black & white only, or uses very few colors (<5) with clear purpose 
  that enhances understanding.
  - 6-8: Acceptable color usage - slightly more colors than necessary but not distracting. Could be simplified without 
  losing meaning.
  - 3-5: Problematic color usage - too many colors or colors used without clear purpose. Would be clearer with fewer colors.
  - 0-2: Poor color usage - excessive number of colors, random color choices, or colors that make the pattern harder 
  to understand.

### Final Evaluation Instructions:

Once you have evaluated each category and assigned scores, sum up all the individual rubric scores. Since each rubric 
has equal weight, no additional multiplication is needed. The \textbf{final score} is simply the sum of all rubric scores. 
Summarize the individual scores and explanations, then provide the final score (out of 60).

### Expected JSON Output:

Please format the final evaluation as a JSON object using the following short keys:
- \textbf{geometry}: Geometric Structure & Symmetry
- \textbf{visual}: Visual Appeal & Clarity
- \textbf{structure}: Structural Coherence
- \textbf{alignment}: Alignment & Positioning
- \textbf{education}: Educational Value & Solvability
- \textbf{color}: Color Usage & Necessity
- \textbf{final\_score}: Final score out of 60

### Example JSON Output:
```json
\{  
  "geometry": \{"score": "<score>", "explanation": "<explanation>"\},
  "visual": \{"score": "<score>", "explanation": "<explanation>"\},
  "structure": \{"score": "<score>", "explanation": "<explanation>"\},
  "alignment": \{"score": "<score>", "explanation": "<explanation>"\},
  "education": \{"score": "<score>", "explanation": "<explanation>"\},
  "color": \{"score": "<score>", "explanation": "<explanation>"\},
  "final_score": "<final_score>"
\}
```
Now, evaluate the turtle graphics task based on the provided image. This image was created using turtle graphics. Please 
assess its quality using the rubrics outlined above, and provide the final evaluation in the JSON format shown in the 
example.
    \end{alltt} \\
    \hline
\end{tabular}
%%%%%%%%%%%%%%%%%%%%%%%%%%%%%%%%%%%%%%% 

%% file: appendix-figs/fig_prompt_code_relabel.tex
% !TEX root =  main.tex
%%%%%%%%%%%%%%%%%%%%%%%%%%%%%%%%%%%%%%%
%%%%%%%%%%%%%%%%%%%%%%%%%%%%%%%%%%%%%%%

\begin{figure*}[h!]
    \centering
    \setlength\tabcolsep{5pt}
    \renewcommand{\arraystretch}{1.2}
    \input{appendix-figs/fig_prompt_code_relabel_text}
    \caption{Prompt template for the CoT labeling for generating the training dataset.}
    \label{fig:prompt_template_code_relabel}
\end{figure*}
%%%%%%%%%%%%%%%%%%%%%%%%%%%%%%%%%%%%%%% 

%% file: appendix-figs/fig_prompt_code_relabel_text.tex
% !TEX root =  main.tex
%%%%%%%%%%%%%%%%%%%%%%%%%%%%%%%%%%%%%%%
%%%%%%%%%%%%%%%%%%%%%%%%%%%%%%%%%%%%%%%
\tiny
\begin{tabular}{|p{0.9\linewidth}|}
    \hline
    % some space
    \\[-6pt]
    \multicolumn{1}{|c|}{\promptheader{Prompt for the CoT Labeling}} \\

    \begin{alltt}
\textbf{\textcolor{red}{<image>}} Your task is to optimize a provided Python code snippet that uses Turtle Graphics, ensuring it is minimal, 
cleanly documented, and fully aligned with the generated image output. You will be provided with:

1. A Python code snippet using Turtle Graphics.
2. The actual image output generated by this code.

## Your Responsibilities:
    
1. \textbf{Describe the Image}  
    - Provide a detailed description of the visual pattern in the image \textbf{without referencing the code}, focusing on 
    geometric shapes, symmetry, colors, and overall structure.
    
2. \textbf{Optimize the Code}  
    - Identify and remove redundant code segments that do not contribute to the visual output, and simplify the logic to 
    enhance readability, ensuring the 
    \textbf{final output remains visually identical}.
    - After optimizing the code, provide a detailed step-by-step explanation of how the code generates the image, linking 
    visual features to the corresponding steps.
    
3. \textbf{Add Documentation and Comments}  
    - Add a descriptive docstring for the provided code snippet, explaining its purpose, parameters, and any outputs. 
    Include clear and concise inline comments to make the code understandable.

## Formatting Instructions:
    - \textbf{Markdown-Only Response}: Format your entire response in markdown, enclosed in a single markdown block.
    - \textbf{Output Focus}: Provide only the optimized `draw(t)` function within the markdown block, excluding any setup or 
    unrelated code.
    
## Provided Code Snippet
    
Below is the code snippet that generates the image you are analyzing:
    
```python
\textbf{\textcolor{blue}{\{code\}}}
```

Please provide your response in the following markdown format:

```markdown
## Image Description

The image displays...

[Provide a detailed description of the visual pattern]

## Analysis & Solution Code

To create the pattern shown in the image using Turtle Graphics, we need to...
[Explain how to create this pattern using Turtle Graphics, describing the logical steps needed to reproduce the image]

Here is the code with comprehensive docstrings and comments to create the pattern:
```python
def draw(t):
    """
    [Function description]
    
    Args:
        t: Turtle graphics object
    """
    # Your simplified code with comments
```
```

\textbf{Important}: Write your response as if you are only looking at the image, without referencing any provided code (e.g., do 
not mention `modified code`, `optimized code`, or `provided code` in your response inside the markdown block).
    \end{alltt} \\
    \hline
\end{tabular}
%%%%%%%%%%%%%%%%%%%%%%%%%%%%%%%%%%%%%%% 

%% file: main.bbl
\begin{thebibliography}{59}
\providecommand{\natexlab}[1]{#1}

\bibitem[{Agrawal et~al.(2024)Agrawal, Antoniak, Hanna, Bout, Chaplot, Chudnovsky, Costa, Monicault, Garg, Gervet, Ghosh, Héliou, Jacob, Jiang, Khandelwal, Lacroix, Lample, Casas, Lavril, Scao, Lo, Marshall, Martin, Mensch, Muddireddy, Nemychnikova, Pellat, Platen, Raghuraman, Rozière, Sablayrolles, Saulnier, Sauvestre, Shang, Soletskyi, Stewart, Stock, Studnia, Subramanian, Vaze, Wang, and Yang}]{agrawal2024pixtral12b}
Pravesh Agrawal, Szymon Antoniak, Emma~Bou Hanna, Baptiste Bout, Devendra Chaplot, Jessica Chudnovsky, Diogo Costa, Baudouin~De Monicault, Saurabh Garg, Theophile Gervet, Soham Ghosh, Amélie Héliou, Paul Jacob, Albert~Q. Jiang, Kartik Khandelwal, Timothée Lacroix, Guillaume Lample, Diego~Las Casas, Thibaut Lavril, and 23 others. 2024.
\newblock {Pixtral 12B}.
\newblock \emph{CoRR}, abs/2410.07073.

\bibitem[{Bai et~al.(2025{\natexlab{a}})Bai, Cai, Chen, Chen, Chen, Cheng, Deng, Ding, Gao, Ge, Ge, Guo, Huang, Huang, Huang, Hui, Jiang, Li, Li, Li, Li, Lin, Lin, Liu, Liu, Liu, Liu, Liu, Liu, Lu, Luo, Lv, Men, Meng, Ren, Ren, Song, Sun, Tang, Tu, Wan, Wang, Wang, Wang, Wang, Xie, Xu, Xu, Xu, Yang, Yang, Yang, Yang, Yu, Zhang, Zhang, Zhang, Zheng, Zhong, Zhou, Zhou, Zhou, Zhu, and Zhu}]{bai2025qwen3vltechnicalreport}
Shuai Bai, Yuxuan Cai, Ruizhe Chen, Keqin Chen, Xionghui Chen, Zesen Cheng, Lianghao Deng, Wei Ding, Chang Gao, Chunjiang Ge, Wenbin Ge, Zhifang Guo, Qidong Huang, Jie Huang, Fei Huang, Binyuan Hui, Shutong Jiang, Zhaohai Li, Mingsheng Li, and 45 others. 2025{\natexlab{a}}.
\newblock {Qwen3-VL Technical Report}.
\newblock \emph{CoRR}, abs/2511.21631.

\bibitem[{Bai et~al.(2025{\natexlab{b}})Bai, Chen, Liu, Wang, Ge, Song, Dang, Wang, Wang, Tang, Zhong, Zhu, Yang, Li, Wan, Wang, Ding, Fu, Xu, Ye, Zhang, Xie, Cheng, Zhang, Yang, Xu, and Lin}]{DBLP:journals/corr/abs-2502-13923}
Shuai Bai, Keqin Chen, Xuejing Liu, Jialin Wang, Wenbin Ge, Sibo Song, Kai Dang, Peng Wang, Shijie Wang, Jun Tang, Humen Zhong, Yuanzhi Zhu, Ming{-}Hsuan Yang, Zhaohai Li, Jianqiang Wan, Pengfei Wang, Wei Ding, Zheren Fu, Yiheng Xu, and 8 others. 2025{\natexlab{b}}.
\newblock {Qwen2.5-VL Technical Report}.
\newblock \emph{CoRR}, abs/2502.13923.

\bibitem[{Belouadi et~al.(2024)Belouadi, Ponzetto, and Eger}]{DBLP:conf/nips/BelouadiPE24}
Jonas Belouadi, Simone~Paolo Ponzetto, and Steffen Eger. 2024.
\newblock {DeTikZify: Synthesizing Graphics Programs for Scientific Figures and Sketches with TikZ}.
\newblock In \emph{{NeurIPS}}.

\bibitem[{Blockly(2022)}]{blocklygames}
Blockly. 2022.
\newblock {G}ames for {T}omorrow's {P}rogrammers.
\newblock \url{https://blockly.games/}.

\bibitem[{Chen et~al.(2023)Chen, Wu, Wang, Su, Chen, Xing, Zhong, Zhang, Zhu, Lu, Li, Luo, Lu, Qiao, and Dai}]{DBLP:journals/corr/abs-2312-14238}
Zhe Chen, Jiannan Wu, Wenhai Wang, Weijie Su, Guo Chen, Sen Xing, Muyan Zhong, Qinglong Zhang, Xizhou Zhu, Lewei Lu, Bin Li, Ping Luo, Tong Lu, Yu~Qiao, and Jifeng Dai. 2023.
\newblock {InternVL: Scaling up Vision Foundation Models and Aligning for Generic Visual-Linguistic Tasks}.
\newblock \emph{CoRR}, abs/2312.14238.

\bibitem[{{CodeHS}(2025)}]{codehs_tracy}
{CodeHS}. 2025.
\newblock {Turtle Graphics with Tracy the Turtle}.
\newblock \url{https://codehs.com/hourofcode/tracy}.
\newblock Accessed: 2025-09-01.

\bibitem[{Code.org(2013)}]{codeorg}
Code.org. 2013.
\newblock Code.org: {L}earn {C}omputer {S}cience.
\newblock \url{https://code.org/}.

\bibitem[{Dai et~al.(2024)Dai, Lee, Wang, Yang, Liu, Barker, Rintamaki, Shoeybi, Catanzaro, and Ping}]{DBLP:journals/corr/abs-2409-11402}
Wenliang Dai, Nayeon Lee, Boxin Wang, Zhuoling Yang, Zihan Liu, Jon Barker, Tuomas Rintamaki, Mohammad Shoeybi, Bryan Catanzaro, and Wei Ping. 2024.
\newblock {{NVLM:} Open Frontier-Class Multimodal LLMs}.
\newblock \emph{CoRR}, abs/2409.11402.

\bibitem[{Deitke et~al.(2024)Deitke, Clark, Lee, Tripathi, Yang, Park, Salehi, Muennighoff, Lo, Soldaini, Lu, Anderson, Bransom, Ehsani, Ngo, Chen, Patel, Yatskar, Callison{-}Burch, Head, Hendrix, Bastani, VanderBilt, Lambert, Chou, Chheda, Sparks, Skjonsberg, Schmitz, Sarnat, Bischoff, Walsh, Newell, Wolters, Gupta, Zeng, Borchardt, Groeneveld, Dumas, Nam, Lebrecht, Wittlif, Schoenick, Michel, Krishna, Weihs, Smith, Hajishirzi, Girshick, Farhadi, and Kembhavi}]{DBLP:journals/corr/abs-2409-17146}
Matt Deitke, Christopher Clark, Sangho Lee, Rohun Tripathi, Yue Yang, Jae~Sung Park, Mohammadreza Salehi, Niklas Muennighoff, Kyle Lo, Luca Soldaini, Jiasen Lu, Taira Anderson, Erin Bransom, Kiana Ehsani, Huong Ngo, Yen{-}Sung Chen, Ajay Patel, Mark Yatskar, Chris Callison{-}Burch, and 32 others. 2024.
\newblock {Molmo and PixMo: Open Weights and Open Data for State-of-the-Art Multimodal Models}.
\newblock \emph{CoRR}, abs/2409.17146.

\bibitem[{Ellis et~al.(2021)Ellis, Wong, Nye, Sabl{\'{e}}{-}Meyer, Morales, Hewitt, Cary, Solar{-}Lezama, and Tenenbaum}]{DBLP:conf/pldi/EllisWNSMHCST21}
Kevin Ellis, Catherine Wong, Maxwell~I. Nye, Mathias Sabl{\'{e}}{-}Meyer, Lucas Morales, Luke~B. Hewitt, Luc Cary, Armando Solar{-}Lezama, and Joshua~B. Tenenbaum. 2021.
\newblock {DreamCoder: bootstrapping inductive program synthesis with wake-sleep library learning}.
\newblock In \emph{{PLDI}}.

\bibitem[{Ester et~al.(1996)Ester, Kriegel, Sander, and Xu}]{DBLP:conf/kdd/EsterKSX96}
Martin Ester, Hans{-}Peter Kriegel, J{\"{o}}rg Sander, and Xiaowei Xu. 1996.
\newblock {A Density-Based Algorithm for Discovering Clusters in Large Spatial Databases with Noise}.
\newblock In \emph{KDD}.

\bibitem[{Grover and Pea(2013)}]{grover2013computational}
Shuchi Grover and Roy Pea. 2013.
\newblock {Computational thinking in K--12: A review of the state of the field}.
\newblock \emph{Educational researcher}.

\bibitem[{Gui et~al.(2025)Gui, Wan, Li, Zhang, Chen, Zhang, Su, Chen, Zhou, Jiang, and Zhang}]{DBLP:conf/www/Gui0LZC0SCZ0025}
Yi~Gui, Yao Wan, Zhen Li, Zhongyi Zhang, Dongping Chen, Hongyu Zhang, Yi~Su, Bohua Chen, Xing Zhou, Wenbin Jiang, and Xiangliang Zhang. 2025.
\newblock {UICopilot: Automating {UI} Synthesis via Hierarchical Code Generation from Webpage Designs}.
\newblock In \emph{{WWW}}.

\bibitem[{Gupta and Kembhavi(2023)}]{DBLP:conf/cvpr/GuptaK23}
Tanmay Gupta and Aniruddha Kembhavi. 2023.
\newblock {Visual Programming: Compositional visual reasoning without training}.
\newblock In \emph{CVPR}.

\bibitem[{He et~al.(2016)He, Zhang, Ren, and Sun}]{DBLP:conf/cvpr/HeZRS16}
Kaiming He, Xiangyu Zhang, Shaoqing Ren, and Jian Sun. 2016.
\newblock {Deep Residual Learning for Image Recognition}.
\newblock In \emph{CVPR}.

\bibitem[{Hu et~al.(2022)Hu, Shen, Wallis, Allen{-}Zhu, Li, Wang, Wang, and Chen}]{DBLP:conf/iclr/HuSWALWWC22}
Edward~J. Hu, Yelong Shen, Phillip Wallis, Zeyuan Allen{-}Zhu, Yuanzhi Li, Shean Wang, Lu~Wang, and Weizhu Chen. 2022.
\newblock {L}o{RA}: {L}ow-{R}ank {A}daptation of {L}arge {L}anguage {M}odels.
\newblock In \emph{{ICLR}}.

\bibitem[{Kwon et~al.(2023)Kwon, Li, Zhuang, Sheng, Zheng, Yu, Gonzalez, Zhang, and Stoica}]{DBLP:conf/sosp/KwonLZ0ZY0ZS23}
Woosuk Kwon, Zhuohan Li, Siyuan Zhuang, Ying Sheng, Lianmin Zheng, Cody~Hao Yu, Joseph Gonzalez, Hao Zhang, and Ion Stoica. 2023.
\newblock {E}fficient {M}emory {M}anagement for {L}arge {L}anguage {M}odel {S}erving with {P}aged{A}ttention.
\newblock In \emph{SIGOPS}.

\bibitem[{Li et~al.(2024{\natexlab{a}})Li, Zhang, Guo, Zhang, Li, Zhang, Zhang, Li, Liu, and Li}]{DBLP:journals/corr/abs-2408-03326}
Bo~Li, Yuanhan Zhang, Dong Guo, Renrui Zhang, Feng Li, Hao Zhang, Kaichen Zhang, Yanwei Li, Ziwei Liu, and Chunyuan Li. 2024{\natexlab{a}}.
\newblock {LLaVA-OneVision: Easy Visual Task Transfer}.
\newblock \emph{CoRR}, abs/2408.03326.

\bibitem[{Li et~al.(2024{\natexlab{b}})Li, Ding, Fang, and Tao}]{DBLP:conf/emnlp/Li0FT24}
Hongyu Li, Liang Ding, Meng Fang, and Dacheng Tao. 2024{\natexlab{b}}.
\newblock {Revisiting Catastrophic Forgetting in Large Language Model Tuning}.
\newblock In \emph{{EMNLP (Findings)}}.

\bibitem[{Li et~al.(2024{\natexlab{c}})Li, Tian, Hu, Luo, Huang, and Ma}]{DBLP:conf/emnlp/LiTHLH024}
Kaixin Li, Yuchen Tian, Qisheng Hu, Ziyang Luo, Zhiyong Huang, and Jing Ma. 2024{\natexlab{c}}.
\newblock {MMCode: Benchmarking Multimodal Large Language Models for Code Generation with Visually Rich Programming Problems}.
\newblock In \emph{{EMNLP (Findings)}}.

\bibitem[{Li and Ellis(2024)}]{DBLP:conf/nips/LiE24}
Wen{-}Ding Li and Kevin Ellis. 2024.
\newblock {Is Programming by Example Solved by LLMs?}
\newblock In \emph{{NeurIPS}}.

\bibitem[{Liu et~al.(2023)Liu, Xia, Wang, and Zhang}]{DBLP:conf/nips/LiuXW023}
Jiawei Liu, Chunqiu~Steven Xia, Yuyao Wang, and Lingming Zhang. 2023.
\newblock {Is Your Code Generated by ChatGPT Really Correct? Rigorous Evaluation of Large Language Models for Code Generation}.
\newblock In \emph{{NeurIPS}}.

\bibitem[{Lu et~al.(2024)Lu, Bansal, Xia, Liu, Li, Hajishirzi, Cheng, Chang, Galley, and Gao}]{DBLP:conf/iclr/LuBX0LH0CG024}
Pan Lu, Hritik Bansal, Tony Xia, Jiacheng Liu, Chunyuan Li, Hannaneh Hajishirzi, Hao Cheng, Kai{-}Wei Chang, Michel Galley, and Jianfeng Gao. 2024.
\newblock {MathVista: Evaluating Mathematical Reasoning of Foundation Models in Visual Contexts}.
\newblock In \emph{{ICLR}}.

\bibitem[{Maloney et~al.(2010)Maloney, Resnick, Rusk, Silverman, and Eastmond}]{DBLP:journals/jeric/MaloneyRRSE10}
John~H. Maloney, Mitchel Resnick, Natalie Rusk, Brian Silverman, and Evelyn Eastmond. 2010.
\newblock {The Scratch Programming Language and Environment}.
\newblock \emph{{ACM} Trans. Comput. Educ.}

\bibitem[{Marbach et~al.(2022)Marbach, Maximova, and Staub}]{DBLP:conf/issep/MarbachMS22}
Jeremy Marbach, Alexandra Maximova, and Jacqueline Staub. 2022.
\newblock {A Tool to Create and Conduct Custom Assessments in Turtle Graphics}.
\newblock In \emph{{ISSEP}}.

\bibitem[{OpenAI(2024{\natexlab{a}})}]{gpt4o}
OpenAI. 2024{\natexlab{a}}.
\newblock {GPT-4o}.
\newblock \url{https://openai.com/index/hello-gpt-4o/}.

\bibitem[{OpenAI(2024{\natexlab{b}})}]{gpt4v}
OpenAI. 2024{\natexlab{b}}.
\newblock {GPT-4v}.
\newblock \url{https://openai.com/index/gpt-4v-system-card/}.

\bibitem[{OpenAI(2025{\natexlab{a}})}]{gpt5}
OpenAI. 2025{\natexlab{a}}.
\newblock {GPT-5}.
\newblock \url{https://openai.com/gpt-5/}.

\bibitem[{OpenAI(2025{\natexlab{b}})}]{o3_and_o4_mini}
OpenAI. 2025{\natexlab{b}}.
\newblock o3 and o4-mini.
\newblock \url{https://openai.com/index/introducing-o3-and-o4-mini/}.

\bibitem[{Padurean and Singla(2024)}]{DBLP:journals/corr/abs-2406-09891}
Victor{-}Alexandru Padurean and Adish Singla. 2024.
\newblock {Benchmarking Generative Models on Computational Thinking Tests in Elementary Visual Programming}.
\newblock In \emph{NeurIPS Track on Datasets and Benchmarks}.

\bibitem[{Pea(1987)}]{pea1987logo}
Roy~D Pea. 1987.
\newblock {Logo programming and problem solving}.

\bibitem[{Python(2024)}]{python_turtle}
Python. 2024.
\newblock {Python Turtle Graphics}.
\newblock \url{https://docs.python.org/3/library/turtle.html}.

\bibitem[{Radford et~al.(2021)Radford, Kim, Hallacy, Ramesh, Goh, Agarwal, Sastry, Askell, Mishkin, Clark, Krueger, and Sutskever}]{DBLP:conf/icml/RadfordKHRGASAM21}
Alec Radford, Jong~Wook Kim, Chris Hallacy, Aditya Ramesh, Gabriel Goh, Sandhini Agarwal, Girish Sastry, Amanda Askell, Pamela Mishkin, Jack Clark, Gretchen Krueger, and Ilya Sutskever. 2021.
\newblock {L}earning {T}ransferable {V}isual {M}odels {F}rom {N}atural {L}anguage {S}upervision.
\newblock In \emph{{ICML}}.

\bibitem[{Ramesh et~al.(2021)Ramesh, Pavlov, Goh, Gray, Voss, Radford, Chen, and Sutskever}]{DBLP:conf/icml/RameshPGGVRCS21}
Aditya Ramesh, Mikhail Pavlov, Gabriel Goh, Scott Gray, Chelsea Voss, Alec Radford, Mark Chen, and Ilya Sutskever. 2021.
\newblock {Z}ero-{S}hot {T}ext-to-{I}mage {G}eneration.
\newblock In \emph{{ICML}}.

\bibitem[{Rismanchian et~al.(2025)Rismanchian, Razeghi, Singh, and Doroudi}]{rismanchian2024turtlebench}
Sina Rismanchian, Yasaman Razeghi, Sameer Singh, and Shayan Doroudi. 2025.
\newblock {TurtleBench: {A} Visual Programming Benchmark in Turtle Geometry}.
\newblock In \emph{{NAACL}}.

\bibitem[{Rodriguez et~al.(2025)Rodriguez, Puri, Agarwal, Laradji, Rajeswar, V{\'{a}}zquez, Pal, and Pedersoli}]{DBLP:conf/aaai/RodriguezPALR0P25}
Juan~A. Rodriguez, Abhay Puri, Shubham Agarwal, Issam~H. Laradji, Sai Rajeswar, David V{\'{a}}zquez, Christopher Pal, and Marco Pedersoli. 2025.
\newblock {StarVector: Generating Scalable Vector Graphics Code from Images and Text}.
\newblock In \emph{{AAAI}}.

\bibitem[{Rozi{\`{e}}re et~al.(2023)Rozi{\`{e}}re, Gehring, Gloeckle, Sootla, Gat, Tan, Adi, Liu, Remez, Rapin, Kozhevnikov, Evtimov, Bitton, Bhatt, Canton{-}Ferrer, Grattafiori, Xiong, D{\'{e}}fossez, Copet, Azhar, Touvron, Martin, Usunier, Scialom, and Synnaeve}]{DBLP:journals/corr/abs-2308-12950}
Baptiste Rozi{\`{e}}re, Jonas Gehring, Fabian Gloeckle, Sten Sootla, Itai Gat, Xiaoqing~Ellen Tan, Yossi Adi, Jingyu Liu, Tal Remez, J{\'{e}}r{\'{e}}my Rapin, Artyom Kozhevnikov, Ivan Evtimov, Joanna Bitton, Manish Bhatt, Cristian Canton{-}Ferrer, Aaron Grattafiori, Wenhan Xiong, Alexandre D{\'{e}}fossez, Jade Copet, and 6 others. 2023.
\newblock {C}ode {L}lama: {O}pen {F}oundation {M}odels for {C}ode.
\newblock \emph{CoRR}, abs/2308.12950.

\bibitem[{Si et~al.(2025)Si, Zhang, Li, Yang, Liu, and Yang}]{DBLP:conf/naacl/SiZLYLY25}
Chenglei Si, Yanzhe Zhang, Ryan Li, Zhengyuan Yang, Ruibo Liu, and Diyi Yang. 2025.
\newblock {Design2Code: Benchmarking Multimodal Code Generation for Automated Front-End Engineering}.
\newblock In \emph{{NAACL}}.

\bibitem[{Singla(2023)}]{DBLP:conf/icer/Singla22}
Adish Singla. 2023.
\newblock {E}valuating {C}hat{GPT} and {GPT-4} for {V}isual {P}rogramming.
\newblock In \emph{ICER - Volume 2}.

\bibitem[{Staub(2021)}]{DBLP:journals/eatcs/Staub21}
Jacqueline Staub. 2021.
\newblock {L}ogo {E}nvironments in the {F}ocus of {T}ime.
\newblock \emph{{B}ulletin of {EATCS}}.

\bibitem[{{Turtle Academy}(2025)}]{turtleacademy}
{Turtle Academy}. 2025.
\newblock {Turtle Academy}.
\newblock \url{https://turtleacademy.com/}.
\newblock Accessed: 2025-09-01.

\bibitem[{Wang et~al.(2024)Wang, Bai, Tan, Wang, Fan, Bai, Chen, Liu, Wang, Ge, Fan, Dang, Du, Ren, Men, Liu, Zhou, Zhou, and Lin}]{DBLP:journals/corr/abs-2409-12191}
Peng Wang, Shuai Bai, Sinan Tan, Shijie Wang, Zhihao Fan, Jinze Bai, Keqin Chen, Xuejing Liu, Jialin Wang, Wenbin Ge, Yang Fan, Kai Dang, Mengfei Du, Xuancheng Ren, Rui Men, Dayiheng Liu, Chang Zhou, Jingren Zhou, and Junyang Lin. 2024.
\newblock {Qwen2-VL: Enhancing Vision-Language Model's Perception of the World at Any Resolution}.
\newblock \emph{CoRR}, abs/2409.12191.

\bibitem[{Wei et~al.(2022)Wei, Wang, Schuurmans, Bosma, Ichter, Xia, Chi, Le, and Zhou}]{DBLP:conf/nips/Wei0SBIXCLZ22}
Jason Wei, Xuezhi Wang, Dale Schuurmans, Maarten Bosma, Brian Ichter, Fei Xia, Ed~H. Chi, Quoc~V. Le, and Denny Zhou. 2022.
\newblock {C}hain-of-{T}hought {P}rompting {E}licits {R}easoning in {L}arge {L}anguage {M}odels.
\newblock In \emph{{NeurIPS}}.

\bibitem[{Wen et~al.(2024)Wen, Ghosh, Staub, and Singla}]{chao2024xlogo}
Chao Wen, Ahana Ghosh, Jacqueline Staub, and Adish Singla. 2024.
\newblock {T}ask {S}ynthesis for {E}lementary {V}isual {P}rogramming in {X}{L}ogo{O}nline {E}nvironment.
\newblock In \emph{{AIED} Track on {L}ate {B}reaking {R}esults}.

\bibitem[{Wen et~al.(2025{\natexlab{a}})Wen, Phung, Mehrotra, Gulwani, Beaty, Nagashima, and Singla}]{DBLP:journals/corr/abs-2512-18388}
Chao Wen, Tung Phung, Pronita Mehrotra, Sumit Gulwani, Roger~E. Beaty, Tomohiro Nagashima, and Adish Singla. 2025{\natexlab{a}}.
\newblock {Exploration vs. Fixation: Scaffolding Divergent and Convergent Thinking for Human-AI Co-Creation with Generative Models}.
\newblock \emph{CoRR}, abs/2512.18388.

\bibitem[{Wen et~al.(2025{\natexlab{b}})Wen, Staub, and Singla}]{DBLP:conf/acl/WenSS25}
Chao Wen, Jacqueline Staub, and Adish Singla. 2025{\natexlab{b}}.
\newblock {Program Synthesis Benchmark for Visual Programming in XLogoOnline Environment}.
\newblock In \emph{{ACL}}.

\bibitem[{Wu et~al.(2025)Wu, Liang, Ge, Guo, Lu, Wang, Shan, and Luo}]{DBLP:conf/naacl/WuLGGLWSL25}
Chengyue Wu, Zhixuan Liang, Yixiao Ge, Qiushan Guo, Zeyu Lu, Jiahao Wang, Ying Shan, and Ping Luo. 2025.
\newblock Plot2code: {A} comprehensive benchmark for evaluating multi-modal large language models in code generation from scientific plots.
\newblock In \emph{{NAACL (Findings)}}.

\bibitem[{XLogoOnline(2024)}]{xlogoonline}
XLogoOnline. 2024.
\newblock {XL}ogo{O}nline {P}latform.
\newblock \url{https://xlogo.inf.ethz.ch/}.

\bibitem[{Xu et~al.(2024)Xu, Sun, Zheng, Geng, Zhao, Feng, Tao, Lin, and Jiang}]{DBLP:conf/iclr/XuSZG0FTLJ24}
Can Xu, Qingfeng Sun, Kai Zheng, Xiubo Geng, Pu~Zhao, Jiazhan Feng, Chongyang Tao, Qingwei Lin, and Daxin Jiang. 2024.
\newblock {WizardLM: Empowering Large Pre-Trained Language Models to Follow Complex Instructions}.
\newblock In \emph{{ICLR}}.

\bibitem[{Yang et~al.(2025)Yang, Shi, Liu, Shui, Wang, Jing, Xu, Zhu, Li, Zhang, Liu, Nie, Cai, and Yang}]{DBLP:conf/iclr/0002SLS0JXZLZLN25}
Cheng Yang, Chufan Shi, Yaxin Liu, Bo~Shui, Junjie Wang, Mohan Jing, Linran Xu, Xinyu Zhu, Siheng Li, Yuxiang Zhang, Gongye Liu, Xiaomei Nie, Deng Cai, and Yujiu Yang. 2025.
\newblock {ChartMimic: Evaluating LMM's Cross-Modal Reasoning Capability via Chart-to-Code Generation}.
\newblock In \emph{{ICLR}}.

\bibitem[{Yue et~al.(2024)Yue, Ni, Zheng, Zhang, Liu, Zhang, Stevens, Jiang, Ren, Sun, Wei, Yu, Yuan, Sun, Yin, Zheng, Yang, Liu, Huang, Sun, Su, and Chen}]{DBLP:conf/cvpr/YueNZ0LZSJRSWYY24}
Xiang Yue, Yuansheng Ni, Tianyu Zheng, Kai Zhang, Ruoqi Liu, Ge~Zhang, Samuel Stevens, Dongfu Jiang, Weiming Ren, Yuxuan Sun, Cong Wei, Botao Yu, Ruibin Yuan, Renliang Sun, Ming Yin, Boyuan Zheng, Zhenzhu Yang, Yibo Liu, Wenhao Huang, and 3 others. 2024.
\newblock {{MMMU:} {A} Massive Multi-Discipline Multimodal Understanding and Reasoning Benchmark for Expert {AGI}}.
\newblock In \emph{{CVPR}}.

\bibitem[{Zelikman et~al.(2022)Zelikman, Wu, Mu, and Goodman}]{star_cot_datasynthesis}
Eric Zelikman, Yuhuai Wu, Jesse Mu, and Noah Goodman. 2022.
\newblock {STaR: Bootstrapping Reasoning With Reasoning}.
\newblock In \emph{{NeurIPS}}.

\bibitem[{Zeng et~al.(2024{\natexlab{a}})Zeng, Liu, Lu, Wang, Liu, Dong, and Tang}]{DBLP:conf/acl/ZengLLWLD024}
Aohan Zeng, Mingdao Liu, Rui Lu, Bowen Wang, Xiao Liu, Yuxiao Dong, and Jie Tang. 2024{\natexlab{a}}.
\newblock {AgentTuning: Enabling Generalized Agent Abilities for LLMs}.
\newblock In \emph{{ACL (Findings)}}.

\bibitem[{Zeng et~al.(2024{\natexlab{b}})Zeng, Xu, Wang, Zhang, Yin, Rojas, Feng, Zhao, Lai, Yu, Wang, Sun, Zhang, Cheng, Gui, Tang, Zhang, Li, Zhao, Wu, Zhong, Liu, Huang, Zhang, Zheng, Lu, Duan, Zhang, Cao, Yang, Tam, Zhao, Liu, Xia, Zhang, Gu, Lv, Liu, Liu, Yang, Song, Zhang, An, Xu, Niu, Yang, Li, Bai, Dong, Qi, Wang, Yang, Du, Hou, and Wang}]{DBLP:journals/corr/abs-2406-12793}
Aohan Zeng, Bin Xu, Bowen Wang, Chenhui Zhang, Da~Yin, Diego Rojas, Guanyu Feng, Hanlin Zhao, Hanyu Lai, Hao Yu, Hongning Wang, Jiadai Sun, Jiajie Zhang, Jiale Cheng, Jiayi Gui, Jie Tang, Jing Zhang, Juanzi Li, Lei Zhao, and 36 others. 2024{\natexlab{b}}.
\newblock {ChatGLM: {A} Family of Large Language Models from {GLM-130B} to {GLM-4} All Tools}.
\newblock \emph{CoRR}, abs/2406.12793.

\bibitem[{Zheng et~al.(2024)Zheng, Zhang, Zhang, Ye, and Luo}]{zheng-etal-2024-llamafactory}
Yaowei Zheng, Richong Zhang, Junhao Zhang, Yanhan Ye, and Zheyan Luo. 2024.
\newblock {LlamaFactory: Unified Efficient Fine-Tuning of 100+ Language Models}.
\newblock In \emph{{ACL}}.

\bibitem[{Zhu et~al.(2025)Zhu, Wang, Chen, Liu, Ye, Gu, Tian, Duan, Su, Shao, Gao, Cui, Wang, Cao, Liu, Wei, Zhang, Wang, Xu, Li, Wang, Deng, Li, He, Jiang, Luo, Wang, He, Shi, Zhang, Shao, He, Xiong, Qu, Sun, Jiao, Lv, Wu, Zhang, Deng, Ge, Chen, Wang, Dou, Lu, Zhu, Lu, Lin, Qiao, Dai, and Wang}]{DBLP:journals/corr/abs-2504-10479}
Jinguo Zhu, Weiyun Wang, Zhe Chen, Zhaoyang Liu, Shenglong Ye, Lixin Gu, Hao Tian, Yuchen Duan, Weijie Su, Jie Shao, Zhangwei Gao, Erfei Cui, Xuehui Wang, Yue Cao, Yangzhou Liu, Xingguang Wei, Hongjie Zhang, Haomin Wang, Weiye Xu, and 32 others. 2025.
\newblock {InternVL3: Exploring Advanced Training and Test-Time Recipes for Open-Source Multimodal Models}.
\newblock \emph{CoRR}, abs/2504.10479.

\bibitem[{Zhuo et~al.(2025)Zhuo, Vu, Chim, Hu, Yu, Widyasari, Yusuf, Zhan, He, Paul, Brunner, Gong, Hoang, Zebaze, Hong, Li, Kaddour, Xu, Zhang, Yadav, and et~al.}]{DBLP:conf/iclr/ZhuoVCH0WYZHPB025}
Terry~Yue Zhuo, Minh~Chien Vu, Jenny Chim, Han Hu, Wenhao Yu, Ratnadira Widyasari, Imam Nur~Bani Yusuf, Haolan Zhan, Junda He, Indraneil Paul, Simon Brunner, Chen Gong, James Hoang, Armel~Randy Zebaze, Xiaoheng Hong, Wen{-}Ding Li, Jean Kaddour, Ming Xu, Zhihan Zhang, and 2 others. 2025.
\newblock {BigCodeBench: Benchmarking Code Generation with Diverse Function Calls and Complex Instructions}.
\newblock In \emph{{ICLR}}.

\bibitem[{Zou et~al.(2024)Zou, Cai, Zhang, and Lee}]{DBLP:conf/emnlp/ZouCZL24}
Bocheng Zou, Mu~Cai, Jianrui Zhang, and Yong~Jae Lee. 2024.
\newblock {VGBench: Evaluating Large Language Models on Vector Graphics Understanding and Generation}.
\newblock In \emph{{EMNLP}}.

\end{thebibliography}
